\documentclass[12pt,a4paper]{journal}

\usepackage[british]{babel}

\usepackage[a4paper,top=2cm,bottom=2cm,left=2.5cm,right=2.5cm,marginparwidth=1.75cm]{geometry}

\usepackage[final]{hyperref} 

\usepackage{amsmath}
\usepackage{graphicx}
\usepackage{hyperref}
\usepackage{orcidlink}
\usepackage[title]{appendix}
\usepackage{mathrsfs}
\usepackage{amsfonts}
\usepackage{booktabs} 
\usepackage{caption}  
\usepackage{threeparttable} 
\usepackage{algorithm}
\usepackage{algorithmicx}
\usepackage{algpseudocode}
\usepackage{listings}
\usepackage{enumitem}
\usepackage{chngcntr}
\usepackage{booktabs}
\usepackage{lipsum}
\usepackage{subcaption}
\usepackage{authblk}
\usepackage[T1]{fontenc}    
\usepackage{csquotes}       
\usepackage{diagbox}

\usepackage{color}
\usepackage{cite}
\usepackage{dirtree}
\usepackage{multicol}
\usepackage{gensymb}
\usepackage{rotating}
\usepackage{multirow}

\usepackage[final]{hyperref} 

\usepackage{setspace}
\usepackage{titlesec}
\titleformat{\section} 
  {\normalfont\Large\bfseries}{\thesection.}{1em}{}
  
\usepackage{lineno} 

\rightlinenumbers 

\usepackage{float}   
\usepackage{caption} 


\title{\textbf{PIV3CAMS: a multi-camera dataset for multiple computer vision problems and its application to novel view-point synthesis}}

\author[1]{\small Sohyeong Kim}
\author[2]{\small Martin Danelljan}
\author[2]{\small Radu Timofte}
\author[2]{\small Luc Van Gool}
\author[1]{\small Jean-Philippe Thiran}
\affil[1]{\small Signal Processing Laboratory 5, EPFL, Switzerland}
\affil[2]{Computer Vision Laboratory, ETH Zurich, Switzerland}

\date{}  

\begin{document}
\maketitle

\begin{abstract}
The modern approaches for computer vision tasks significantly rely on machine learning, which requires a large number of quality images. While there is a plethora of image datasets with a single type of images, there is a lack of datasets collected from multiple cameras. In this thesis, we introduce Paired Image and Video data from three CAMeraS, namely PIV3CAMS, aimed at multiple computer vision tasks. The PIV3CAMS dataset consists of 8385 pairs of images and 82 pairs of videos taken from three different cameras: Canon D5 Mark IV, Huawei P20, and ZED stereo camera. The dataset includes various indoor and outdoor scenes from different locations in Zurich (Switzerland) and Cheonan (South Korea). Some of the computer vision applications that can benefit from the PIV3CAMS dataset are image/video enhancement, view interpolation, image matching, and much more. We provide a careful explanation of the data collection process and detailed analysis of the data.
The second part of this thesis studies the usage of depth information in the view synthesizing task. In addition to the regeneration of a current state-of-the-art algorithm, we investigate several proposed alternative models that integrate depth information geometrically. Through extensive experiments, we show that the effect of depth is crucial in small view changes. Finally, we apply our model to the introduced PIV3CAMS dataset to synthesize novel target views as an example application of PIV3CAMS.
\end{abstract}


\tableofcontents
\section{Introduction}
\subsection{Motivation}
Computer vision is a field of artificial intelligence that aims to develop systems that can interpret and understand the visual world as the human visual system does. One of the earliest well-known projects in the computer vision field is the summer computer vision project at MIT in 1966\cite{papert1966summer}, intending to make a computer explain what it saw. Early attempts at scene understanding involved extracting edges\cite{roberts1963machine}, labeling lines\cite{Huffman1971ImpossibleOA, Clowes:1971:ST:3015362.3015366}, representing objects as interconnections of smaller surfaces, feature-based correspondence algorithms\cite{marr1979computational, grimson1984computational}, optical flows\cite{horn1981determining}, and motion estimation\cite{ullman1979interpretation}. Since then, much research has been done on the rigorous mathematical analysis of images and quantitative aspects of computer vision. Most modern studies on computer vision apply linear algebra, projective and differential geometry, as well as statistics in conjunction with machine learning techniques and complex optimization frameworks. For example, Convolutional Neural Networks algorithms\cite{lecun1999object} have already shown promising results with superior accuracy, even surpassing human-level accuracy on some tasks\cite{he2016deep}.

Several factors have converged to bring about a recent breakthrough in the computer vision field, including the advancement of hardware and the amount of data available. The growth of computing power has enabled the iterative learning process with deep layers of neural network architectures. The built-in cameras in smartphones and smaller digital cameras have made it easier to take photos and videos, resulting in a tremendous amount of data becoming accessible for research. The development of other types of image acquisition hardware, such as structured-light 3D cameras, radar imaging, Light Detection and Ranging (LIDAR) scanners, and magnetic resonance imaging, has given access to different types of images beyond RGB images, enriching computer vision research.

Over time, computer vision has partly merged with other closely related fields such as image processing, photogrammetry, and computer graphics, broadening its spectrum of potential applications. These applications of computer vision include autonomous navigation, robotic control processes, 3D scene reconstruction, novel view synthesis, virtual reality (VR), augmented reality (AR), and vision-based human-computer interfaces. As these applications become more popular in our lives, much research in computer vision has already started and produced notable progress.

As there is a wide range of applications for computer vision, the organization of its system is highly application-dependent. Not only the algorithm to complete a task but also the means of image acquisition and pre-processing those images are related to the application's specific objective. For example, for large-scale object detection, classification, and segmentation tasks, the ImageNet\cite{imagenet} and MS COCO\cite{mscoco} datasets can be perfect to use. On the other hand, for specific tasks such as house number detection, one can use SVHN (Street View House Numbers)\cite{netzer2011reading} for better performance. Some datasets, such as KITTI\cite{kitti}, have pre-processed the collected data in various formats so that it can be used for different tasks in applications for autonomous vehicles.

Much of the recent approaches for computer vision greatly rely on machine learning, which drives the need for benchmark datasets for training and evaluation.
While there is a plethora of image datasets with a single type of images, there is a lack of datasets collected from multiple imaging sensors. To address this gap, we have collected a paired image dataset. We have also chosen novel view synthesis as an example of a computer vision problem to investigate. Synthesizing new views of a scene becomes more critical as it is highly related to many compelling applications such as scene reconstruction, AR/VR, free-viewpoint image-based rendering, and more.


\subsection{Contribution}
\begin{figure}[h!]
  \centering
  \includegraphics[width = \textwidth]{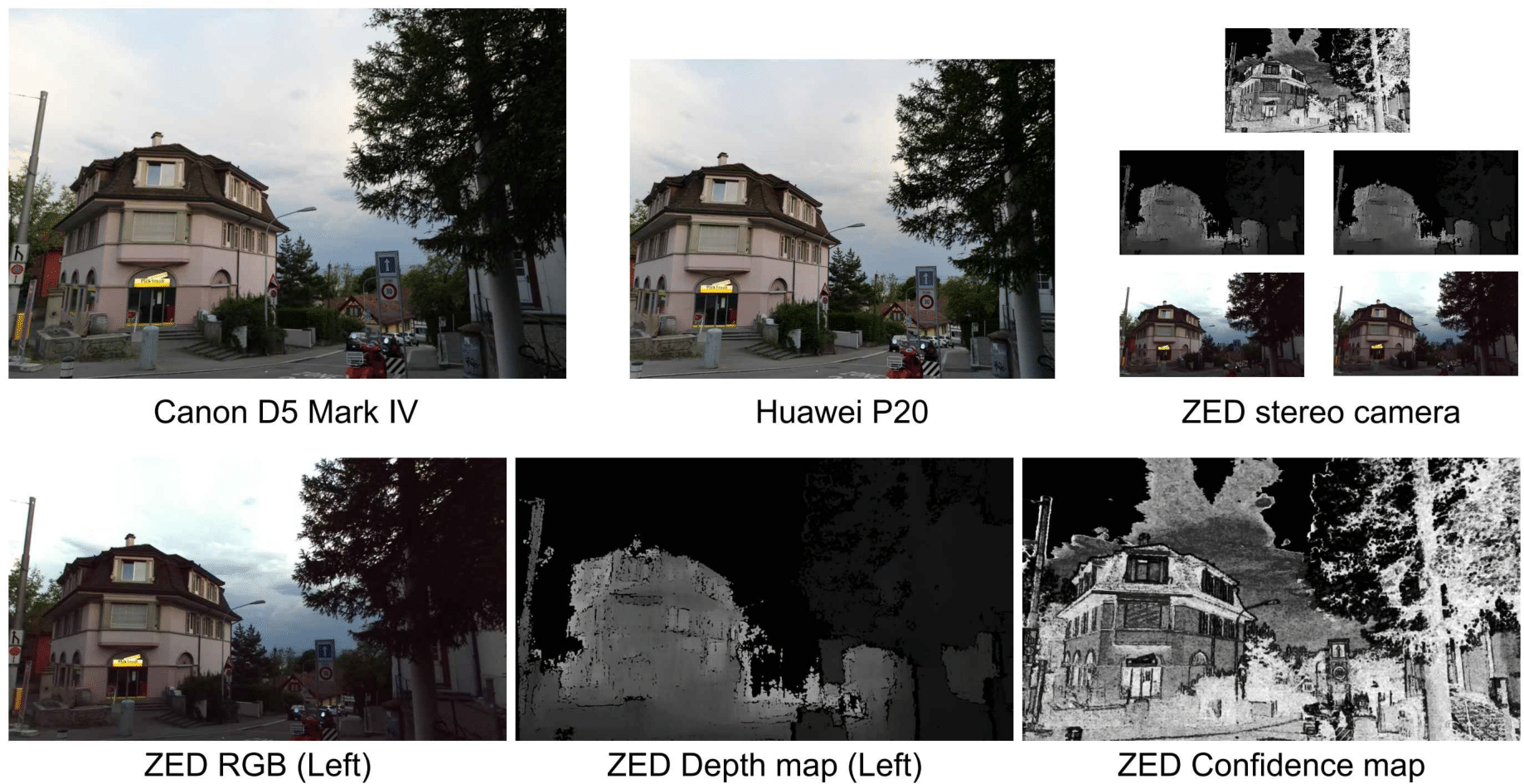}
  \caption{\textbf{Examples from PIV3CAMS dataset} The upper row shows images from different cameras with relative resolution sizes(Canon: 6720x4480, P20: 5120x3480, and ZED:2208x1242). The lower row shows the enlarged RGB-D image and a confidence map from ZED stereo camera.} 
  \label{fig:sampleImage}
\end{figure}

This thesis concerns two core parts of the modern computer vision pipeline, namely data collection and system design. For the first task of the thesis, we introduce Paired Image and Video data (Fig~\ref{fig:sampleImage}) from three CAMeras (PIV3CAMS) collected for multiple computer vision tasks. Next, investigating novel view synthesis with a learning-based approach is our second aim for the thesis. Specifically, we wanted to experiment with how using depth information would affect the performance of view synthesis.

As the first main contribution of this thesis, the PIV3CAMS dataset is proposed. It consists of images and videos from three different cameras (Huawei P20, Canon 5D Mark IV, and ZED stereo camera). These cameras are attached to a static rig, and we have collected data from two different cities: Zurich, Switzerland, and Cheonan, South Korea. The purpose of this dataset is to push forward the development of computer vision algorithms such as view interpolation, image matching, and image/video super-resolution. It includes RGB-D images and videos, depth confidence maps, and positions of the stereo camera for videos. Compared to other existing paired datasets, such as the KITTI\cite{kitti} dataset, which is mainly comprised of road and traffic scenes, our dataset includes various scenes for both outdoor and indoor settings obtained using hand-held devices.

\begin{figure}[h!]
  \centering
  \includegraphics[height =6cm ]{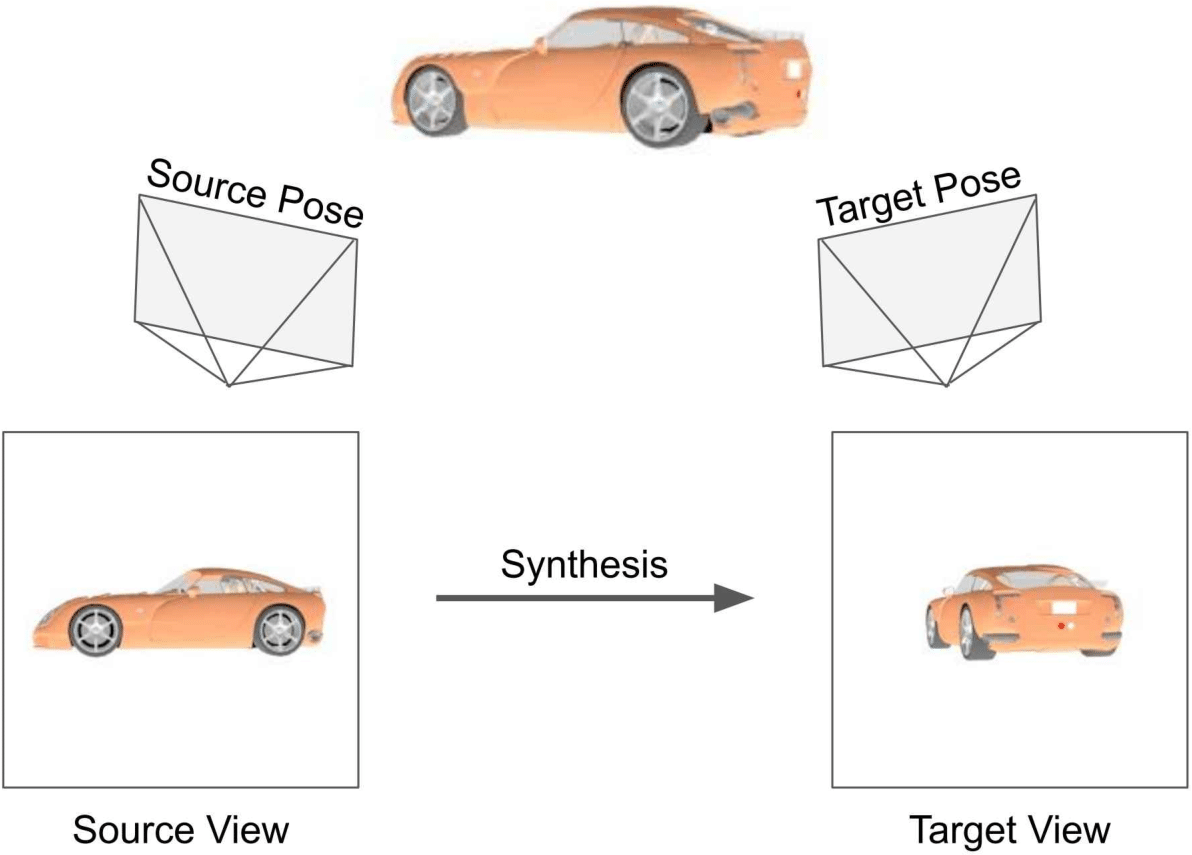}
  \caption{\textbf{Novel view Synthesis task} The goal of this task is to synthesize the target view from the source view.}
  \label{fig:nvsintro}
\end{figure}

As the second main contribution, this thesis investigates the problem of novel view synthesis (Fig~\ref{fig:nvsintro}). The goal of the novel view synthesis problem is to generate a new image from a different point of view from the given source image. While reproducing a state-of-the-art novel view synthesis algorithm\cite{chen2019nvs}, we investigated the effect of utilizing depth information for the view generation task. We have reorganized their networks by replacing the depth branch with a ground truth depth map to warp the source view. To see how sparse depth affects performance, we have also trained the networks with both dense and sparse depth maps. We used three different datasets to validate our approach: SYNTHIA\cite{synthia}, KITTI\cite{kitti}, and PIV3CAMS.

In summary, this work makes the following contributions:
(1) Introduction of a new image and video dataset aimed at multiple computer vision tasks.
(2) Reproduction of a state-of-the-art novel view synthesis algorithm.
(3) Investigation of the effect of depth information in the novel view synthesis task by creating a variant network model from the baseline network model.

In Section~\ref{ch2}, we will discuss existing datasets that aim to tackle similar tasks as PIV3CAMS and related works on novel view synthesis. In Section~\ref{ch3}, we will provide the technical details of the collection process and raw data of PIV3CAMS. We will also describe how the raw data is processed and rendered for specific tasks such as novel view synthesis and image and video super-resolution. In Section~\ref{ch4}, we will explain the baseline network for the novel view synthesis task and the variations we made to the networks to use depth information. In Section~\ref{ch5}, we will show the details of training and testing the networks, and we will discuss the results. Finally, in Section~\ref{ch6}, the conclusion and future work for the thesis will be presented.

\section{Related Works} \label{ch2}

\subsection{Datasets for Scene understanding}
In the last decade, many image datasets have been released, and they have played a significant role in computer vision research. This section discusses some of these datasets according to their data organization. Most of the datasets available for research are primarily focused on scene understanding, which includes tasks such as object detection and recognition, image classification, and segmentation. While some datasets, like DPED\cite{DPED}, aim to enhance image quality, we have included this dataset in our discussion because its data collection setup was similar to ours.

\subsubsection{Single-image datasets}
The ease of capturing and annotating single RGB images has led communities to release large-scale RGB image-only datasets. 

PASCAL VOC\cite{pascalvoc}, CIFAR-10, and CIFAR-100\cite{cifar} were among the earliest image datasets used for image classification tasks. The images were labeled in several classes: 20, 10, and 100 for PASCAL VOC, CIFAR-10, and CIFAR-100, respectively. For training and validation, images in the PASCAL VOC dataset were segmented, and objects within the images were annotated with bounding boxes.

ImageNet\cite{imagenet} is one of the largest image datasets, providing more than 14 million images, about 1 million of which are annotated with bounding boxes. It also provides descriptive words and SIFT features for the images.

More recent datasets such as COCO\cite{mscoco} and Open Images Dataset V5\cite{OpenImages} provide more complex annotations beyond just segmentation and bounding boxes to allow for cross-task training and analysis. For example, COCO provides panoptic segmentation by assigning a semantic label and instance ID for each pixel of an image. It also provides key points for people's poses if there are person instances. This allows data to be used for person detection and localization of key points for poses. The OpenImages Dataset V5 annotated all images already containing bounding box annotations with visual relationships between objects and other objects in the images. These visual relationship annotations pushed scene understanding further into 'images to words.'

Some datasets consist of single sets of videos with segmentation annotations. DAVIS\cite{davis} and YouTube-VOS\cite{youtubevos} provide object segmentation in the video frames. These datasets aim for video instance segmentation tasks.

The roles of single image and video datasets are very limited to object detection, classification, and recognition since they only provide 2D-limited information of the scenes. Table \ref{table:1} summarizes common datasets of single images or videos.

\begin{sidewaystable}[p]
\centering
\begin{tabular}{||p{4cm} p{1cm} p{3cm} p{3cm} p{3cm} p{3cm}||} 
 \hline
 Dataset & Year & \# Images & Video length \newline (in minutes) & \# Categories & \# Objects \newline annotated \\ [0.5ex] 
 \hline\hline
 PASCAL VOC\cite{pascalvoc} & 2010 & 500,000 & - & 80 & 27,450 \\ 
 CIFAR-10\cite{cifar} & 2009 & 60,000 & - & 10 & -\\
 CIFAR-100\cite{cifar} & 2009 & 60,000 & - & 100 & -\\
 ImageNet\cite{imagenet} & 2014 & 14,197,122 & - & 200 & 534,309 \\
 COCO\cite{mscoco} & 2015 & 2,500,000 & - & 80 & 886,284 \\ 
 OpenImages\cite{OpenImages} & 2017 & 9,178,275 & - & 600 & 15,440,132 \\
 DAVIS\cite{davis} & 2017 & 10,459 & 5.17 & - & 376 \\
 Youtube-VOS\cite{youtubevos} & 2018 & - & 334.81 & 94 & 197,272 \\[1ex] 
 \hline
\end{tabular}
\caption{Comparison of the single images and video datasets used for scene understanding tasks}
\label{table:1}
\end{sidewaystable}

\subsubsection{Paired datasets}
With the advancement of image acquisition hardware and sensors, it has become possible to capture images and videos of the same scene from different sensors. The visual data acquired from various sensors is formed into paired data. This additional paired data, on top of a single RGB image, provides researchers with extra information about the scenes. Table \ref{table:2} summarizes common datasets of paired images or videos.

One of the most popular ways of collecting paired images is using depth-sensing cameras such as Microsoft Kinect, Intel RealSense, and Asus Xtion. The paired data obtained from the depth-sensing camera is commonly called an 'RGB-D' image. The advantage of having depth information in an image is that it enables the extraction and utilization of the geometrical structure of scenes.

Some datasets are labeled or annotated with bounding boxes to provide a more general understanding of the world. NYU Depth V1\cite{nyuv1} and NYU Depth V2\cite{nyuv2} labeled images with dense multi-class labels. The Cornell-RGBD-Dataset\cite{cornellrgbd} labeled objects in a point cloud format. The Berkeley 3-D Object (B3DO)\cite{b3do} and RGB-D Scenes\cite{rgbdscene} datasets annotated objects with bounding boxes. SUNRGB-D\cite{sunrgbd} merged the RGB-D images from NYU Depth V2, B3DO, and SUN3D\cite{sun3d} and annotated the whole dataset with 2D polygons and 3D bounding boxes with accurate object orientations, as well as a 3D room layout and scene categories.

There are also some other datasets that go beyond the single, static-frame modality. SUN3D\cite{sun3d} and ViDRILO\cite{vidrilo} collected RGB-D data in video format. Specifically, the SUN3D dataset includes eight annotated sequences consisting of RGB-D images, camera poses, object segmentation, and point clouds registered into global coordinate frames.

While the majority of RGB-D datasets capture data in indoor environments, the DIML/CVL RGB-D dataset\cite{dimlcvl} comprises both indoor and outdoor RGB-D images. Compared to our dataset, the DIML/CVL RGB-D dataset consists of a much larger number of data, with 1M indoor images and 1M outdoor images. However, unlike our dataset, which includes RGB-D video data, their dataset is limited to image data.

Other than RGB-D datasets, another type of paired dataset using different cameras exists. DPED\cite{DPED} captured natural scenes using four different cameras by triggering them simultaneously. They used an iPhone 3GS, BlackBerry Passport, Sony Xperia Z, and Canon 70D DSLR to collect the data. This dataset is useful for training image enhancement algorithms, for example, to improve image quality from low to DSLR quality. Similarly, our dataset could be used for image enhancement tasks and further extended to video enhancement tasks.

\begin{sidewaystable}[p]
\centering
\begin{tabular}{||p{4cm} p{1cm} p{4cm} p{1.5cm} p{4.5cm} p{2cm}||} 
 \hline
 Dataset & Year & \# Images & Video & Sensors & Scene type \\ [0.5ex] 
 \hline\hline
 NYU V1\cite{nyuv1} & 2011 & 2283 frames & - & Kinect & indoor \\ 
 NYU V2\cite{nyuv2} & 2012 & 1449 frames \newline from 464 scenes & - & Kinect & indoor \\
 Cornell-RGBD dataset\cite{cornellrgbd} & 2011 & 52 scenes & - & Kinect & indoor\\
 B3OD\cite{b3do} & 2011 & 849 frames \newline from 79 scenes & - & Kinect & indoor \\
 RGB-D Scenes\cite{rgbdscene} & 2011 & 8 scenes & - & Kinect & indoor \\ 
 SUN RGB-D dataset\cite{sunrgbd} & 2015 & 10,335 frames & - & Kinect v1, Kinect v2, \newline Intel RealSense, \newline Axus Xiton & indoor \\
 SUN3D\cite{sun3d} & 2013 & 8 scenes & o & Kinect & indoor \\
 ViDRILO\cite{vidrilo} & 2015 & 22,454 frames \newline from 5 scenes & o & Kinect & indoor \\
 DIML/CVL RGB-D dataset\cite{dimlcvl} & 2018 & 1M frames (indoor), \newline 1M frames (outdoor) & - & Kinect v2, ZED stereo, Built-in Streo & indoor, outdoor \\
 DPED\cite{DPED} & 2017 & 16,291 frames & - & iPhone 3GS, BlackBerry Passport, Sony Xperia, Canon 70D DSLR & indoor \\[1ex] 
 \hline
\end{tabular}
\caption{Comparison of the paired images and video datasets used for scene understanding tasks}
\label{table:2}
\end{sidewaystable}

\subsubsection{Multi-modal datasets}
The scene understanding problem in computer vision has become more important with the rise of autonomous vehicles. Due to this specific application setting, several image and video datasets with a view from a car have been released, as summarized in Table~\ref{table:3}. These datasets typically comprise images from multiple cameras, range sensor (lidar, radar) data, and GPS/IMU data. Multi-modal datasets are expensive to collect and annotate due to the difficulties of integrating, synchronizing, and calibrating multiple sensors.

KITTI\cite{kitti} was the pioneering multimodal dataset, providing dense point clouds from lidar sensors as well as gray and color images and GPS/IMU data. It offers their raw dataset with various annotations and settings so that it can be used for many computer vision tasks.

The recently released H3D dataset\cite{h3d} includes 160 crowded traffic scenes with a total of 1 million labeled instances in 27k frames. It provides full 360-degree object annotations in point clouds, whereas the KITTI dataset annotates objects in front of the car. However, all cameras in H3D are front-facing, meaning that image data only covers 180 degrees. Additionally, both KITTI and H3D datasets do not provide nighttime data.

The KAIST multispectral dataset\cite{kaistdataset} is a multimodal dataset for driving scenarios that provides various illumination conditions, including nighttime.

The Cityscapes dataset\cite{cityscape} also collected urban street scenes in 50 different cities. It provides high-quality pixel-level annotations for 5k images and coarse annotations for 20k images.

BDD100k\cite{bdd100k} and Apolloscape\cite{apolloscape} released much larger datasets compared to other multimodal datasets, containing 100k and 144k RGB images, respectively. BDD100k also contains images captured in different weather conditions and illumination settings.

It is true that most of the multimodal datasets target the specific application of autonomous driving, so their visual data are limited to road scenes. Therefore, diversity in captured scenes is needed for broader applications.

\begin{sidewaystable}[p]
\centering
\begin{tabular}{||p{3cm} p{1cm} p{2cm} p{2cm} p{7cm} p{3cm}||} 
 \hline
 Dataset & Year & \# scenes & \# images & sensors & Locations \\ [0.5ex] 
 \hline\hline
 KITTI\cite{kitti} & 2012 & 22 & 150,000 & 2 Grayscale camera, 2 RGB cameras, 4 Varifocal lenses, Lidar, GPS/IMU & Karlsruhe \\ 
 H3D\cite{h3d} & 2019 & 160 & 83,000 & 3 RGB cameras, Lidar, GPS/IMU & San Francisco \\
 KAIST\cite{kaistdataset} & 2018 & - &89,000 & RGB-thermal camera, RGB stereo camera, Lidar, GPS/IMU & Daejeon \\
 Cityscapes\cite{cityscape} & 2016 & - & 25,000 & Stereo camera, GPS/IMU, Thermometer & Germany \\
 BDD100k\cite{bdd100k} & 2017 & 100,000 & 100M & RGB-thermal camera, RGB stereo camera, Lidar, GPS/IMU & NewYork, \newline San Francisco \\
 ApolloScape\cite{apolloscape} & 2018 & - & 144,000 &  4 RGB cameras, 2 laser scanners, GNSS/IMU & China \\ [1ex] 
 \hline
\end{tabular}
\caption{Comparison of the multi-modal datasets used for scene understanding tasks}
\label{table:3}
\end{sidewaystable}

\subsubsection{Synthetic datasets}
While real-world datasets are useful for enabling computer vision applications to work well in real-world scenarios, collecting all the data is very laborious. To avoid the difficulties in collecting real scenes, several synthetic datasets were generated via simulation.

ShapeNet\cite{shapenet} and ObjectNet3D\cite{objectnet3d} are large-scale databases of 3D objects classified into 3,135 and 100 categories, respectively. ShapeNet provides many semantic annotations for each 3D model, such as consistent rigid alignments, parts, and physical sizes.

SYNTHIA\cite{synthia}, CARLA\cite{carla}, and Virtual KITTI\cite{virtualkitti} simulated virtual urban areas using game engines and acquired photo-realistic frames rendered from multiple viewpoints in the simulated world. CARLA allows flexible configuration of the agent's sensor suite so that users can extract numerous frames of the same scene using different pseudo-sensors.

These synthetic datasets also provide a large amount of annotated images and measurements without requiring too much human labor. However, a domain gap still exists between synthesized images and real ones. In general, models learned in real scenarios generalize better in real-world applications.

\subsection{Novel view Synthesis}
Novel view synthesis is the task of synthesizing new views from a given source image. Its practical applications range from computer vision and computer graphics to virtual/augmented reality. The novel view synthesis problem aims to mimic humans' ability to build a mental understanding of the 3D properties of scenes from just 2D components. Humans can learn to combine and process all the information, such as viewpoints and shape understandings, to predict a 3D statistical understanding of the world. In computer vision, the integration of data from 2D images is not trivial, and understanding each type of information is tackled separately. In this section, we will introduce past research that has tackled novel view synthesis tasks.

\subsubsection{Geometry-based View Synthesis}
One of the classical approaches to view synthesis is directly estimating the underlying 3D structures of the object or scene and applying geometrical transformations to the pixels in the input\cite{geomethod01},\cite{geomethod02},\cite{geomethod03},\cite{geomethod04},\cite{geomethod05}. Given abundant source data from multiple views, one can interpolate the views by exploiting the 3D geometry of the scene\cite{mutliviewstereo}. \cite{motionclipfromdepth} and \cite{motionto3d} predicted depth maps from the stereo images created by small motion and then reconstructed the 3D structure of scenes. \cite{lightfieldrendering} and \cite{lf_frombasline} generated new views from arbitrary camera positions without depth information but with light field arrays rendered from images. This approach is successful as it can accurately transfer original colors, textures, and local features from the source view to the target view. However, it is still difficult to hallucinate new parts due to dis-occlusion, and it is unable to recover the desired target views with only a handful of images.

\subsubsection{Learning-based View Synthesis}
With the emergence of neural networks and the introduction of large-scale RGB-D datasets\cite{kitti} and synthetic 3D model datasets\cite{synthia},\cite{shapenet}, learning-based approaches have been applied to tackle novel view synthesis problems.

\cite{pixelgen01}, \cite{pixelgen02_multiviewfromSingleImg}, and \cite{pixelgen03_deepViewMorphing} built networks that learn how to generate pixels in the target views. Directly regressing pixels can generate structurally consistent results; however, it is prone to generating blurry results. To overcome this blurriness, \cite{flowpred01_deepstero}, \cite{flowpred02_cycleConsistency}, and \cite{viewSynAppearanceFlow} tried to predict the flows occurring from view changes. While these approaches could generate realistic textures, they were unable to generate regions not present in the source images. Recent research has merged these two concepts, pixel generation and flow prediction, to generate the target views \cite{TVSN}, \cite{multiview2novelview}, and \cite{nvsmachines}. They also used generative adversarial networks (GANs) to generate target views that are perceptually better than previous works.

Furthermore, some novel view synthesis frameworks such as \cite{pixelgen02_multiviewfromSingleImg}, \cite{multiview2novelview}, and \cite{stereoMag} aggregated multiple source views to derive the 3D statistical information of the objects or scenes. Aggregating multiple source views is only possible when there are images from numerous views of the same scene in the dataset. Additionally, it is challenging to incorporate understandings from various source images. On the other hand, \cite{TVSN} and \cite{nvsmachines} used a single source image to generate a target view by modeling auto-encoder networks to extract geometrically relevant features such as depth and masks. Then, by explicitly transforming geometric information, they could derive images with target views.


\section{Paired Image and Video dataset: PIV3CAMS dataset} \label{ch3}

In this chapter, we will provide a detailed description of the newly collected dataset, Paired Image and Video from 3 Cameras (PIV3CAMS) dataset. The primary motivation for creating this dataset is to promote applications related to paired images based on deep learning. We will describe how the data was collected and processed. We will then present the analysis results of the collected images and videos and propose some computer vision tasks where our dataset can be utilized.


\subsection{Motivation} \label{ch3.1}
With the increasing number of smartphone users in recent years, more and more people are taking pictures and videos with their smartphones. Even though advancements in smartphone camera lenses and image processing technology have minimized the gap in image and video quality between smartphones and DSLR cameras, there is still a need for better quality smartphone images and videos. At the same time, recent breakthroughs in AR/VR technology have produced many appealing applications in fields such as healthcare, education, e-commerce, and manufacturing. Many of these applications are facilitated by the availability of depth information during training and/or testing. Inspired by these needs, we have decided to create a dataset with various environments that could support different applications: image and video enhancement, novel view synthesis, and scene matching.

\begin{itemize}
    \item[] When designing and collecting a dataset, there are several important aspects to consider:

    \item \textbf{Selection of cameras} - What kinds of cameras should be used to cover the different computer vision applications we are motivated by? Which models should we choose?
    \item \textbf{Rig design} - How do we fix multiple cameras together?
    \item \textbf{Synchronization} - How are we going to trigger all cameras at the same time? 
    \item \textbf{Calibration} -  Which tools are we going to use to calibrate the cameras?
    \item \textbf{Locations} - Where do we collect the images and videos to ensure a variety of scenes?
    \item \textbf{Capture mode} - Which modes should we use when taking images and videos?
    \item \textbf{Resolution} - At what resolution should we capture the images and videos?

\end{itemize}
These aspects are carefully addressed in this section. Please refer to the following subsections for further details.

\subsection{Overview of the dataset}
\begin{figure}[p]
    \centering
    \includegraphics[width=\textwidth]{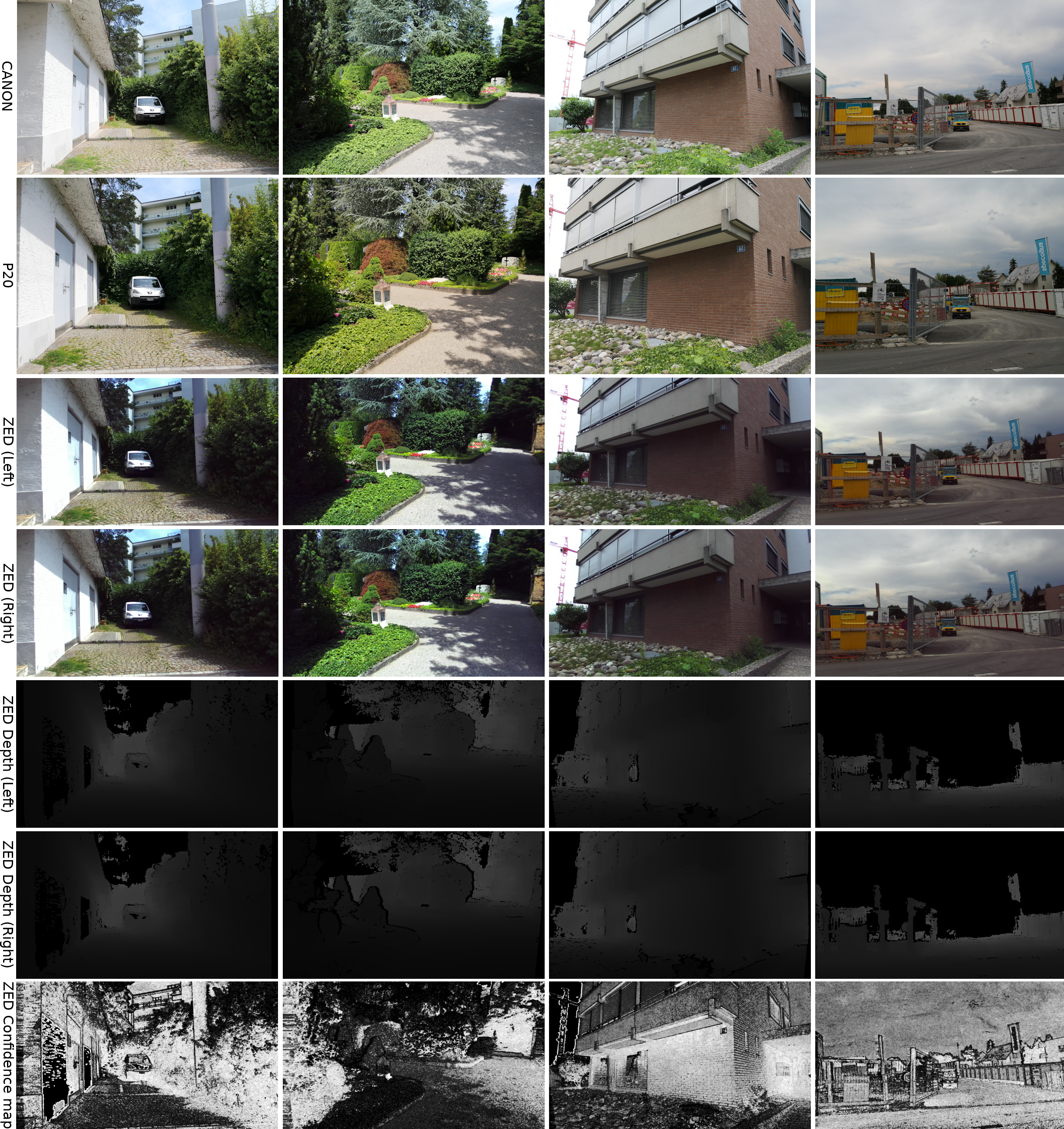}
    \caption{The samples from PIV3CAMS image dataset}
    \label{fig:image_samples}
\end{figure}

\begin{figure}[h!]
    \centering
    \includegraphics[width=\textwidth]{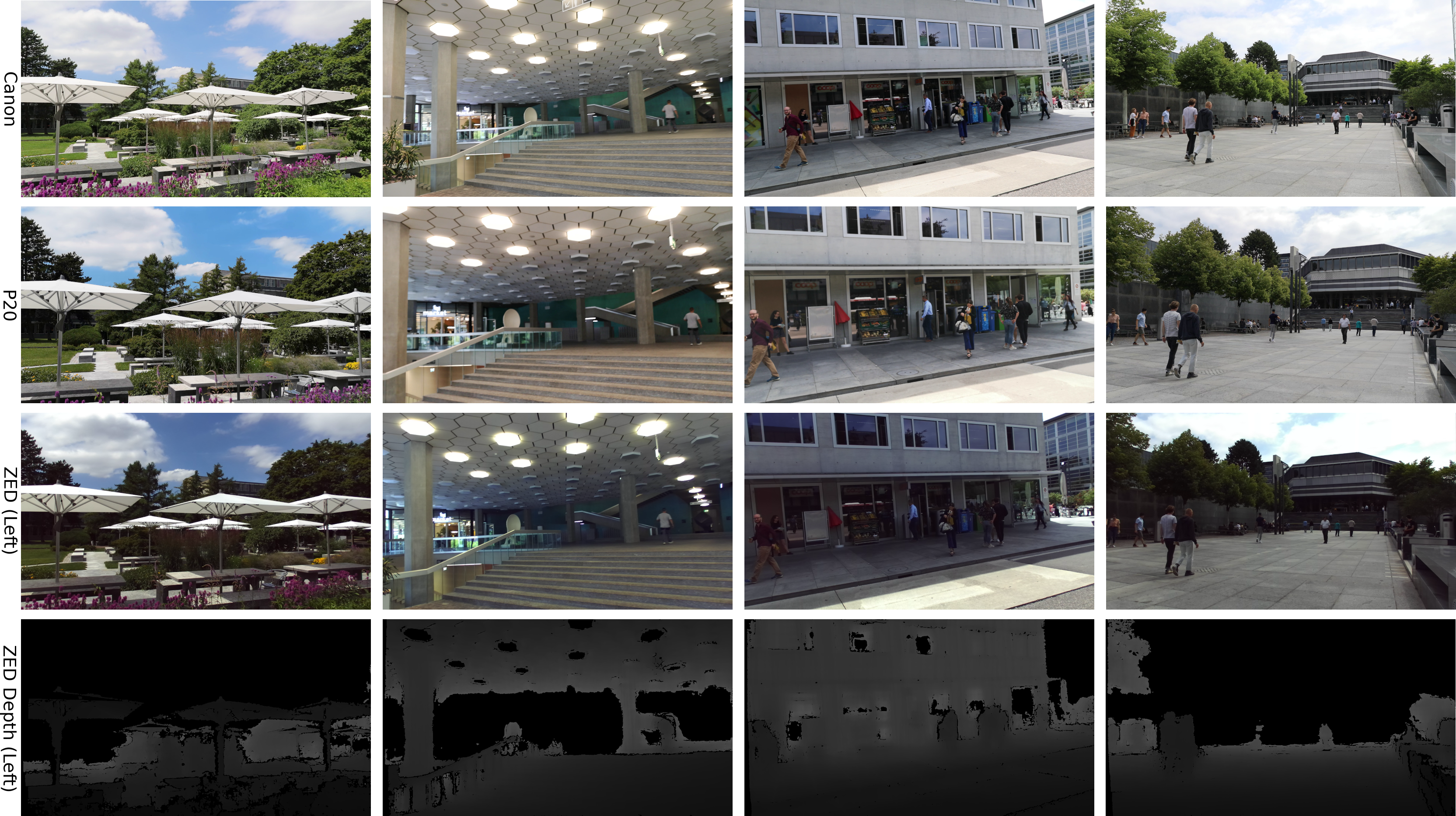}
    \caption{The frame samples from PIV3CAMS video dataset}
    \label{fig:video_samples}
\end{figure}

The dataset consists of over 8,000 sets of paired images and 82 different video scenes acquired by three different cameras: Huawei P20, Canon 5D Mark IV, and ZED stereo camera. This dataset was collected over two months, from June to July 2019, in Zurich (Switzerland) and Cheonan (South Korea), both indoors and outdoors. Unlike existing outdoor data using stereo cameras mounted on a moving vehicle, ours was taken using handheld stereo cameras. Table~\ref{table:pic3cam_img} and Table~\ref{table:pic3cam_vid} summarize the details of the image and video collection settings, respectively. Figures \ref{fig:image_samples} and \ref{fig:video_samples} show subsets of the dataset as examples.

\begin{sidewaystable}[p]
\centering
\begin{tabular}{||p{3.5cm} | p{4cm} p{4cm} p{7cm}||} 
 \hline \hline
  & Huawei P20 & Canon Mark IV 5D & ZED stereo \\ [0.5ex]
 \hline\hline
 Color Resolution & 5120 x 3840 (4:3) & 6720 x 4480 (3:2) & 2208 x 1242 (16:9) \\
 Depth Resolution & & & 2208 x 1242 (16:9) \\
 Range of Depth & & & 0.5m to 20 m \\ 
 Focus Mode & Auto focus & Auto focus & Auto focus \\
 Raw file Format & DNG & CR2 & SVO \\
 Data Acquisition & RGB images, \newline RAW files & RGB images, \newline RAW files & Rectified left and right RGB images, Left and right depth images, Left confidence maps, Calibration parameters, RAW files \\
 \hline
 \multicolumn{1}{||p{3cm}|}{Collection Area} & \multicolumn{3}{c||}{Zurich, Choenan, Indoor, Outdoor} \\[1ex]
 \hline
 \hline
\end{tabular}
\caption{Summary of PIV3CAMS dataset settings for \textbf{image}}
\label{table:pic3cam_img}
\end{sidewaystable}

\begin{sidewaystable}[p]
\centering
\begin{tabular}{||p{3.5cm} | p{4cm} p{4cm} p{7cm}||} 
 \hline \hline
  & Huawei P20 & Canon Mark IV 5D & ZED stereo \\ [0.5ex]
 \hline\hline
 Color Resolution & 3840x2160 (16:9) & 1920 x 1080 (16:9) & 1920x1080 (16:9) \\
 Depth Resolution & & & 1920x1080 (16:9) \\
 Range of Depth & & & 0.5m to 20 m \\ 
 FPS & 29.76 & 29.97 & 30 (10 fps for depth video) \\
 Video Codec & H.265(HEVC) & H.264(AVC) & Lossless(PNG) \\
 Data Acquisition & RGB Videos & RGB Videos & RGB videos(rectified left), \newline Depth videos(left), Calibration parameters, Poses of the camera \\
 \hline
  \multicolumn{1}{||c|}{Video lengths} & \multicolumn{3}{c||}{between 25 and 50 seconds} \\
 \multicolumn{1}{||c|}{Collection Area} & \multicolumn{3}{c||}{Zurich, Indoor, Outdoor} \\[1ex]
 \hline
 \hline
\end{tabular}
\caption{Summary of PIV3CAMS dataset settings for \textbf{videos}}
\label{table:pic3cam_vid}
\end{sidewaystable}

\subsection{Setup}
\subsubsection{Hardware: Selection of cameras}
We aimed to select cameras that differ in their image and video quality to cover applications such as image and video enhancement. At the same time, we wanted our dataset to provide 3D information about the scenes. Therefore, we selected three different cameras: Huawei P20 (P20), Canon 5D Mark IV (Canon), and ZED stereo camera (ZED), each representing a smartphone camera, DSLR camera, and 3D camera, respectively.

We wanted to use a smartphone well-known for its camera performance. There were many models we could choose for smartphones: Samsung Galaxy series, Apple iPhones, and more. Among them, we decided on the Huawei P20 since it was one of the top-tier smartphones at the time of this thesis and is known for its good camera.

Similarly, we wanted to use a high-performance DSLR camera, so we chose the Canon 5D Mark IV. It is a professional-grade 30.1-megapixel full-frame DSLR camera and was one of the best DSLR cameras at the time of this work.

The difference in sensor size, image processor, and lenses makes the DSLR camera special. Notably, the sensor size of the Canon is 36.0x24.0mm, about 20 times larger than the sensor size of the P20, which is 1/1.7-inch (7.6x5.7mm), and 35 times larger than the sensor size of the ZED, which is 1/3-inch (4.8x3.6mm). The bigger the sensor size, the better the low-light sensitivity, depth of field, and lower diffraction the camera has.

For the 3D camera, we chose the ZED due to its high resolution and high frame rate for capturing 3D images and videos. Compared to other well-known 3D cameras such as Microsoft Kinect v2 and Intel RealSense D415/D435, the ZED has a much greater depth-sensing range, up to 20 meters indoors and outdoors. Additionally, it has better light sensitivity compared to active Time-of-Flight 3D cameras like the Kinect.

\subsubsection{Hardware: Rig design}
\begin{figure}[h]
  \centering
  \includegraphics[height = 7cm]{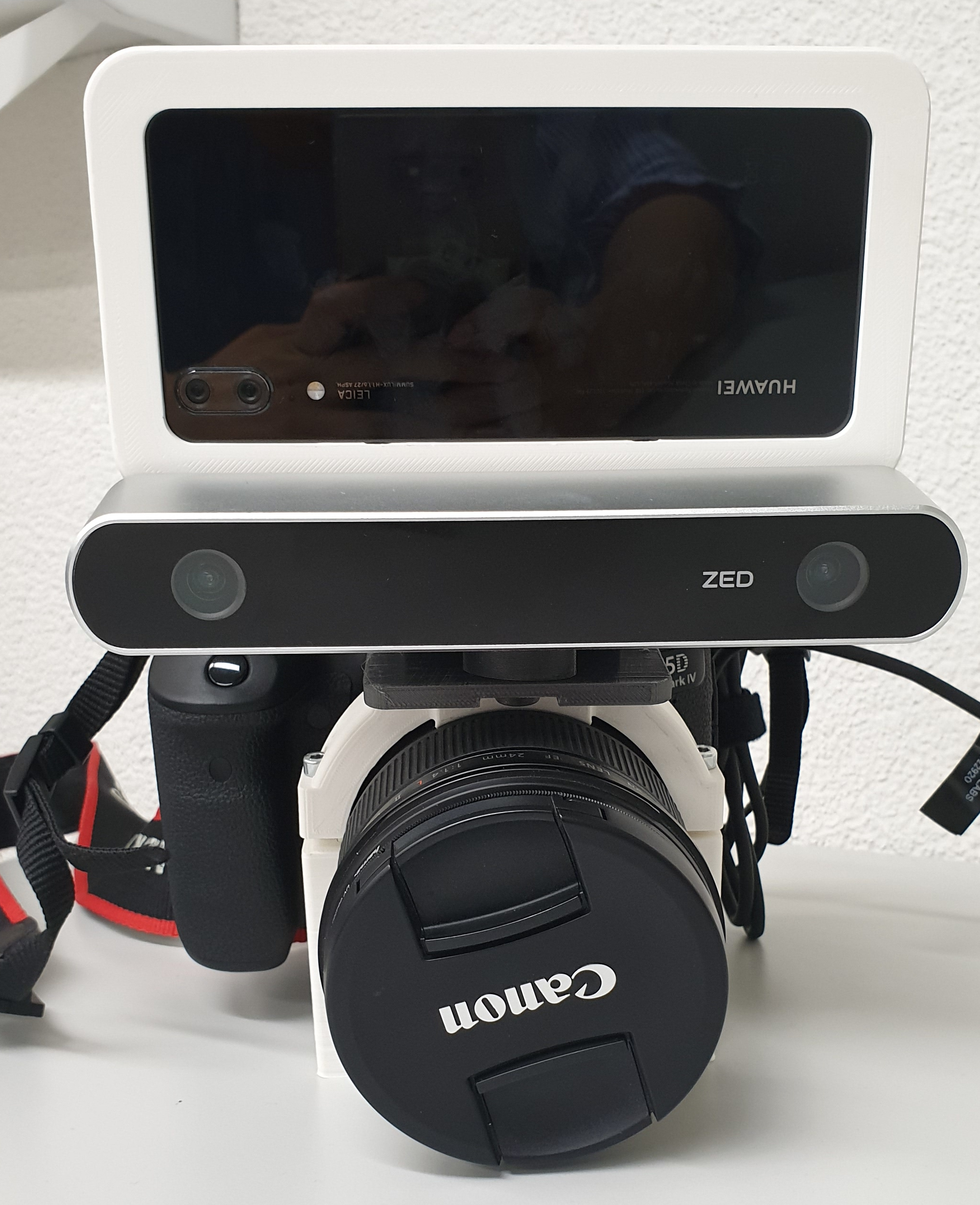}
  \includegraphics[height = 7cm]{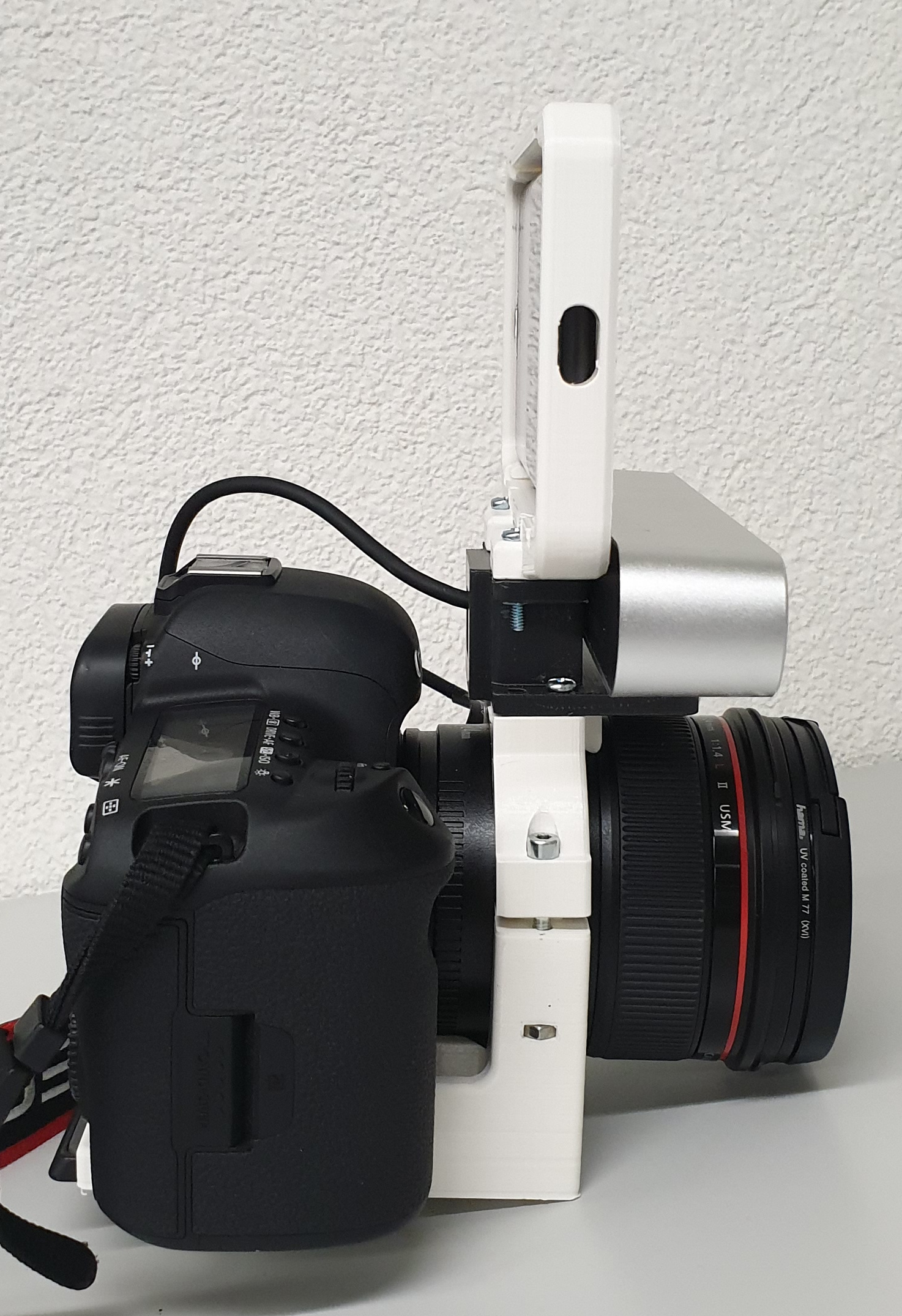}
  \caption{\textbf{Rig setup for cameras}. (Left) Front view of the 3D printed rig  with cameras mounted on it. (Right) Side view of the rig with cameras mounted on it.} 
  \label{fig:rigsetup2}
\end{figure}

To fix the position of the cameras, we printed a customized rig with a 3D printer and mounted the cameras on it, as shown in Figure \ref{fig:rigsetup2}. These were then attached to a tripod for stabilization. A laptop (Dell XPS 15-9570) with an Intel Core i7-8750H CPU (12 x 2.20GHz), 16GB DDR4 RAM, and a 4GB GPU (GeForce GTX 1050 Ti) was used throughout the data collection.

\begin{figure}[h!]
    \centering
    \includegraphics[height = 7cm]{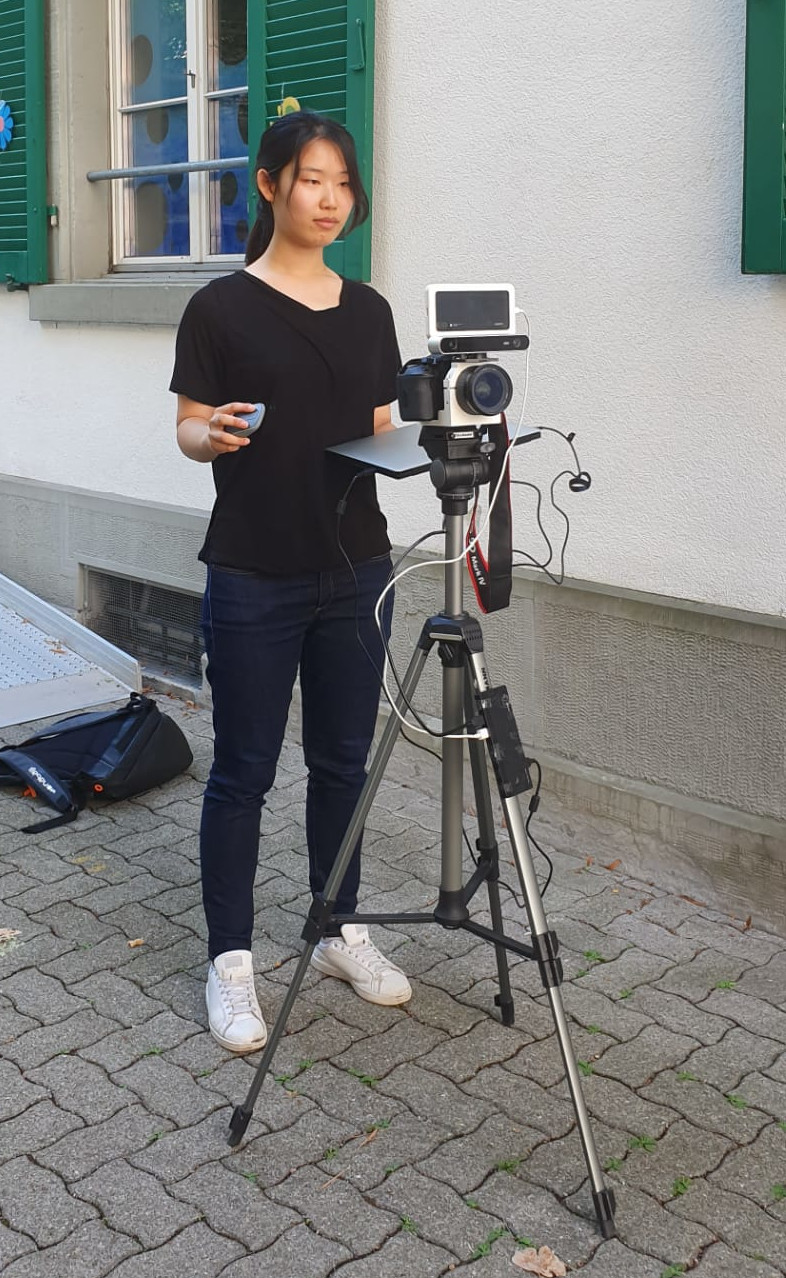}
    \includegraphics[height = 7cm]{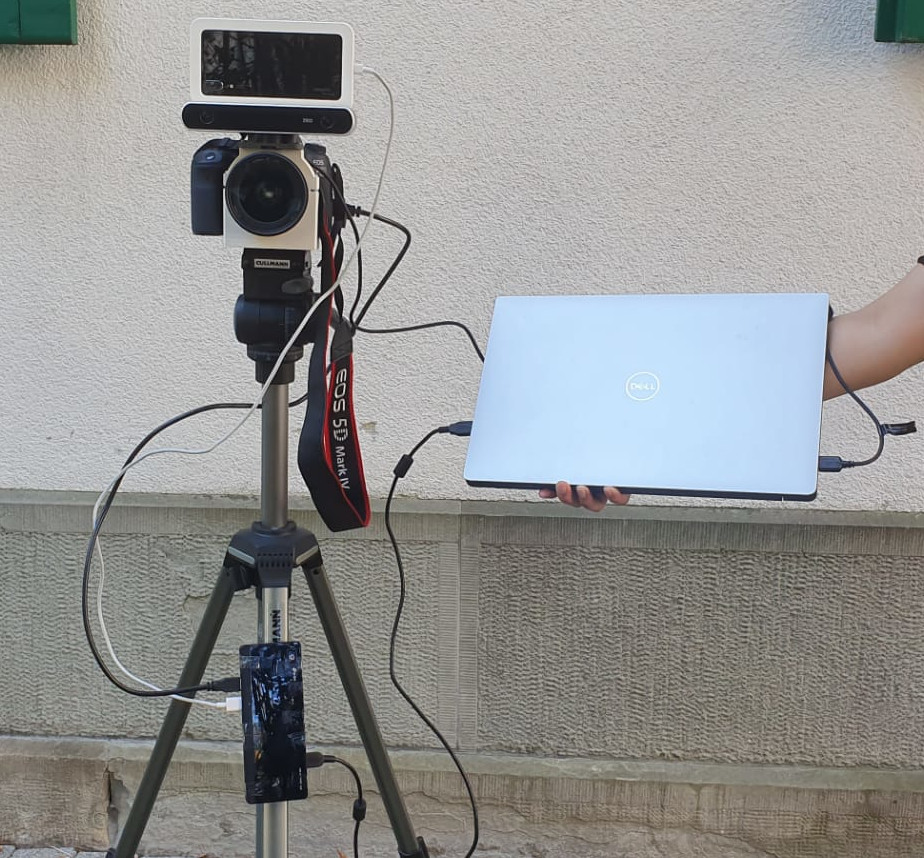}
    \caption{\textbf{Overall setup for data collection}. (Left)All scenes were captured steadily with a stationary tripod. The collector used a mouse to trigger cameras while carrying the laptop. (Right) The Canon and P20 cameras were connected to a USB hub, while the ZED camera was connected directly to the laptop.} 
    \label{fig:capture_setup}
\end{figure}

To achieve precise multi-sensor data alignment between cameras and to make collection easier, we developed scripts that trigger multiple cameras concurrently. The script runs on Linux OS (Ubuntu 18.04.3 LTS) and requires ZED SDK version 2.8 with the CUDA 10 library. With a click of the mouse's left button, the script triggers the shooting of the three cameras connected to the laptop.

Due to a lack of USB ports on the laptop, we connected the P20 and Canon to the laptop using a USB hub (Fig \ref{fig:capture_setup}). The ZED camera was connected directly to the computer through a USB 3.0 port, as the USB hub could not source enough power to run the ZED camera.

\subsubsection{Synchronization: Triggering multiple cameras at the same time}
\begin{figure}[h]
      \centering
      \includegraphics[width=0.8\textwidth]{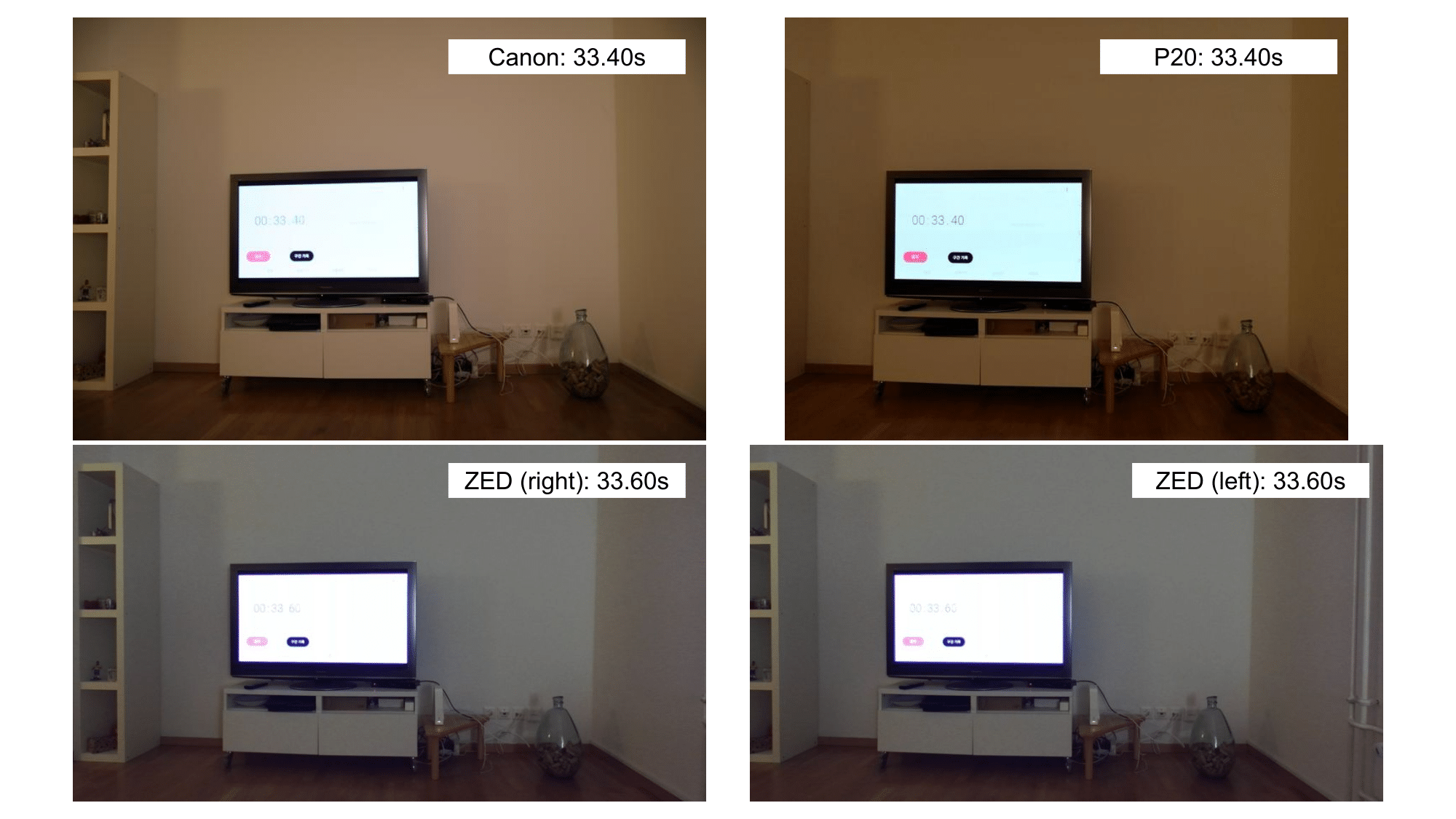}
      \caption{\textbf{Synchronization test}. Testing simultaneous camera triggering using the script. Images are captured from Canon, P20, ZED right, and ZED left (left to right, top to bottom). Over 20 tests were conducted to average synchronization time differences for capturing images and videos.} 
      \label{fig:synctest}
    \end{figure}

We tested the cameras' exposure time using a timer and achieved an average time difference of less than 0.5 seconds, which can be regarded as instantaneous exposure.  Figure~\ref{fig:synctest} shows the images testing the synchronization.

\subsubsection{Calibration: Stereo camera calibration}
\begin{figure}[h]
      \centering
      \includegraphics[width = 0.28\textwidth]{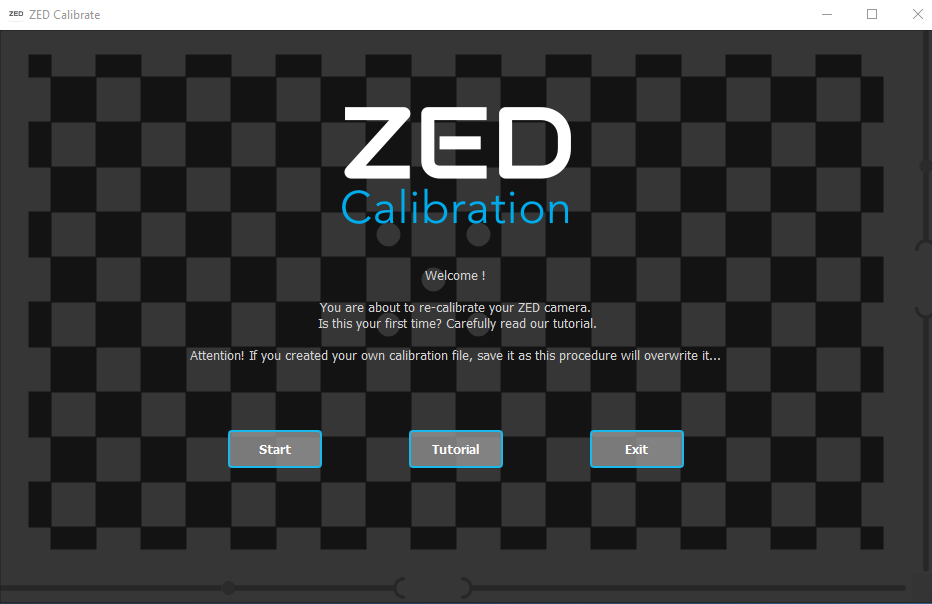}
      \includegraphics[width = 0.32\textwidth]{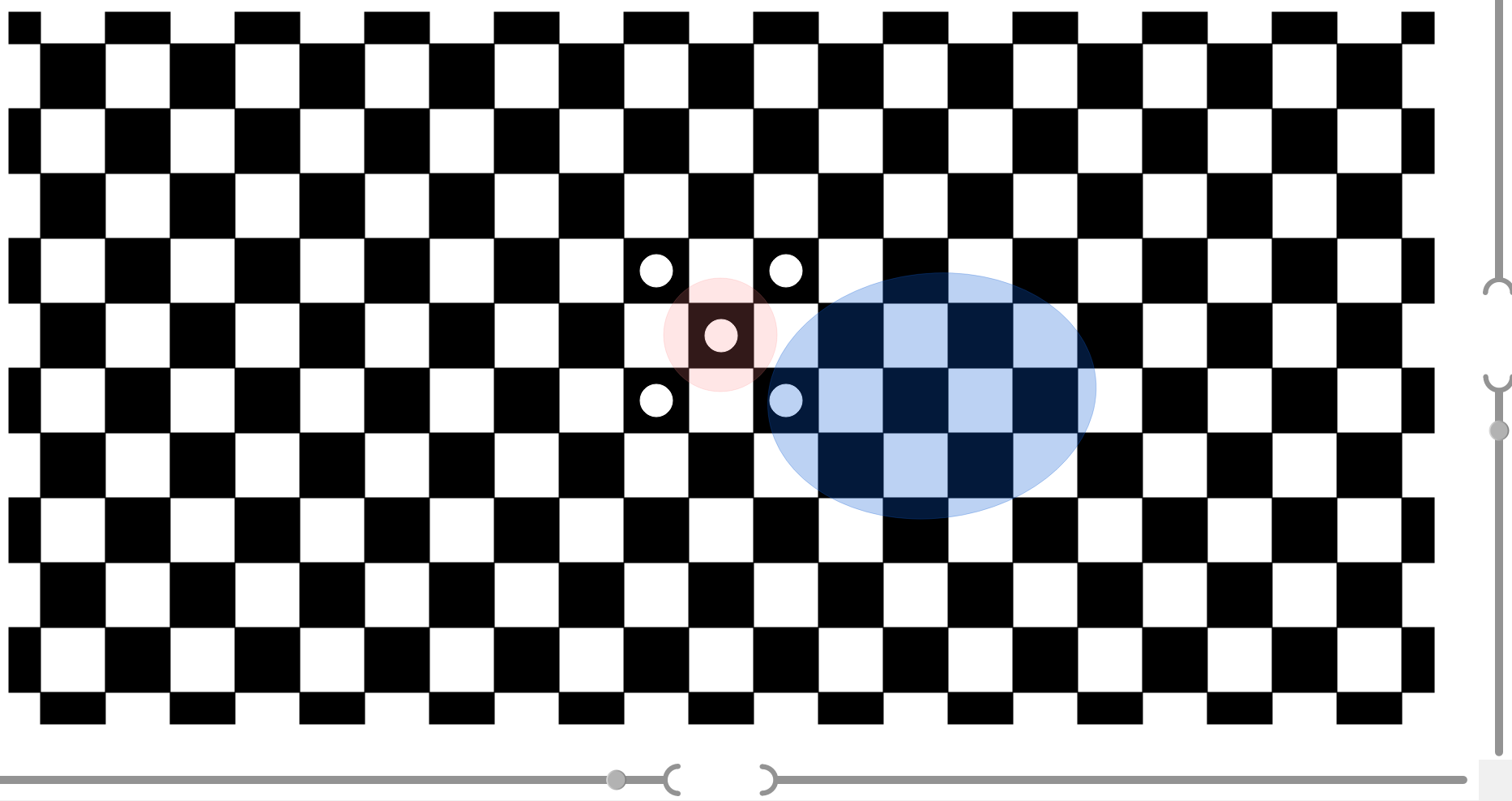}
      \includegraphics[width = 0.32\textwidth]{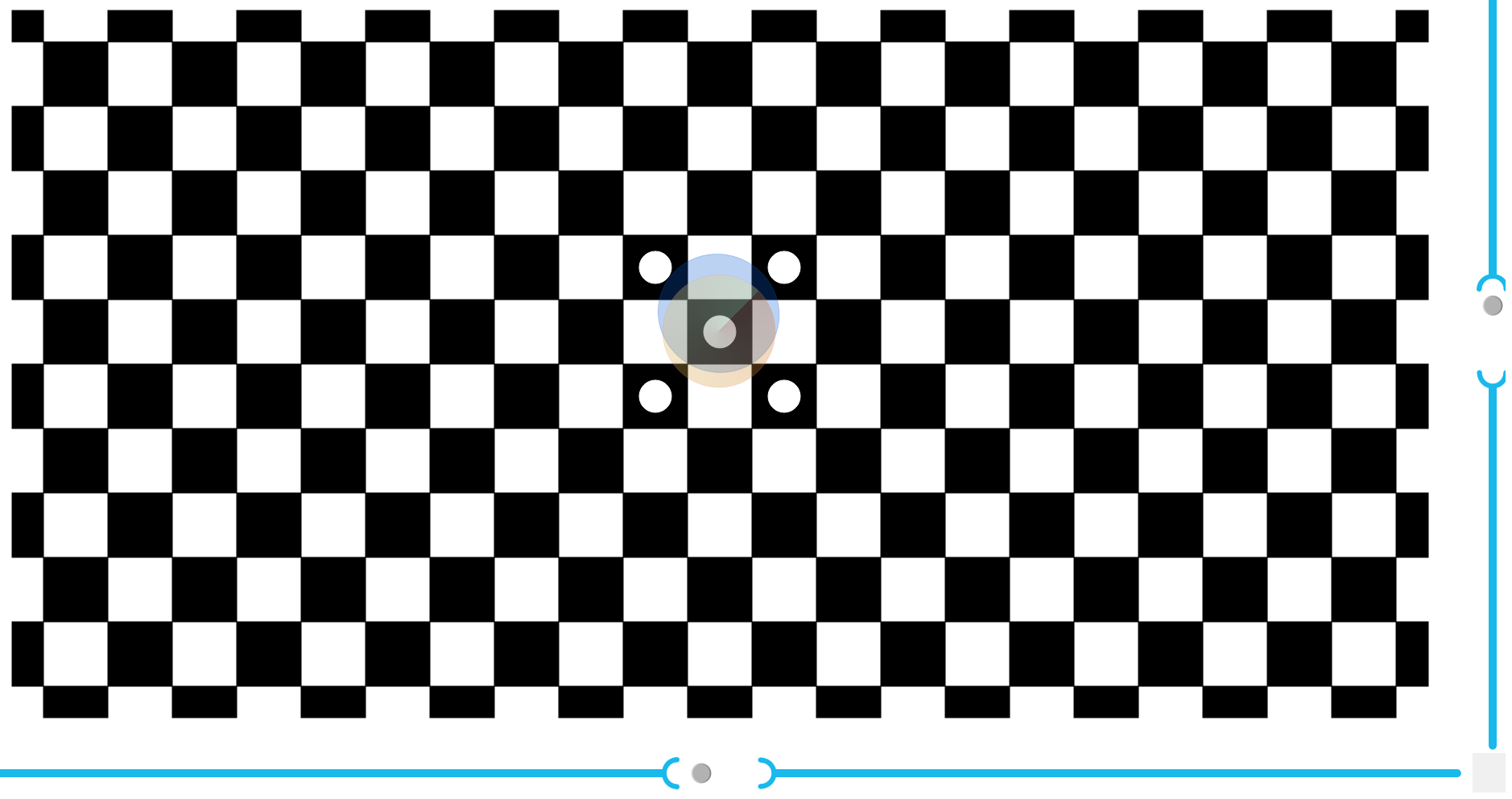}
      \caption{\textbf{Calibration of the ZED stereo camera using ZED SDK}. The goal is to align the blue dot (ZED camera view) with the red dot.} 
      \label{fig:calibarionZED}
    \end{figure}

We performed the calibration for the ZED camera using the calibration toolbox provided with the ZED SDK. This calibration app works by having the ZED camera recognize checkerboard patterns shown on a monitor (Fig~\ref{fig:calibarionZED}). Calibration is completed after moving the ZED camera further back and tilting it in each direction through several steps. The calibrated parameters are used to rectify the raw images from the ZED camera and compute the depth map of the scenes. These calibrated parameters are included within our dataset.

\vspace{-1em}   
\subsection{Data Acquisition}
\subsubsection{Locations and times}
\begin{figure}[p]
  \centering
  \includegraphics[width = \textwidth]{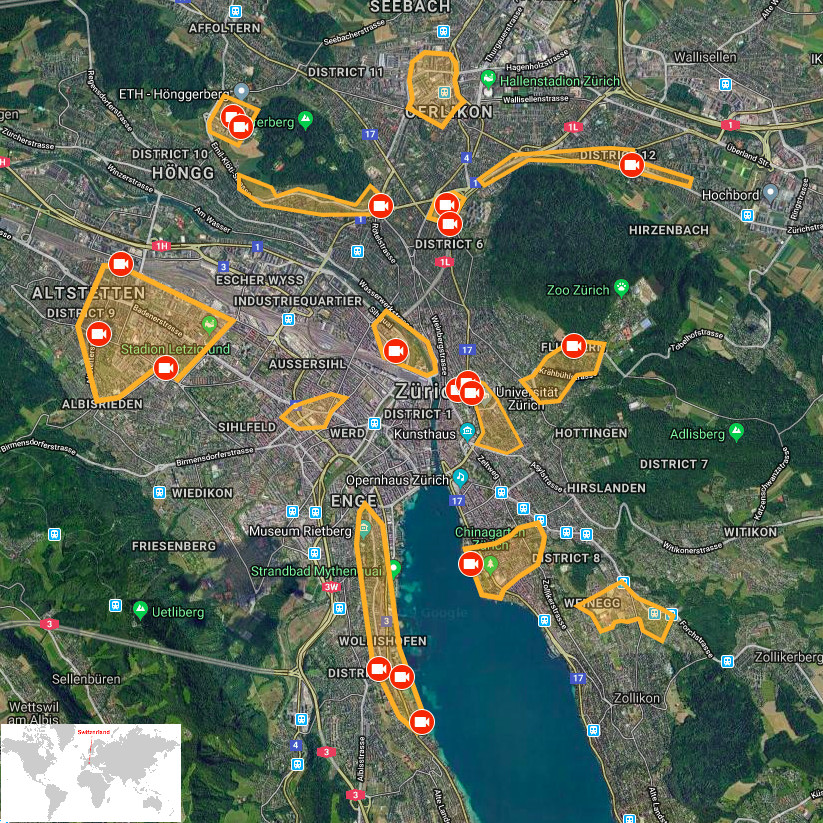}
  \includegraphics[width = \textwidth]{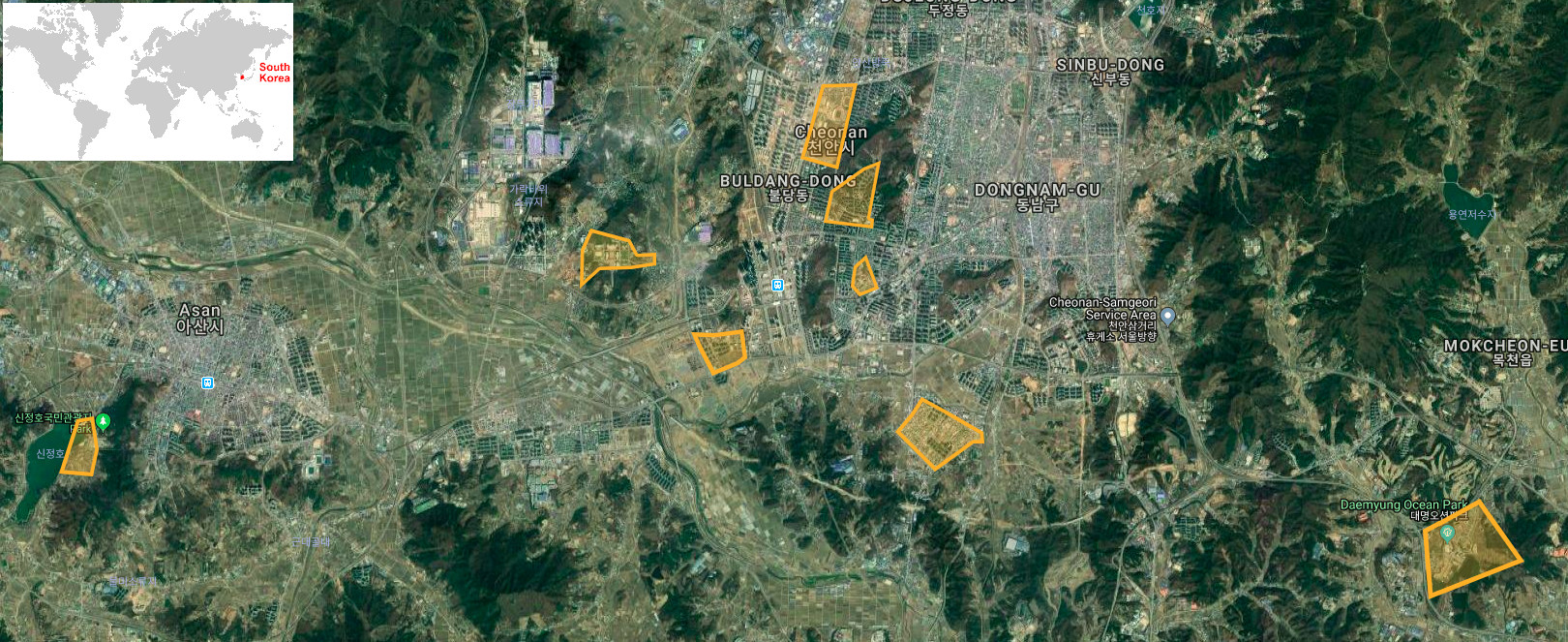}
  \caption{\textbf{Locations of data collection}. The map shows data collection areas in Zurich, Switzerland (top) and Cheonan, South Korea (bottom). Orange-shaded areas indicate where images were taken, and red pins mark video recording locations.} 
  \label{fig:recordedzone}
\end{figure}

The PIV3CAMS dataset was obtained using cameras mounted on a tripod, capturing various outdoor and indoor scenes from different locations in Zurich (Switzerland) and Cheonan (South Korea), as shown in Figure~\ref{fig:recordedzone}.

As mentioned in section \ref{ch3.1}, we aimed to ensure a variety of scenes in our dataset. Therefore, we made a list of scene types before choosing the locations. Some examples from the list include: \textit{business area}, \textit{residential area}, \textit{playground}, \textit{construction site}, \textit{park}, \textit{lake}, \textit{cafeteria}, \textit{office}, and \textit{gym}. We then carefully selected areas within the cities for shooting to minimize overlap between scene types. To ensure diversity in the dataset, we collected images in the early morning, afternoon, evening, and at night. 

\subsubsection{Image data collection and processing}
\textbf{Capture mode and resolution}
We captured images in the AUTO mode of each camera since it is the most commonly used mode. We also set the image resolution to the maximum achievable for each camera for the best quality pictures.

The image resolutions are 6720$\times$4480, 5120$\times$3840, and 2208$\times$1242 for Canon, P20, and ZED, respectively. For the Canon and P20, we saved both the RAW files and JPEG-compressed RGB images. For the ZED, we recorded images in Stereolab's SVO format, which contains extra metadata such as timestamps and sensor data. We chose the LOSSLESS (PNG) compression mode for recording the SVO files. 

\vspace{1em}
\textbf{Processing}
After the collection, we pre-processed the dataset to make it suitable for research. First, we processed the ZED's SVO files to extract images. Using the ZED API, we loaded recorded SVO files and extracted left and right RGB images, left and right depth images, depth confidence maps, and other calibration parameters. When extracting depth images, we chose the ULTRA mode, which offers the highest depth range and better preserves Z-accuracy along with the sensing range (from 0.5 to 20 meters). We stored depth images in millimeters using a 16-bit PNG format. Next, we manually sorted out pairs of photos corrupted by blurriness, light reflection, and incorrect color mapping.

\vspace{1em}
\textbf{Data format}
Here, we describe the structure and format of the image dataset. There are a total of 8,385 pairs of images, and the total size of the dataset is over 900GB (approximately 300GB for compressed images and 600GB for RAW files). In addition to the image files, we provide the timestamps for when each pair of images was taken and the calibration parameters for the ZED camera.

\begin{multicols}{2}
\footnotesize
\dirtree{%
.1 IMAGE\_DATA.
.2 Compressed.
.3 CANON.
.4 \textbf{Canon\_[6digits].JPG}.
.3 P20.
.4 \textbf{P20\_[6digits].jpg}.
.3 ZED.
.4 color.
.5 left.
.6 \textbf{ZED\_left\_[6digits].png}.
.5 right.
.6 \textbf{ZED\_right\_[6digits].png}.
.4 depth.
.5 left.
.6 \textbf{ZED\_depth\_left\_[6digits].png}.
.5 right.
.6 \textbf{ZED\_depth\_right\_[6digits].png}.
.4 confidence.
.5 \textbf{ZED\_confidence\_map\_[6digits].png}.
}
\columnbreak
\dirtree{%
.1 IMAGE\_DATA.
.2 RAW.
.3 CANON.
.4 \textbf{Canon\_[6digits].CR2}.
.3 P20.
.4 \textbf{P20\_[6digits].dng}.
.3 ZED.
.4 \textbf{ZED\_[6digits].svo}.
.3 \textbf{calib.txt}.
.3 \textbf{time.txt}.
}
\end{multicols}

\subsubsection{Video data collection and processing}
\textbf{Capture mode and resolution}
Similar to the image collection, we captured videos using the DEFAULT video recording modes of each camera. We set the frame rate to 30 frames per second (fps) and configured the resolution to the maximum achievable at 30 fps. 
The video frame resolutions are 1920$\times$1080, 3840$\times$2160, and 1920$\times$1080 for Canon, P20, and ZED, respectively. For the Canon and P20, the videos are compressed using H.264 and H.265 coding methods, respectively. For the ZED camera, the video is recorded in SVO format with LOSSLESS (PNG) mode.

We recorded the videos in two different settings: one with a static camera position and the other with a moving camera position. In the first setting, the tripod was fixed in one spot, and the cameras were moderately panned to minimize fluctuations from walking. In the second setting, the videos were recorded while walking to mimic a natural recording situation.
Although we synchronized the triggering time of all cameras to within 0.5 seconds, this time difference can result in up to a 15-frame misalignment in the videos. To reduce frame shifts between videos, we displayed a timer at the beginning of each video.

\vspace{1em} \noindent
\textbf{Processing}
Similar to the image collection, we pre-processed the collected videos. First, we extracted frames from each camera's videos. We extracted ZED's SVO files into both RGB and depth frames from the left lens of the ZED camera. For depth images, we used the ULTRA mode as we did for the image sets, with a sensing range from 0.5 to 20 meters. Depth frames are stored in millimeters using a 16-bit PNG format. Canon's and P20's videos were also converted into frames.
We then matched the starting frame for each pair of scenes by checking the timer shown at the beginning of the videos. Depending on the length of the original videos, we cropped the recordings to have lengths between 25 and 50 seconds.

\begin{figure}[h]
  \centering
  \includegraphics[width = 0.2\textwidth]{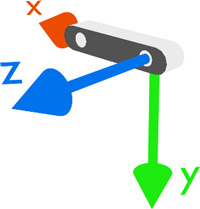}
  \caption{\textbf{Coordinate system of ZED}} 
  \label{fig:zedcoordinate}
\end{figure}

We also retrieved the ZED camera poses from the SVO file. The ZED uses a three-dimensional Cartesian coordinate system (X, Y, Z) with a right-handed, y-down configuration, and coordinates in millimeters. The motion of the ZED's left eye is described in the absolute real-world space, called the \textit{World Frame}, located where the ZED first started motion tracking; the very first frame of the video. (Fig~\ref{fig:zedcoordinate}). The pose data of each frame, including rotation and translation values, is stored in a text file.

\vspace{1em} \noindent
\textbf{Data format}
We describe the structure and format of the video dataset below. There are a total of 82 pairs of videos with different scenes. The total size of the dataset is over 1.2TB (around 700GB for extracted frames and 500GB for raw videos). In addition to the raw videos and their frames, we also provide pose data of the ZED's left lens for each scene.

\begin{multicols}{2}
\dirtree{%
.1 VIDEO\_DATA.
.2 Frames.
.3 set[2digits].
.4 canon\_frames.
.6 \textbf{[6digit].png}.
.4 p20\_frames.
.6 \textbf{[6digit].png}.
.4 zed\_frames.
.5 left.
.6 \textbf{[6digit].png}.
.5 depth.
.6 \textbf{[6digit].png}.
.5 \textbf{calib.txt}.
.5 \textbf{pose.txt}.
}
\columnbreak
\dirtree{%
.1 VIDEO\_DATA.
.2 RAW.
.3 \textbf{Canon\_[2digit].MOV}.
.3 \textbf{P20\_[2digit].mp4}.
.3 \textbf{ZED\_[2digit].svo}.
}
\end{multicols}

\subsection{Data Analysis}
We analyze the statistics of the collection time, color distribution, density of the depth map, and the number of object instances in PIV3CAMS. We expect this analysis to give readers an understanding of how the dataset is composed and provide ideas for future work with this dataset. 

\subsubsection{Collection time}
\begin{figure}[h!]
  \centering
  \includegraphics[width = 0.4\textwidth]{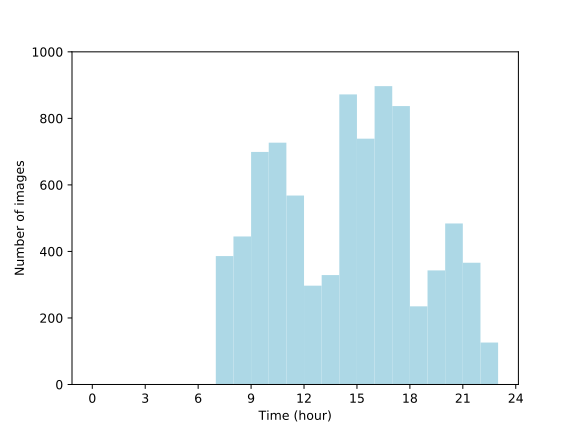}
  \includegraphics[width = 0.4\textwidth]{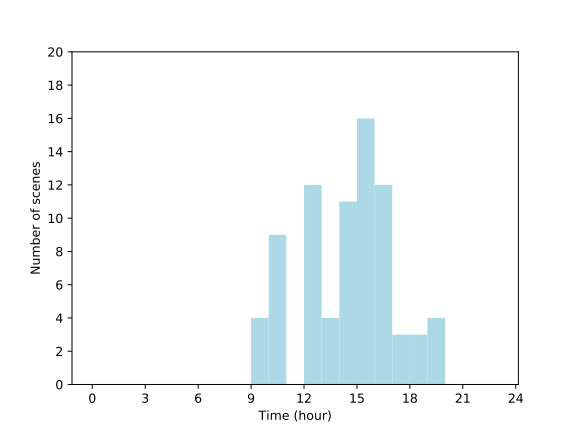}
  \caption{\textbf{Number of scenes taken at different times of the day}. Histogram of the scenes of images (left) and videos (right).}
  \label{fig:time}
\end{figure}

Our dataset was collected at different times of the day, as seen in Figure ~\ref{fig:time}. While images were taken throughout the day from 6 a.m. to 11 p.m., videos were mostly taken during the daytime. Considering that the collection took place during the summer, there are limited numbers of night scenes. Due to difficulties in capturing consistent quality RGB-D images from ZED cameras in low light, many images and videos captured at night were deleted during the pre-processing step.

\subsubsection{RAW image data}
The RAW images that we saved, along with other compressed images, are unprocessed outputs directly from the camera, showing more shades of colors and a better representation of image parameters such as white balance, contrast, exposure, gain, and more. Our RAW images, from the Canon and P20, are three to four times larger in size than the compressed images.

\begin{figure}[p]
  \centering
  \includegraphics[width = \textwidth]{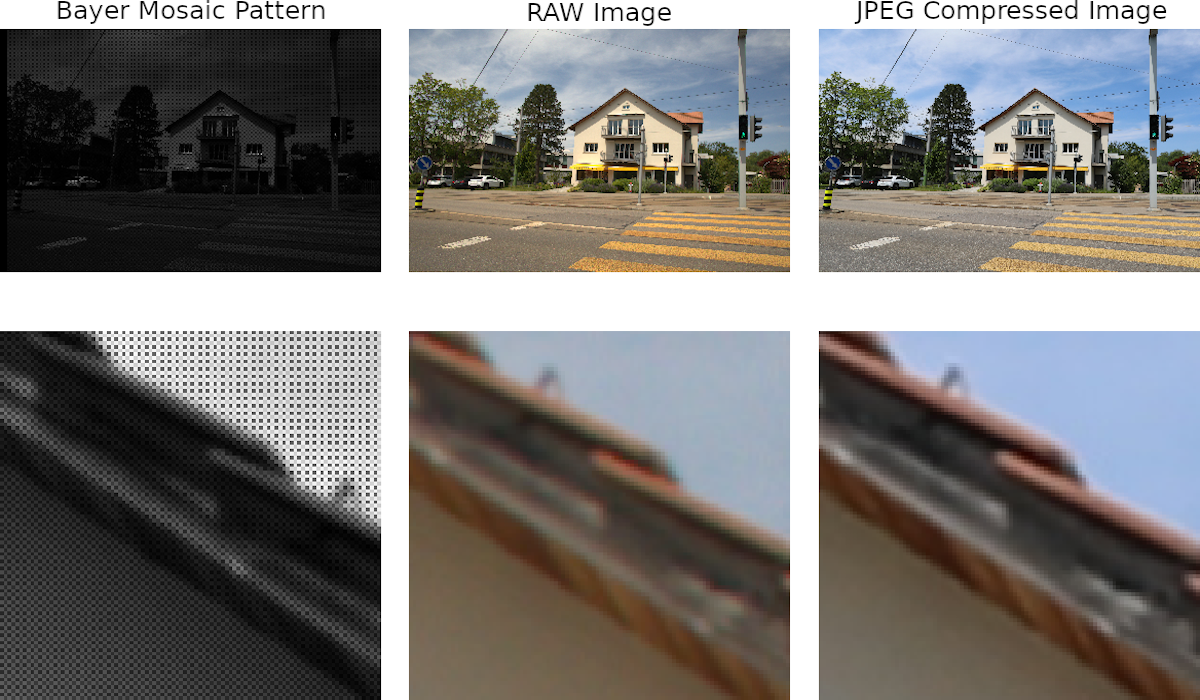}
  \caption{\textbf{Example of a RAW image from Canon}. Images in the left column show Bayer mosaic pattern of a RAW image. The middle column shows the RAW image and the right column shows the JPEG compressed image. }
  \label{fig:bayer}
  \vspace{0.5cm}
  \includegraphics[width = \textwidth]{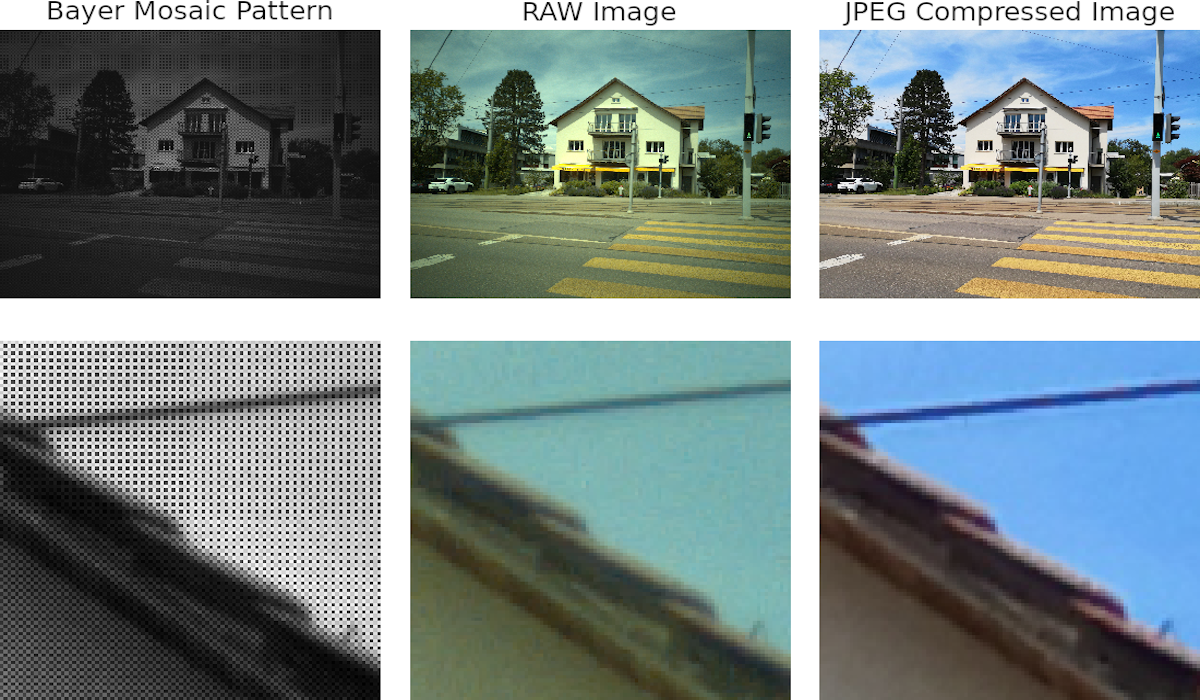}
  \caption{\textbf{Example of a RAW image from P20}. The left column shows the Bayer mosaic pattern of a RAW image, the middle column shows the RAW image, and the right column shows the JPEG compressed image. }
  \label{fig:bayer2}
\end{figure}

The pixels in RAW images are overlaid with the Bayer color filtering arrangement, also known as the Bayer mosaic pattern. It filters each pixel to record a single primary color (red, green, or blue) and its pattern consists of 50\% green, 25\% red, and 25\% blue. Various demosaicing algorithms can be used to interpolate RGB values in the Bayer mosaic pattern to obtain a full-color image.  Figures~\ref{fig:bayer} and ~\ref{fig:bayer2} show the RAW images and their Bayer mosaic pattern for the Canon and P20 images. Note that the ZED camera outputs processed images, so we do not save RAW images from the ZED.

As shown in Figures~\ref{fig:bayer} and ~\ref{fig:bayer2}, the RAW images are relatively darker at the edges. It is also possible to compare how much the image is processed by comparing the RAW image and a JPEG compressed image. For the P20, the JPEG compressed image has visually good quality compared to its RAW image. 

\subsubsection{Object instance analysis}
Although we have not annotated our scenes with objects, we analyzed the PIV3CAMS dataset using an object detector to examine the diversity of the scenes. We used the \textit{TensorFlow Object Detection API}\cite{objectAPI} provided by Google, which includes popular object detection networks with pre-trained models. We used the Faster R-CNN detector\cite{frcnn} with models trained on the Open Images Dataset V4\cite{OpenImages}, which has 600 classes of object instances.

\begin{figure}[h]
  \centering
  \includegraphics[width =\textwidth]{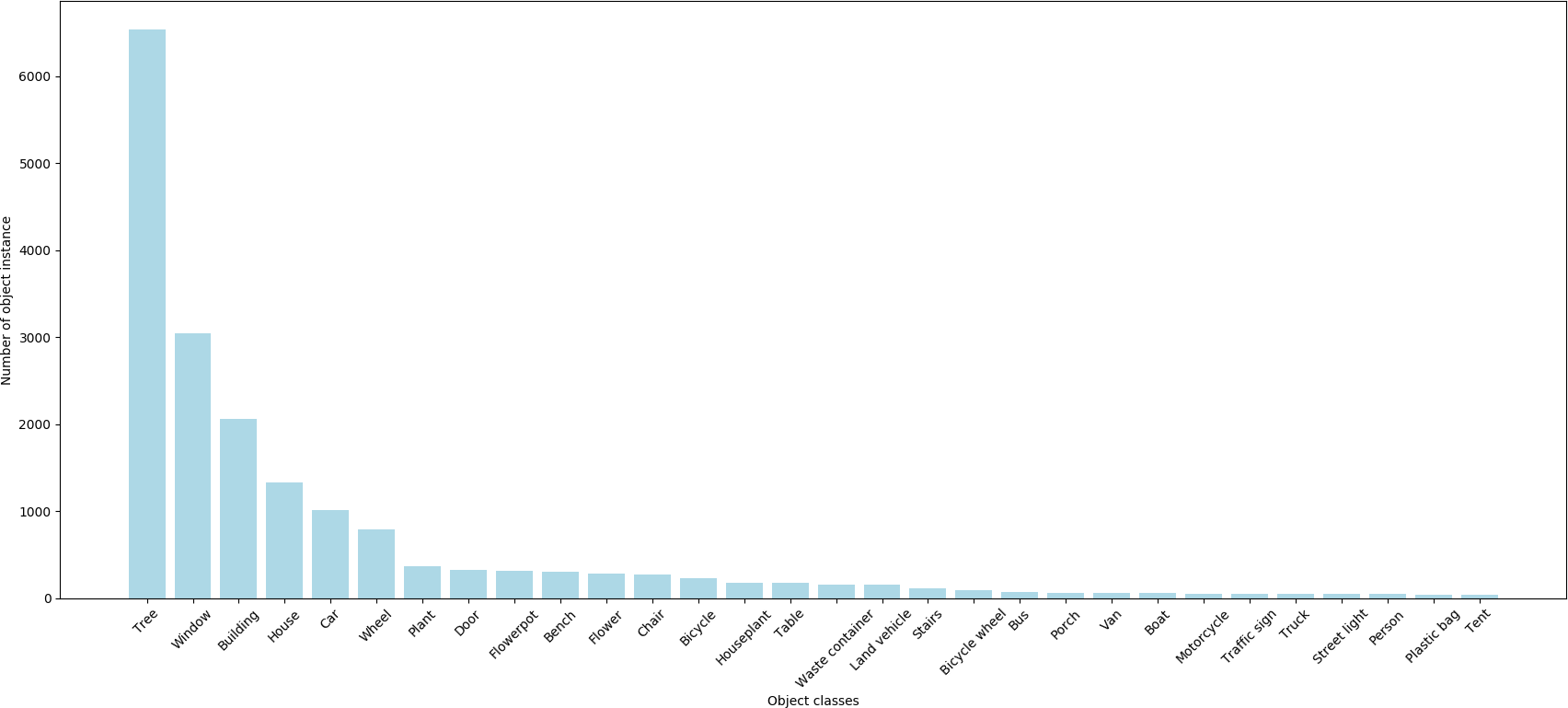}
  \caption{\textbf{Top 30 most frequent objects found in the image dataset}. The histogram shows 30 most frequent objects found throughout the image dataset.} 
  \label{fig:object_img}
\end{figure}

For the image dataset, we used all 8,385 images to test the object detector. On average, 3.656 object instances were detected per scene. Figure~\ref{fig:object_img} shows the top 30 most frequently seen object instance  in the image dataset. From this figure, we notice that \texttt{tree} appears in almost every scene, likely because many greens are planted throughout the cities. The \texttt{window}, \texttt{building}, \texttt{house}, and \texttt{car} are the next most frequent objects. Since most of the image collection locations were within urban areas, it was inevitable to capture houses and cars.

\begin{figure}[h]
  \centering
  \includegraphics[width = \textwidth]{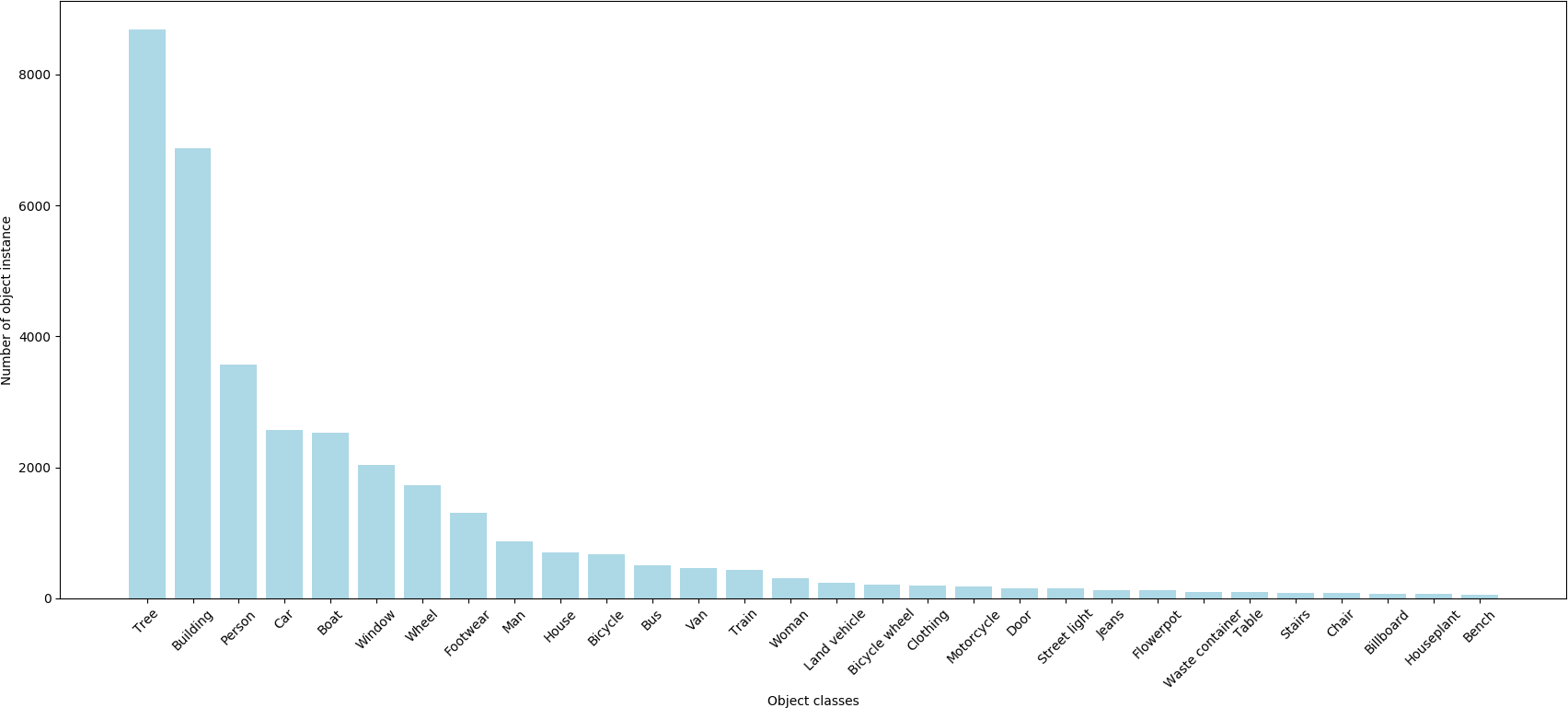}
  \caption{\textbf{Top 30 most frequent objects found in the video dataset}. The histogram shows 30 most frequent objects found throughout the video dataset. } 
  \label{fig:object_vid}
\end{figure}

For the video dataset, we sampled video frames every ten frames to avoid excessive overlap between frames. On average, 5.019 object instances were detected per scene. Figure \ref{fig:object_vid} shows the top 30 most frequently seen object instances in the video dataset. Similar to the image dataset, \texttt{tree} and \texttt{building} are the two most frequently appearing objects. While we focused on collecting static scenes for the image sets, we captured more dynamic scenes for the video sets. Therefore, there are more active object instances, such as \texttt{person}, \texttt{car}, and \texttt{boat}, appearing in the videos compared to the images.

\section{Novel View Synthesis using RGB-D images } \label{ch4}
Along with data collection, the second aim of this thesis is to explore a computer vision task: novel view synthesis. As discussed in Chapter 2, there are two main approaches to the novel view synthesis problem: geometry-based and learning-based approaches. The method we chose to investigate is \cite{nvsmachines}, which is the state-of-the-art method at the time of writing this thesis. Their method involves training an encoder-decoder network given a pair of RGB images of source and target views and their poses. Our goal was first to reproduce their method as our baseline. Then, we modified parts of their network to apply depth images and investigate the effectiveness of using depth information in the novel view synthesis task.

In this chapter, we will briefly explain the baseline method and how we re-implemented their models. Then we will introduce several network variations that we modified from the baseline network. The experimental results on each network will be discussed in the next chapter.

\subsection{Regenerated Baseline Network}
\begin{figure}[h]
  \centering
  \includegraphics[width = 0.8\textwidth]{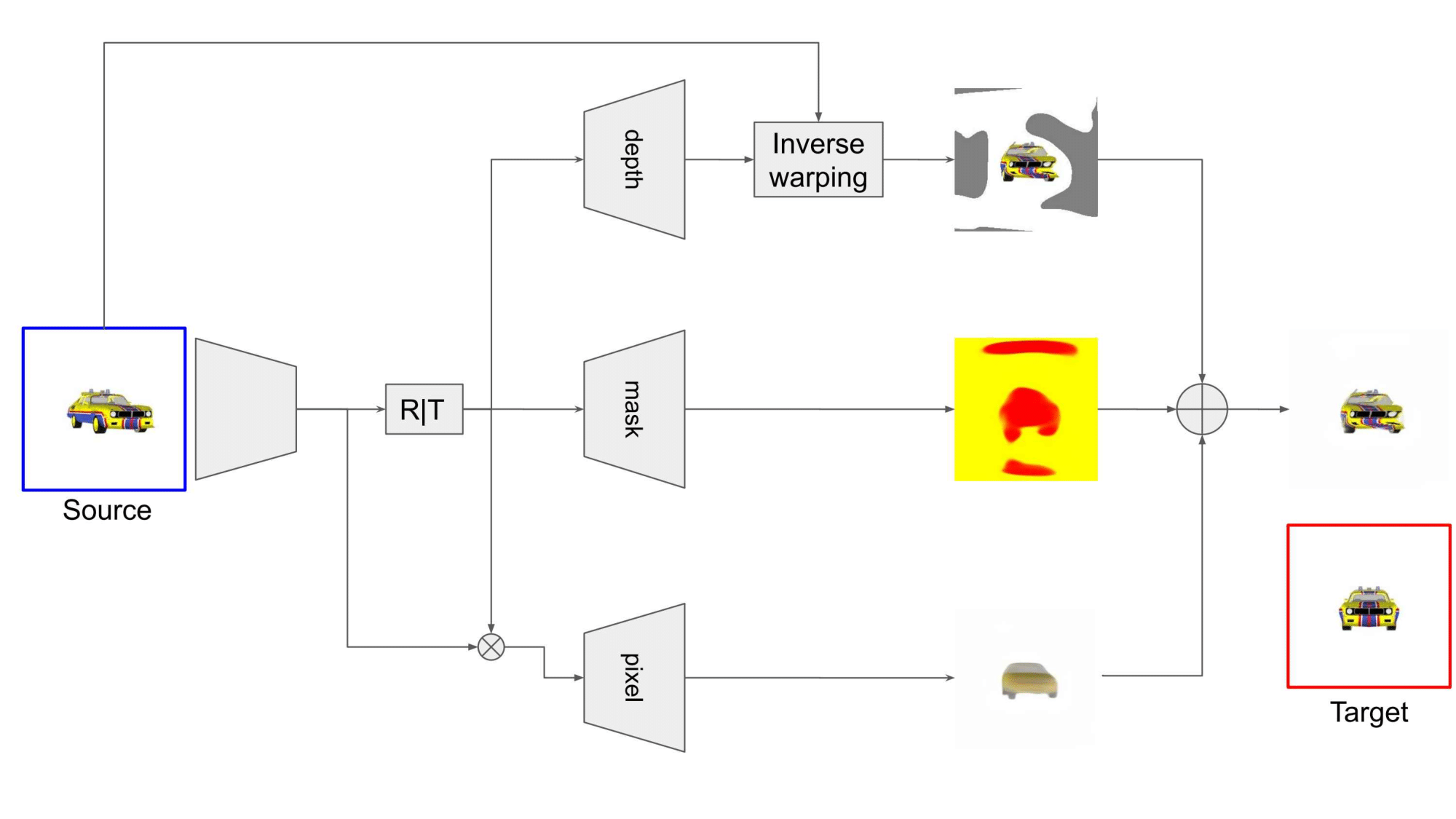}
  \caption{\textbf{Baseline Network Architecture} The network consists of an encoder and three decoders. The depth decoder predicts the depth map, the pixel decoder predicts the RGB image, and the mask decoder merges their outputs. Given a source view and a transformation matrix, the network outputs the target view.} 
  \label{fig:baseline_model}
\end{figure}

The current state-of-the-art learning-based novel view synthesis method has been proposed by \cite{nvsmachines}. The main idea of their work is to train a network with different branches that learn the depth and the pixels of the target view. The network has an encoder-decoder structure with three different branches that learn the depth, mask, and pixels of the target view (Fig~\ref{fig:baseline_model}).

Inspired by \cite{TAE}, they transformed the latent vector, an output of the encoder network, according to the given pose matrix between the source and target views. This transformed latent vector is then given as an input to three different decoders. The depth decoder predicts the depth map of the target view and uses it to inverse warp the source view to the target view. The pixel decoder directly predicts the RGB images of the target view. Finally, the mask decoder learns to create a mask that merges the inverse warped image from the depth decoder branch and the result of the pixel decoder.

\subsubsection{Encoder and Decoder modules}
\begin{figure}[h]
  \centering
  \includegraphics[width = 0.45\textwidth]{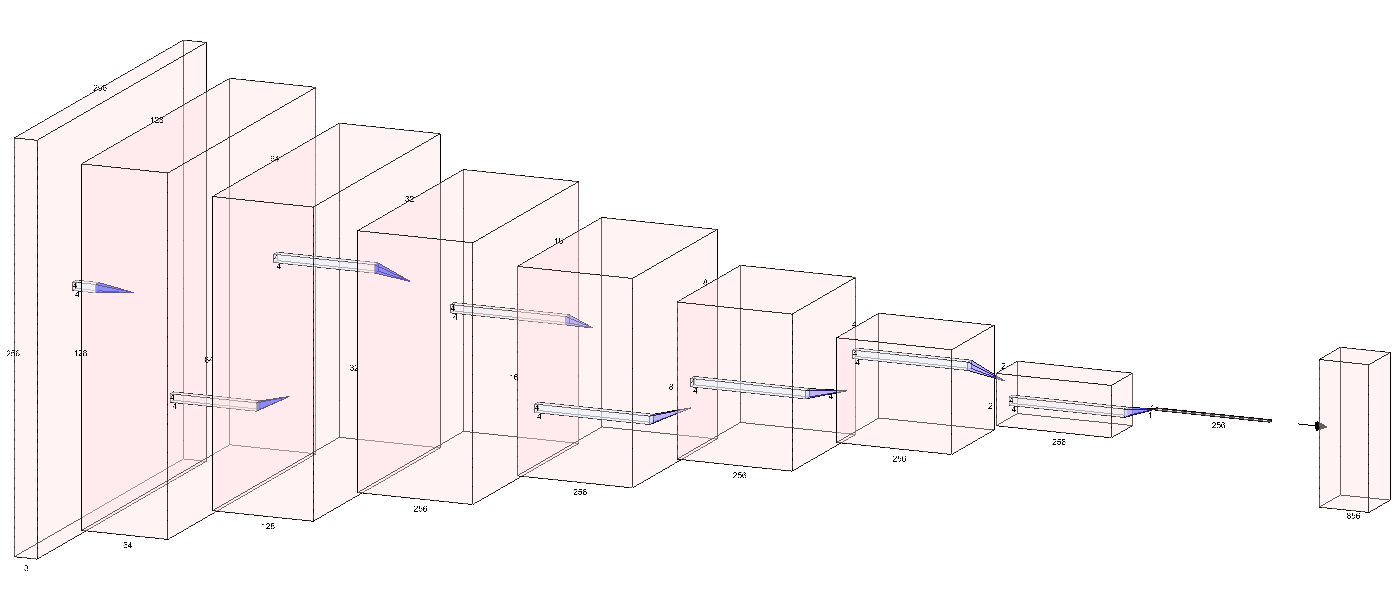}
  \includegraphics[width = 0.45\textwidth]{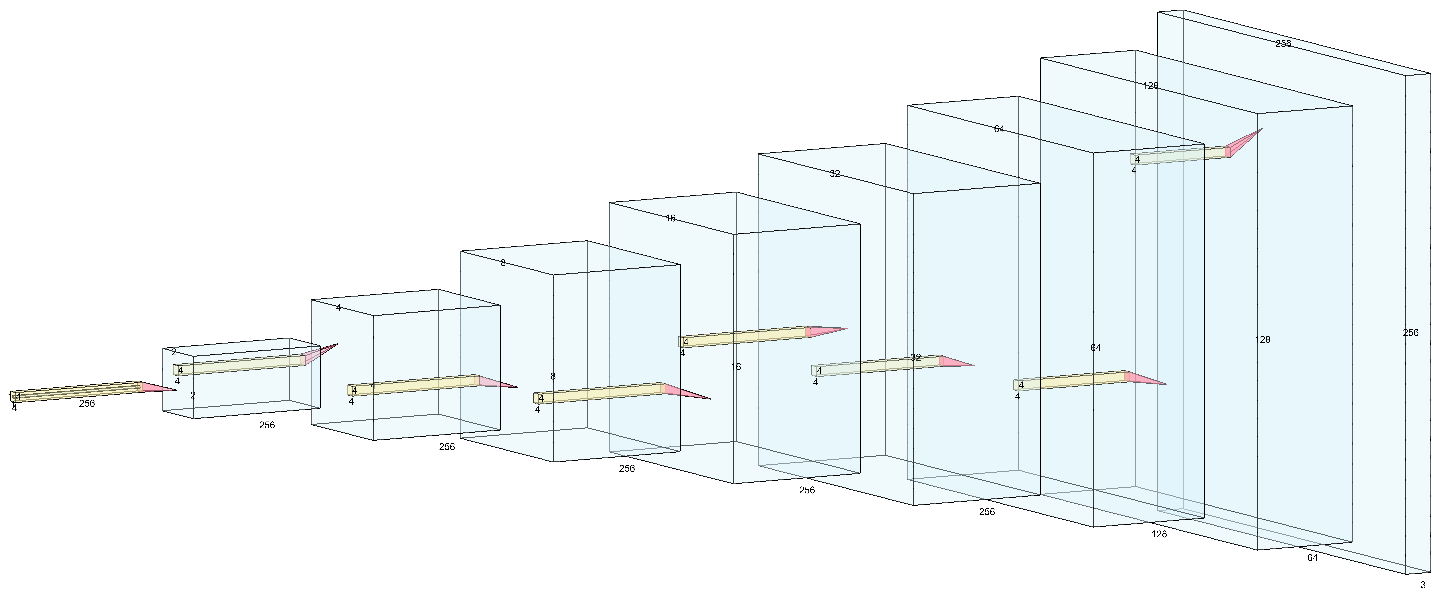}
  \caption{\textbf{Encoder and Decoder module diagram}. The encoder (top) and decoder (bottom) modules each consist of 7 convolution layers with batch normalization and leaky ReLU activation functions between each layer.} 
  \label{fig:network}
\end{figure}

The authors of \cite{nvsmachines} did not release their source code or supplementary materials at the time of this project. Therefore, we relied solely on the network description from their paper and designed the encoder and decoder modules as shown in Figure~\ref{fig:network}.

We constructed the encoder module with 7 convolution layers (stride of 2, kernel size of 4, and padding size of 1) with batch normalization and the leaky rectified linear unit (leaky ReLU) as the activation function. At the end of the encoder module, a fully connected layer outputs the latent vector with the chosen size.

The decoder module also consists of 7 deconvolution layers (stride of 2, kernel size of 4, and padding size of 1). Similar to the encoder module, we added batch normalization between each convolution layer and used the leaky ReLU as the activation function. A fully connected layer is attached at the beginning of each decoder to receive the rotated latent vector as input. Note that the output of each decoder varies by its usage. For example, the pixel branch decoder produces an RGB image of size \(Batch Size \times W \times H \times 3\) while the depth and mask branch decoders produce a depth map and a mask of size \(Batch Size \times W \times H \times 1\).  

\subsubsection{Transforming latent vector}
\cite{nvsmachines} deployed a Transforming Auto-Encoder (T-AE) network\cite{TAE} by applying a 3D transform to the latent vector \(z^T = T_{src \rightarrow tgt }(z)\).  The latent vector z is represented as \(z \in R^{n\times 3}\), where n is a hyper-parameter. This representation is then multiplied with the ground-truth transformation \(T_{src \rightarrow tgt }\) which describes the camera pose between the source view (\(I_{src}\)) and the target view (\(I_{tgt}\)). This 3D transformed latent vector is given as input to the decoders in each branch.

\subsubsection{Depth branch}
The decoder in the depth branch decodes \(z^T\) into a target depth mask \(D^{pred}_{tgt}\). We compute the warped target view \(I^{warped}_{tgt}\) deterministically via perspective projection using \(D^{pred}_{tgt}\),and \(I_{src}\). 

Each pixel \((x_{t}, y_{t})\) in the depth mask is mapped to 3D coordinates [X, Y, Z] using the intrinsic parameter matrix K.
\begin{equation}\label{eq:1}
    [X,Y,Z]^{T} = K^{-1} D^{pred}_{tgt}(x_{t}, y_{t}) [x_{t}, y_{t}, 1]^{T}
\end{equation}
We compute the new 3D coordinates of the pixel's location in the source view by multiplying transformation matrix \(T_{tgt \rightarrow src}\) which describes the camera rotation from the target view to the source view. These coordinates are then projected onto the source image plane.
\begin{equation} \label{eq:2}
    [x^{tgt}_{s}, y^{tgt}_{s}, 1]^{T} = K T_{tgt \rightarrow src } [X,Y,Z,1]^{T}
\end{equation}

Through the above equation, we obtain the correspondence between the pixels in the target view and their positions in the source view. Note that the pixel positions derived from equation~\ref{eq:2} are non-integer.  Therefore, we used a differentiable bilinear sampling method proposed in \cite{spatialtransformNetwork} to compute the pixel values in the target view from the source view.
\begin{equation} \label{eq:3}
    I^{warped}_{tgt}(x_{t}, y_{t}) = \sum_{x_{s}}\sum_{y_{s}} I_{src} max(0, 1-|x_{s} - x^{tgt}_{s}|) max(0, 1-|y_{s} - y^{tgt}_{s}|) 
\end{equation}

Finally, we warp the source view to the target view using the pixel correspondences. The operations, from equation~\ref{eq:1} to equation~\ref{eq:3}, are indicated as an \textit{inverse warping} block in Figure~\ref{fig:baseline_model}. This operation is crucial for generating target views as it helps preserve the texture and local details of the source view.

\subsubsection{Pixel branch and Mask branch}
The decoder in the pixel branch decodes \(z^T\) into a target RGB image \(I^{pixel}_{tgt}\). The decoder in the mask branch decodes \(z^T\) into a weighted mask \(M^{pred}\) which is then used to fuse the results from the pixel and depth branch. 

\subsubsection{Fusion}
While the predicted target view from the depth branch contains textures and local details close to the ground-truth target view, it is difficult to reconstruct occluded parts in the source view. On the other hand, the predicted target view from the pixel branch has a similar structure to the ground-truth target view but lacks local details. Therefore, \cite{nvsmachines} employed a fusion of the predicted target views from the depth and pixel branches using the mask. The final predicted target view \(I^{pred}_{tgt}\) is given by equation~\ref{eq:4} where \(\cdot\) denotes element-wise multiplication.
\begin{equation}\label{eq:4}
    I^{pred}_{tgt} = (1 - M^{pred}) \cdot I^{warped}_{tgt}  + M^{pred} \cdot I^{pixel}_{tgt} 
\end{equation}

\subsubsection{Training}
The baseline model requires only pairs of source and target views and their transformation information to train the networks. We applied the L1 loss between the predicted image outputs from each branch and the ground truth. All three networks are trained by minimizing the reconstruction loss, as shown in equation~\ref{eq:5}.
\begin{equation}\label{eq:5}
    loss_{recon} = || I^{pred}_{tgt} - I_{tgt}||_{1}  +|| I^{warped}_{tgt} - I_{tgt}||_{1} + || I^{pixel}_{tgt} - I_{tgt}||_{1}
\end{equation}

Additionally, we applied a least-square GAN loss\cite{lsgan} and a perceptual loss\cite{perceptualloss} to the output of the pixel branch to enhance realism.
\begin{equation}\label{eq:6}
    loss_{LSGAN} = (1 - Dis(I^{pixel}_{tgt}))^{2}
\end{equation}
\begin{equation}\label{eq:7}
     loss_{VGG} = || F_{vgg16}(I_{tgt}) - F_{vgg16}(I^{pixel}_{tgt} )||_{2}
\end{equation}
where \(Dis\) is a discriminator and \(F_{vgg16}\) is a feature-extracting function base on a pre-trained VGG network\cite{vgg16}. 

The final loss is defined as:
\begin{equation} \label{eq:8}
    loss = \lambda_{L1} loss_{recon}  + \lambda_{GAN} loss_{LSGAN} + \lambda_{VGG} loss_{VGG} 
\end{equation}
where \(\lambda_{L1}, \lambda_{GAN}\), and \(\lambda_{VGG} \) are the weight parameters.

\subsection{Variation Networks}
Our motivation is to investigate the effectiveness of using ground-truth depth maps instead of predicting depth masks from the network. To experiment with our ideas, we made several variations to the baseline network by replacing the depth decoder with ground-truth depth maps. Additionally, we examined the importance of the mask, which plays a significant role in fusing the results of the depth and pixel branches. Therefore, we altered the depth and mask branches of the baseline network while keeping other parts unchanged.

In summary, we classified our proposed networks into two main categories. Within each category, we made small changes, such as using different types of input depth maps (source or target depth map) and masks for final prediction. We explain the details of the modified networks in the following sub-sections.

\begin{itemize}
    \item[(1)] Without depth decoder. 
    \begin{itemize}
        \item Using target depth maps. 
        \item Using source depth maps. 
    \end{itemize}
    \item[(2)] Without both depth decoder and mask decoder.
    \begin{itemize}
        \item Using target depth maps and using no mask.
        \item Using target depth maps and using visible source pixels in the target view as a mask. 
        \item Using source depth maps and using visible source pixels in the target view as a mask.
    \end{itemize}
\end{itemize}

\subsubsection{Variation 1: Without depth decoder}
\begin{figure}[h!]
  \centering
  \includegraphics[width = \textwidth]{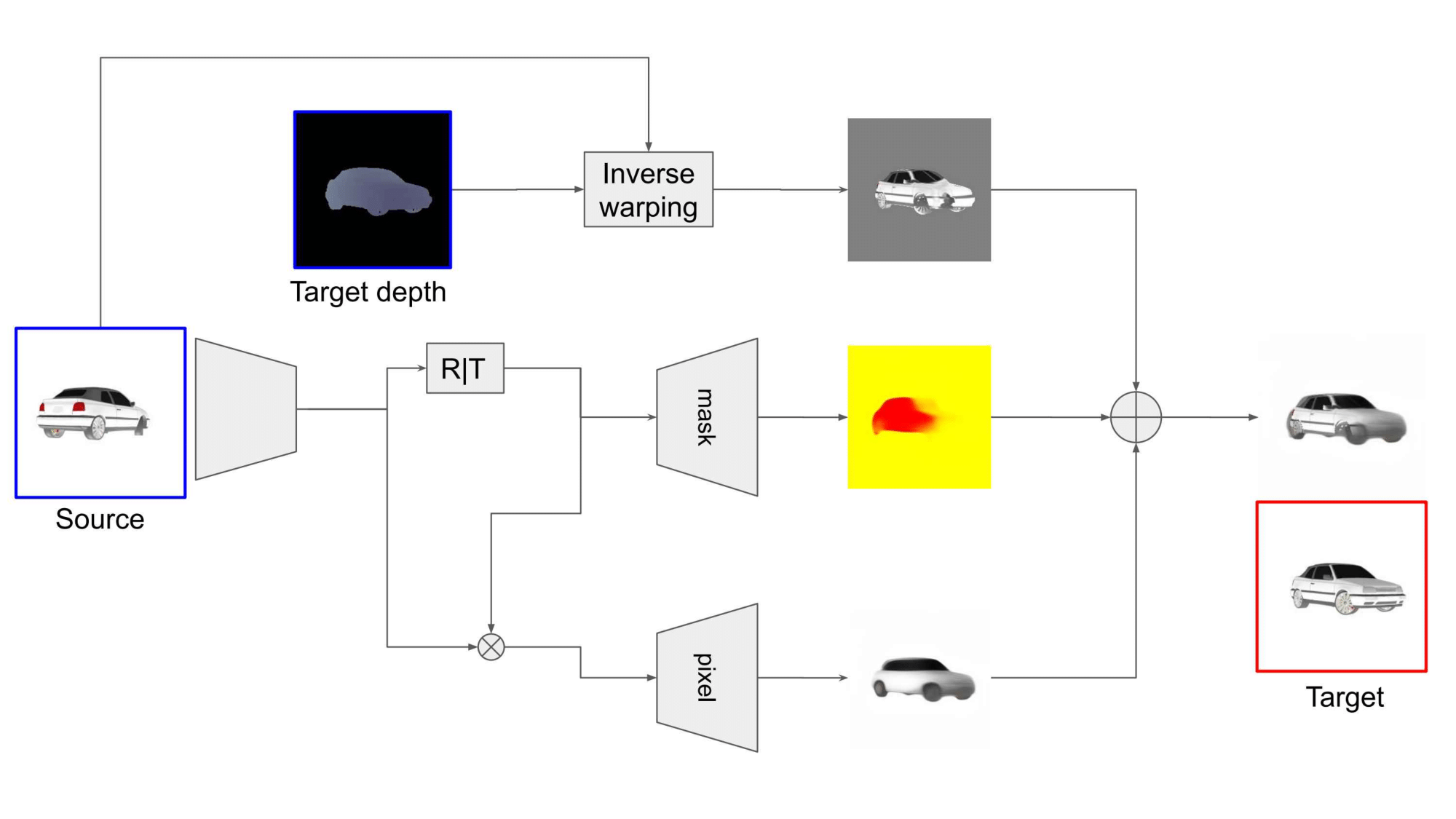}
  \caption{\textbf{ND-Tgt Network model}. This variation replaces the depth decoder with ground-truth target depth. The encoder and two decoders (pixel and mask) predict the target RGB image and a fusion mask. Given a source view, ground-truth target depth, and a pose matrix, the network outputs the predicted target view. } 
  \label{fig:var1_tgt_model}
\end{figure}

\textbf{Using target depth maps}
\newline
To investigate the effectiveness of using ground-truth depth maps for novel view synthesis, our first approach is to replace the depth decoder with a ground-truth target depth map. Since the role of the depth decoder in the baseline model is to predict the target depth mask for inverse warping, we have instead used the ground-truth target depth directly, as shown in Figure \ref{fig:var1_tgt_model}. We will refer to this variation model as \textit{No Depth Decoder - Target Depth Map (ND-Tgt)} for simplicity.

The encoder and decoder modules are the same as in the baseline network. Additionally, the transformation of the latent vector \(z\) and the inverse warping of the target depth maps to produce warped target images are done as explained earlier. Note that there are no weights to train in the depth branch of the network. The final loss is defined as in equation \ref{eq:8}, with \(loss_{LSGAN}\) and \(loss_{VGG} \) as defined in equations \ref{eq:6} and \ref{eq:7}, respectively.
\begin{equation} \label{eq:8}
    loss = \lambda_{L1}|| I^{pred}_{tgt} - I_{tgt}||_{1}  + \lambda_{L1}|| I^{pixel}_{tgt} - I_{tgt}||_{1} + \lambda_{GAN} loss_{LSGAN} + \lambda_{VGG} loss_{VGG} 
\end{equation}

\begin{figure}[h!]
  \centering
  \includegraphics[width = \textwidth]{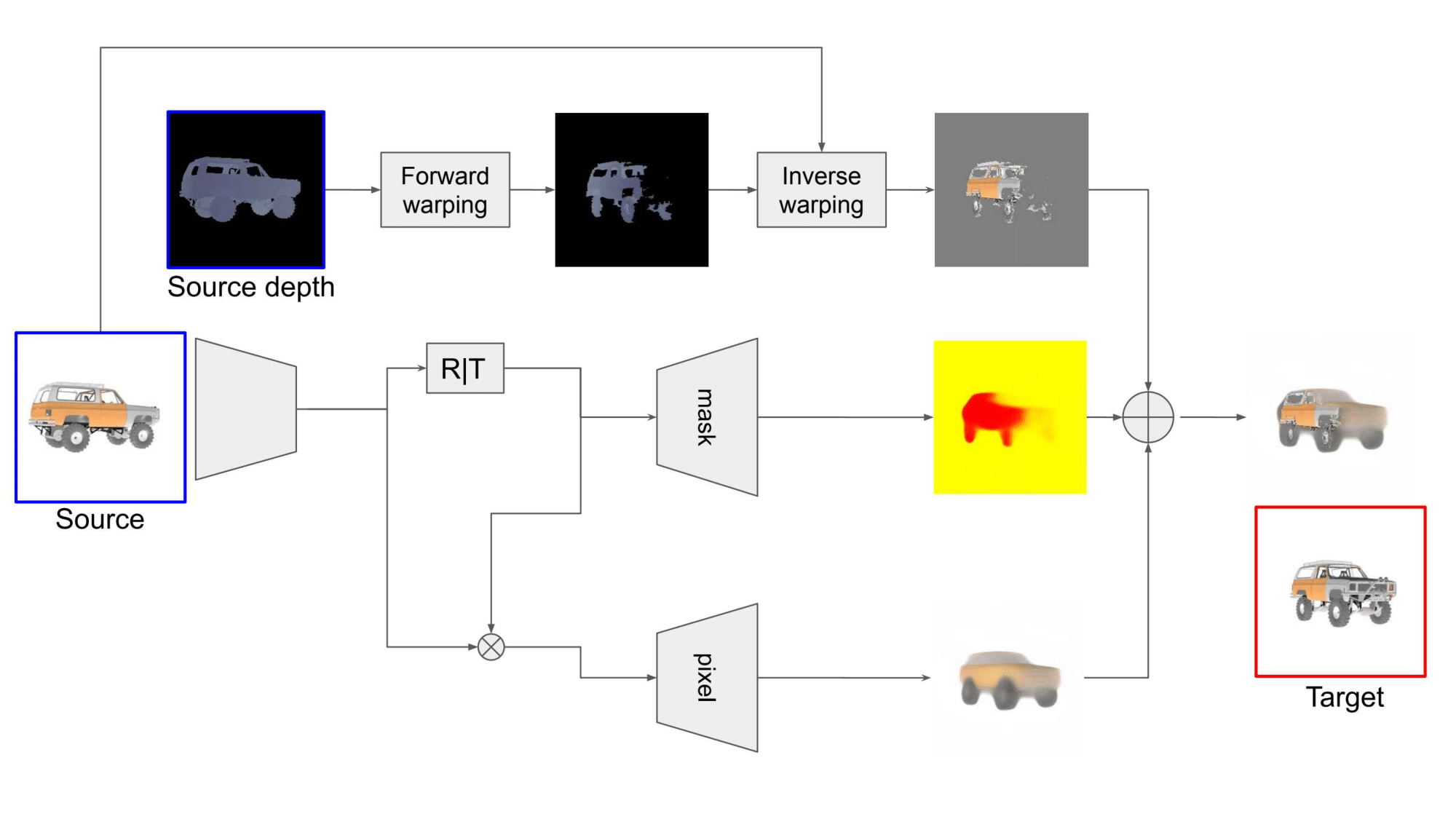}
  \caption{\textbf{ND-Src Network model}. This variation replaces the depth decoder with the ground-truth source depth. The encoder and two decoders (pixel and mask) predict the target RGB image and a fusion mask. Given a source view, ground-truth source depth, and a pose matrix, the network outputs the predicted target view.} 
  \label{fig:var1_src_model}
\end{figure}

\noindent
\textbf{Using source depth maps}
\newline
Normally, RGB-D images come as a pair of source RGB image and source depth map in real scenarios. Therefore, we modified the \textit{ND-Tgt} network to use the source depth map as an input, as shown in Figure Figure~\ref{fig:var1_src_model}. We will refer to this variation model as \textit{No Depth Decoder - Source Depth Map (ND-Src)} for simplicity.

\begin{figure}[h]
  \centering
  \includegraphics[width = \textwidth]{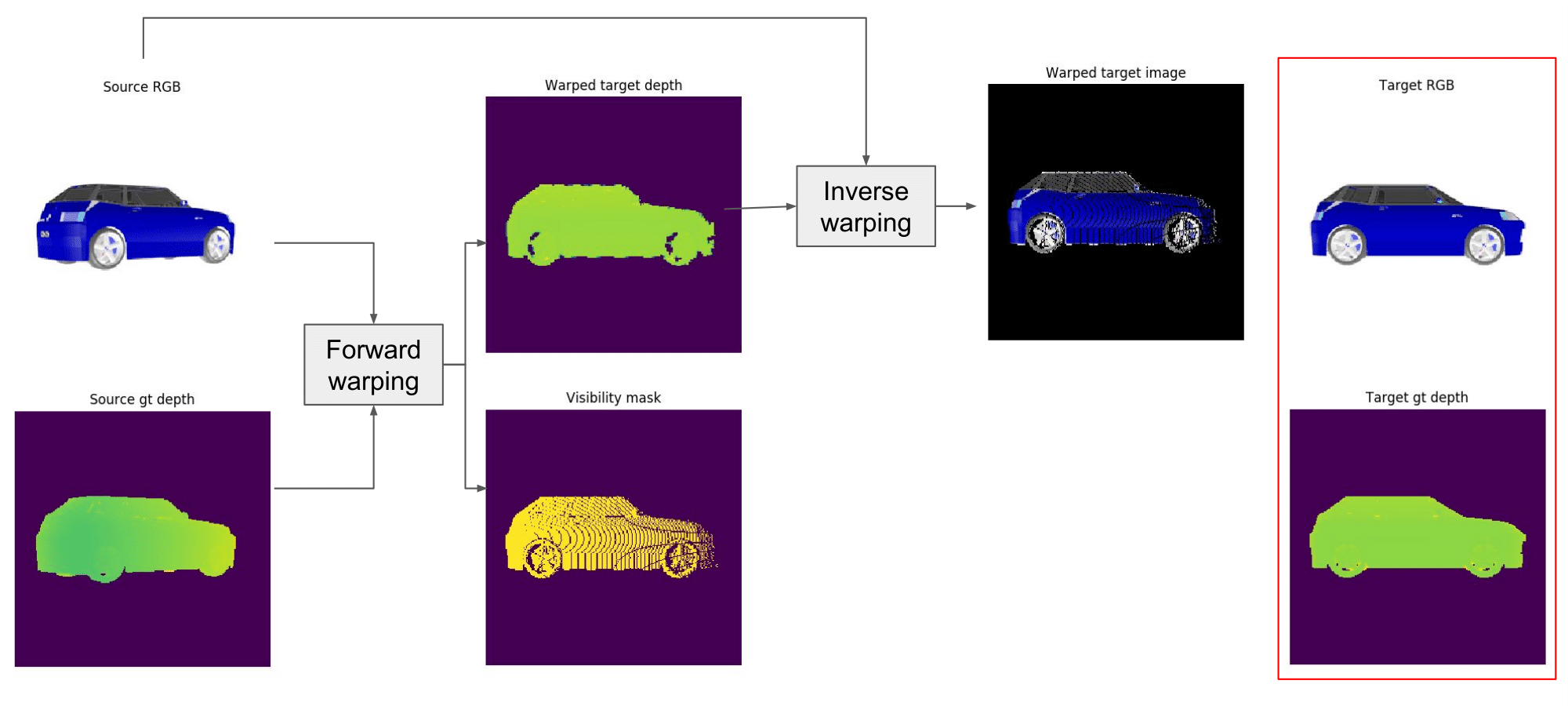}
  \caption{\textbf{Example of depth branch using source RGB-D image}. Given a source RGB-D image as an input, we compute the warped target depth map and visibility mask via forward-warping. Then, the source image and warped target depth map are used to compute the warped target image via inverse warping.} 
  \label{fig:forward_warping}
\end{figure}

From the source depth map \(D^{gt}_{src}\), we computed the warped target depth map \(D^{warped}_{tgt}\) using the forward projective projection. This warped target depth \(D^{warped}_{tgt}\) is used to compute the warped target image \(I^{warped}_{tgt}\). The forward warping process is similar to the inverse warping process explained previously, except for the transformation matrix  \(T_{src \rightarrow tgt}\),which describes the camera rotation from the source view to the target view. Using forward warping, we can also compute a visibility mask that shows pixels visible in the target view. Figure~\ref{fig:forward_warping} shows the depth branch using the source RGB-D image to compute the warped target depth and visibility mask. This warped target depth is then used with the source view to generate the target view.

For the mask and pixel branches, we used the same encoder and decoder as the baseline network. The latent  vector \(z\) is transformed and the inverse warping of the target are done as well. Since there are no weights for thedepth branch, the final loss is defined as equation~\ref{eq:8}

\subsubsection{Variation 2: Without both depth decoder and mask decoder}
Although the role of the depth branch is to provide more accurate texture and local details to the predicted target view, this information is only available for the parts that are not occluded from the source view. Therefore, we were motivated to use the visibility map instead of learning the mask from the decoder.

\begin{figure}[h!]
  \centering
  \includegraphics[width = 0.8\textwidth]{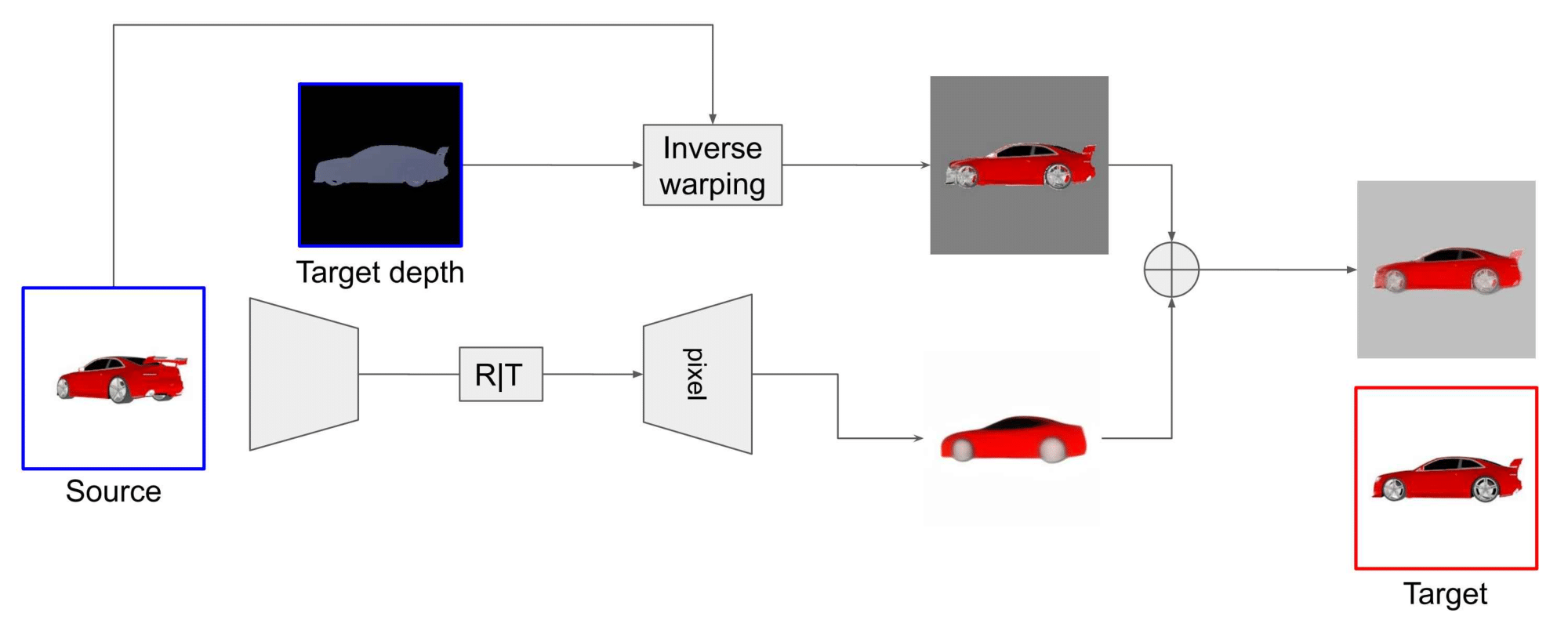}
  \caption{\textbf{NDNM-Tgt Network model}. This variation uses an encoder and a pixel decoder, replacing the depth decoder with ground-truth target depth. The pixel decoder predicts the target RGB image. Fusion is done by averaging pixel values from the warped target view and the pixel branch output. Given a source view, ground-truth target depth, and a pose matrix, the network outputs the predicted target view.} 
  \label{fig:var2_tgt_model}
\end{figure}

\vspace{1em} \noindent
\textbf{Using target depth maps and using no mask}
\newline
First, we modified the baseline network by removing both the depth and mask decoders, as shown in Figure~\ref{fig:var2_tgt_model}. We will refer to this variation model as \textit{No Depth Decoder, No Mask Decoder - Target Depth Map (NDNM-Tgt)} for simplicity.

The final predicted target image \(I^{pred}_{tgt} \) is achieved by averaging the result from the depth branch and the pixel branch. 
\begin{equation} \label{eq:10}
    I^{pred}_{tgt} = \frac{1}{2} I^{warped}_{tgt}  + \frac{1}{2} I^{pixel}_{tgt} 
\end{equation}

Since the NDNM-Tgt network consists of only one encoder and one decoder, the total loss is solely on the pixel branch, as shown in equeation~\ref{eq:11}
\begin{equation} \label{eq:11}
    loss = \lambda_{L1}|| I^{pred}_{tgt} - I_{tgt}||_{1}  + \lambda_{L1}|| I^{pixel}_{tgt} - I_{tgt}||_{1} + \lambda_{GAN} loss_{LSGAN} + \lambda_{VGG} loss_{VGG} 
\end{equation}

\noindent
\textbf{Using target depth maps and visible source pixels in the target view as a mask}
\newline
Next, we computed the visibility mask \(M^{vis}\) from the source RGB-D image using forward-warping projection. This mask is then used to merge the results from the depth and pixel branches (Eq.~\ref{eq:12}). The visibility mask is originally sparse, as shown in Figure~\ref{fig:forward_warping}. We filled its gaps using the closing operation of mathematical morphology to create a dense visibility mask. For our experiment, we tried using both sparse and dense visibility masks.
\begin{equation} \label{eq:12}
    I^{pred}_{tgt} = M^{vis} \cdot I^{warped}_{tgt}  + (1-M^{vis}) \cdot I^{pixel}_{tgt} 
\end{equation}

\begin{figure}[h!]
  \centering
  \includegraphics[width = 0.9\textwidth]{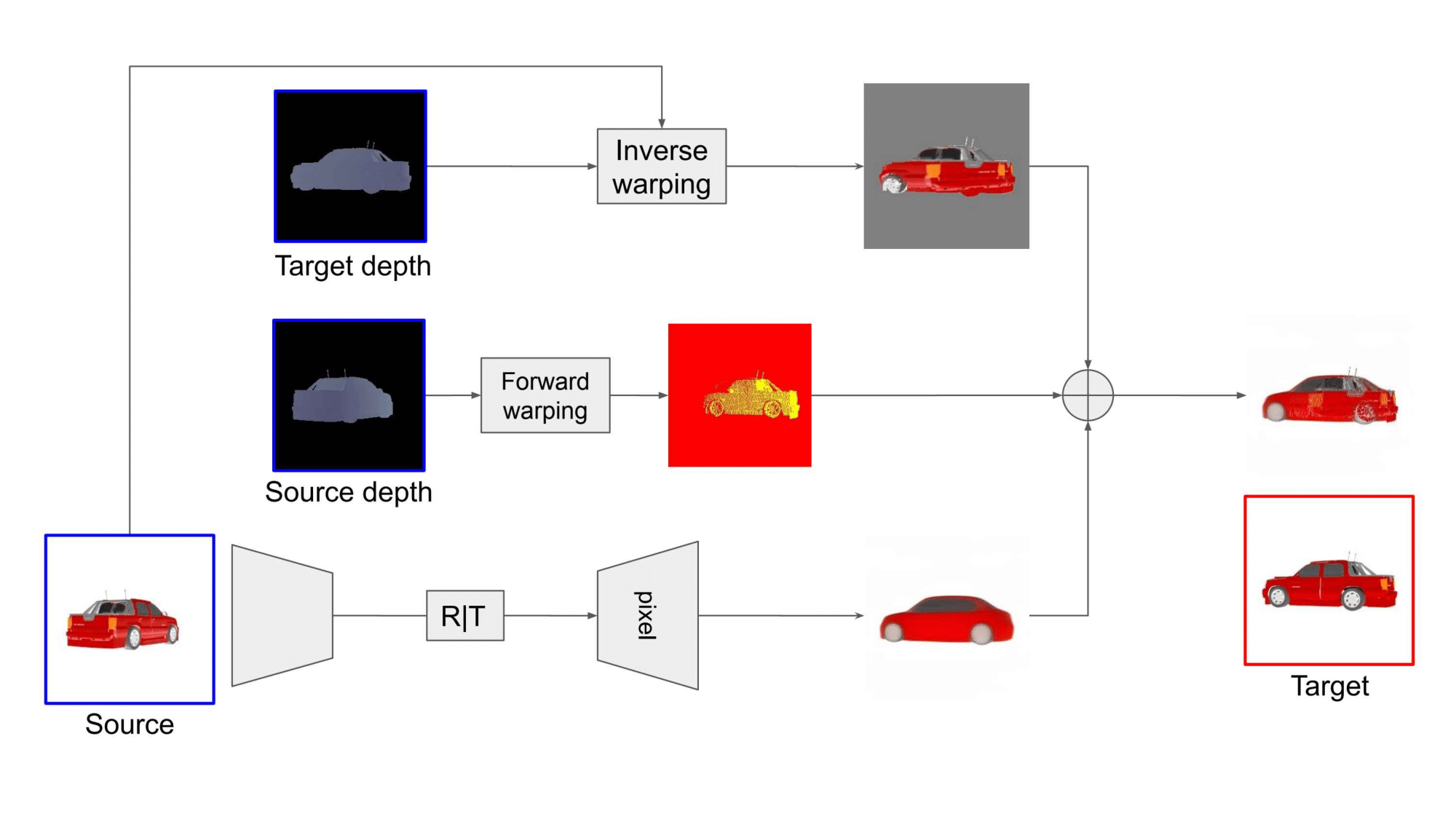}
  \caption{\textbf{NDVM-Tgt Network model}.This variation uses an encoder and a pixel decoder, replacing the depth decoder with ground-truth target depth. The pixel decoder predicts the target RGB image. Fusion between depth and pixel branches is done using the visibility mask from forward-warping projection.} 
  \label{fig:var2_vm_tgt_model}
\end{figure}

We will refer to this variation model as \textit{No Depth Decoder, Visibility Mask - Target depth map (NDVM-Tgt)} for simplicity. Figure~\ref{fig:var2_vm_tgt_model} shows the diagram of this model. Note that this model uses the same method for transforming the latent vector \(z\). Additionally, the pixel branch is trained with GAN loss~\ref{eq:6} and VGG loss~\ref{eq:7} for realistic results. Therefore, the final loss is defined the same as in equation~\ref{eq:11}.  

\begin{figure}[h]
  \centering
  \includegraphics[width = \textwidth]{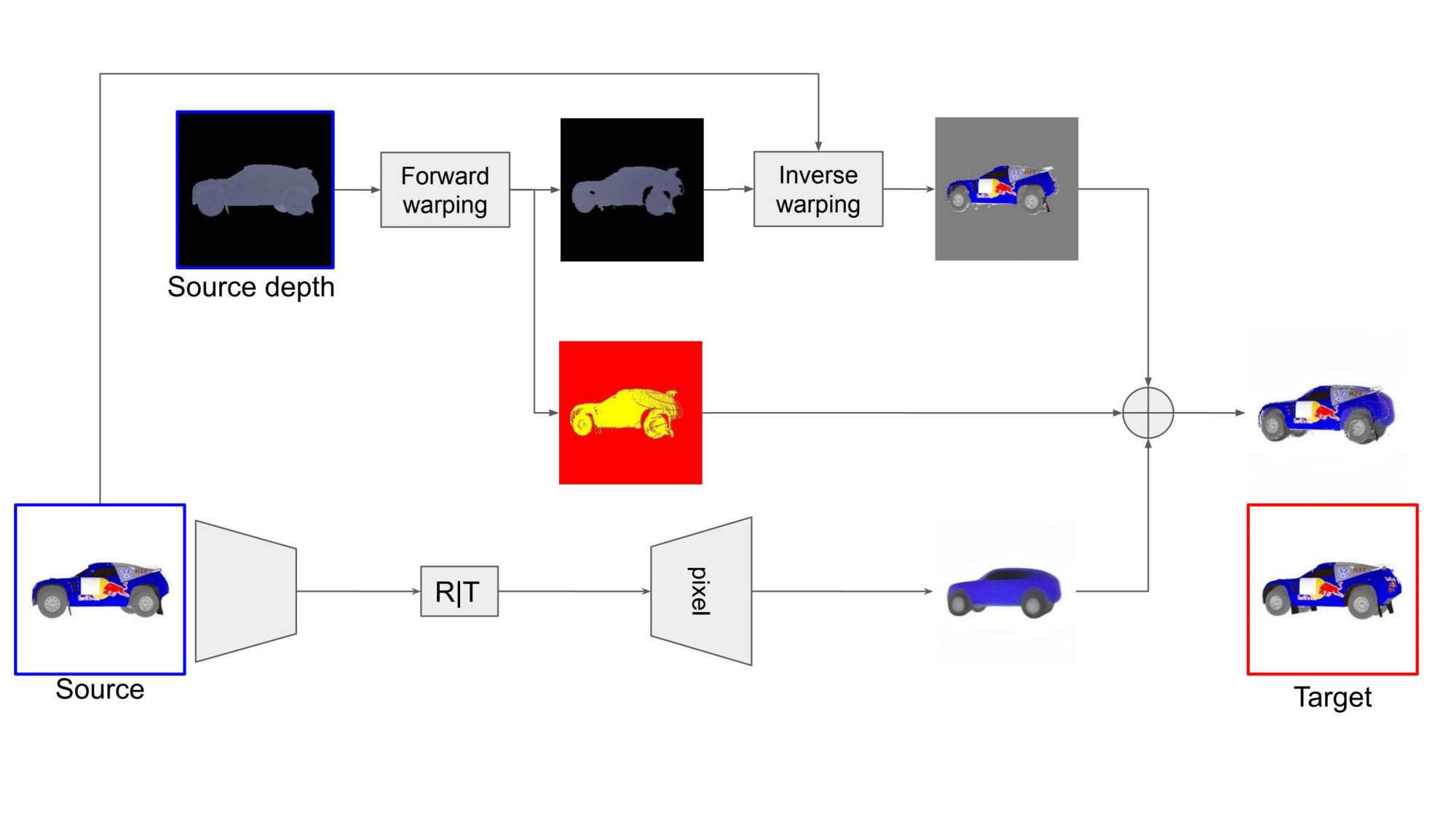}
  \caption{\textbf{NDVM-Src Network model}. This variation uses an encoder and a pixel decoder. The source RGB-D image is used to compute the warped target depth map and visibility mask, which in turn predict the warped target image. The pixel decoder predicts the target RGB image. Fusion between the depth and pixel branches is done using the visibility mask from forward-warping projection. Given a source RGB-D and a pose matrix, the network outputs the predicted target view.} 
  \label{fig:var2_vm_src_model}
\end{figure}

\vspace{1em}\noindent
\textbf{Using source depth maps and visible source pixels in the target view as a mask}
\newline
As mentioned above, in realistic scenarios, it is challenging to provide ground-truth target depth maps with source RGB-D images. Therefore, our final variation model uses only the source RGB-D image as input to predict the image in a target view, as shown in Figure~\ref{fig:var2_vm_src_model}. This model will be referred to as \textit{No Depth Decoder, Visibility Mask - Source Depth Map (NDVM-Src)} for simplicity.

The encoder-decoder part for the pixel branch remains the same as in the baseline model. The loss function applied to this model is the same as the \textit{NDVM-Tgt} model (Eq \ref{eq:11}). Similar to the \textit{NDVM-Tgt} model, we trained this model with both sparse and dense visibility masks. The main difference of this model compared to \textit{NDVM-Tgt} is that it computes the warped target depth map from the source depth map. The warped target depth map is a sparse depth map, which contains less information when warping a source image into the target view.

\subsection{Experiments}
We conducted experiments on two challenging datasets: the synthetic car dataset from ShapeNet\cite{shapenet} and real driving scenes from KITTI\cite{kitti}. First, we trained and evaluated the baseline model and four different proposed variation models with the synthetic car dataset. Then we trained the models with real scenes from KITTI. Finally, we chose the most promising model to test on our dataset, PIV3CAMS. The metrics, mean absolute error (L1) and structural similarity (SSIM) index, widely used to assess reconstruction quality, were used to evaluate the models quantitatively.
In this section, we describe the details about the organization of the training and testing sets and technical implementations.

\subsubsection{Datasets: ShapeNet}
The synthetic car dataset from ShapeNet\cite{shapenet} consists of 8,195 car object files that can be rendered using 3D computer graphics software such as Blender.

We used 6,555 car objects for training and 1,640 for testing, as suggested in \cite{nvsmachines}. For training, we rendered 18 views for each car object by changing its azimuth between 0\degree and 360\degree with a step size of 20\degree. We then randomly sampled two views from each car object and used them as a pair of inputs to the networks. For testing, we rendered 180 views for each car object by changing its azimuth between 0\degree and 360\degree with a step size of 2\degree. For each car object, we chose 15 pairs of different source and target views, resulting in a total of 24,600 instances used during evaluation.

\subsubsection{Datasets: KITTI}
The KITTI dataset\cite{kitti} contains city road scenes captured from the frontal views of cars. They provide sub-datasets with different settings for various computer vision tasks.

We chose the sub-dataset for the \textit{odometry} task following \cite{nvsmachines}. Since the \textit{odometry} dataset does not come with depth maps, we downloaded the corresponding ground-truth sparse depth maps from the \textit{depth completion} dataset. We excluded sequences from the \textit{odometry} dataset if there were no corresponding depth maps. Additionally, we prepared dense depth maps using the depth completion model proposed by \cite{depthcompletion}.

In total, we used 21,432 frames to train the models and 1,071 frames for testing. For both training and testing, we randomly sampled two images within a three-frame difference. We also center-cropped each frame to form an image with dimensions of 256$\times$256.

\subsubsection{Datasets: PIV3CAMS}
As an example of a computer vision application for PIV3CAMS, we tested novel view synthesis on our proposed dataset. We selected three different ZED RGB-D scenes from the video set since they include depth maps and transformation data between frames. We did not perform any pre-processing on these sets except for cropping the center of each frame to match the dimensions of the other datasets, 256$\times$256. We prepared 2,000 frames from three different scenes, randomly sampled within a three-frame distance, as the source and target views for the test.

\subsection{Metrics}
\subsubsection{Mean Absolute Error (L1)}
The mean absolute error, also called L1 loss, is defined as
\begin{equation}
  L_{1} = \frac{\sum_{i=1}^{h} \sum_{j=1}^{w} |x(i,j) - y(i,j)|}{hw}
\end{equation}
where \(h\) and \(w\) denotes the height and width of an image. This metric is used to measure per-pixel color absolute differences between predicted target images (\(I^{pred}_{tgt}\)) and the ground-truth target images (\(I_{tgt}\)). 

\subsubsection{Structural SIMilarity (SSIM) Index}
Structural SIMilarity (SSIM) Index\cite{SSIM} measures the structural similarity between two images by calculating on various windows of images. The measure between two windows \(x \) and \(y \) of size \(N \times N\) is defined as
\begin{equation}
    SSIM(x,y) = \frac {(2\mu _{x}\mu _{y}+c_{1})(2\sigma _{xy}+c_{2})}{(\mu _{x}^{2}+\mu _{y}^{2}+c_{1})(\sigma _{x}^{2}+\sigma _{y}^{2}+c_{2})}
\end{equation}
where \(\mu_{x}, \mu_{y}\) are the average of \(x \) and \(y \), \( \sigma_{x}^{2},  \sigma_{y}^{2}\) are the variance of \(x \) and \(y \), \( \sigma_{xy}\) is covariance of \(x \) and \(y \), and \(c_{1}, c_{2}\) are variables added to stabilize the division. For our evaluation, we set window size \(N = 11 \) and \(c_{1}=0.01^{2}, c_{2}=0.03^{2}\) as proposed in \cite{SSIM}. This metric ranges from -1 to 1, indicating the perceptual quality of two images. Note that when two images are similar, the value is closer to 1.

\subsection{Training Details}
For a fair comparison between various models, we set the parameters the same for all training sessions. The networks were trained with the Adam optimizer, using a learning rate of 0.00006, \(\beta_{1}\) as 0.5 and \(\beta_{2}\) as 0.999. The weights for the losses were chosen so they have similar absolute values during training: \(\lambda_{L1} = 10, \lambda_{GAN} = 2, \lambda_{VGG} = 0.5\). The batch size was set to 16. The maximum number of training epochs was 200 with 375 iterations for the car dataset and 100 with 1,375 iterations for the KITTI dataset.
\section{Results and Discussion} \label{ch5}
\subsection{Novel view synthesis for Synthetic Objects}
\begin{figure}[h!]
  \centering
  \includegraphics[width = \textwidth]{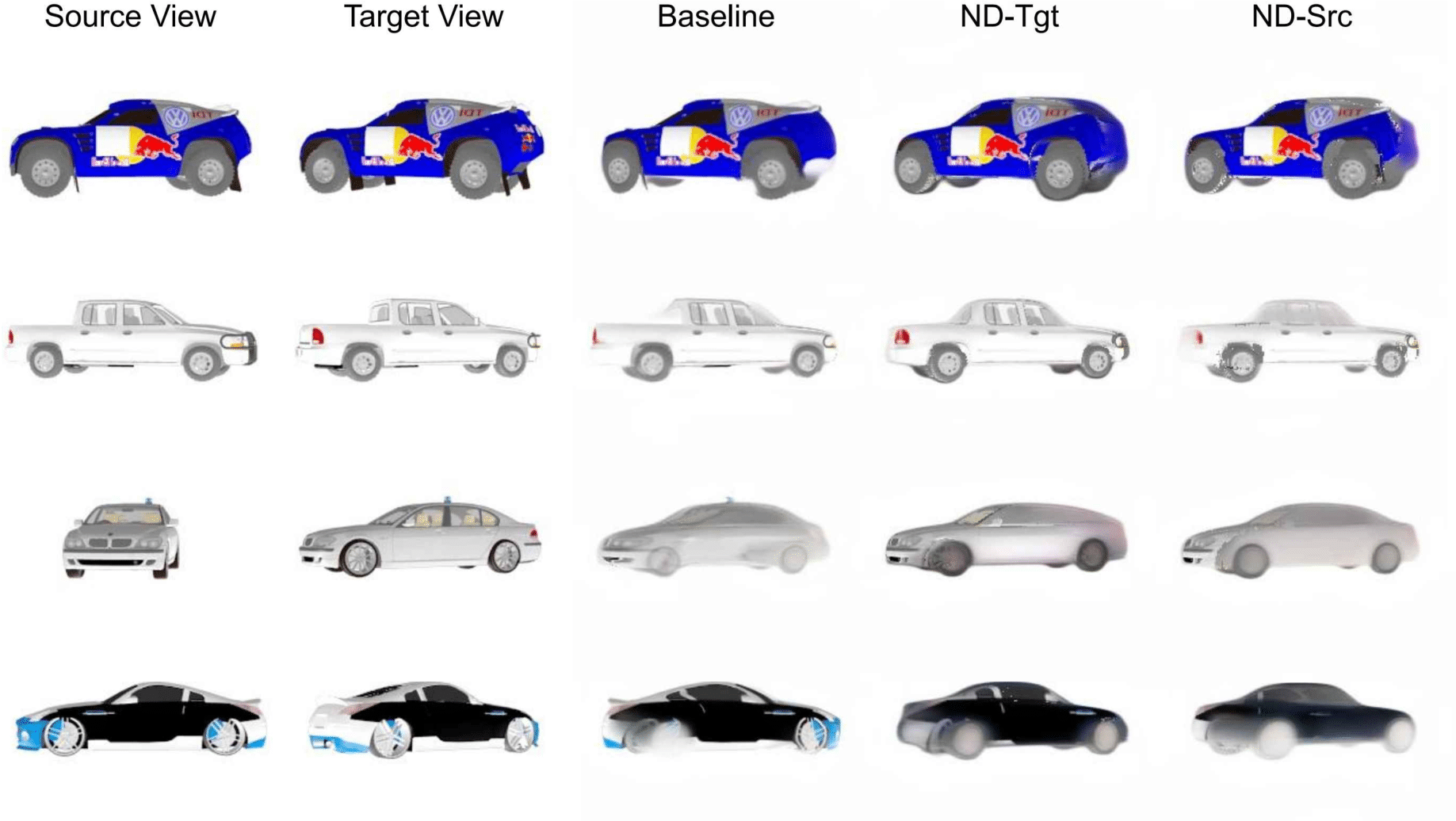}
  \includegraphics[width = \textwidth]{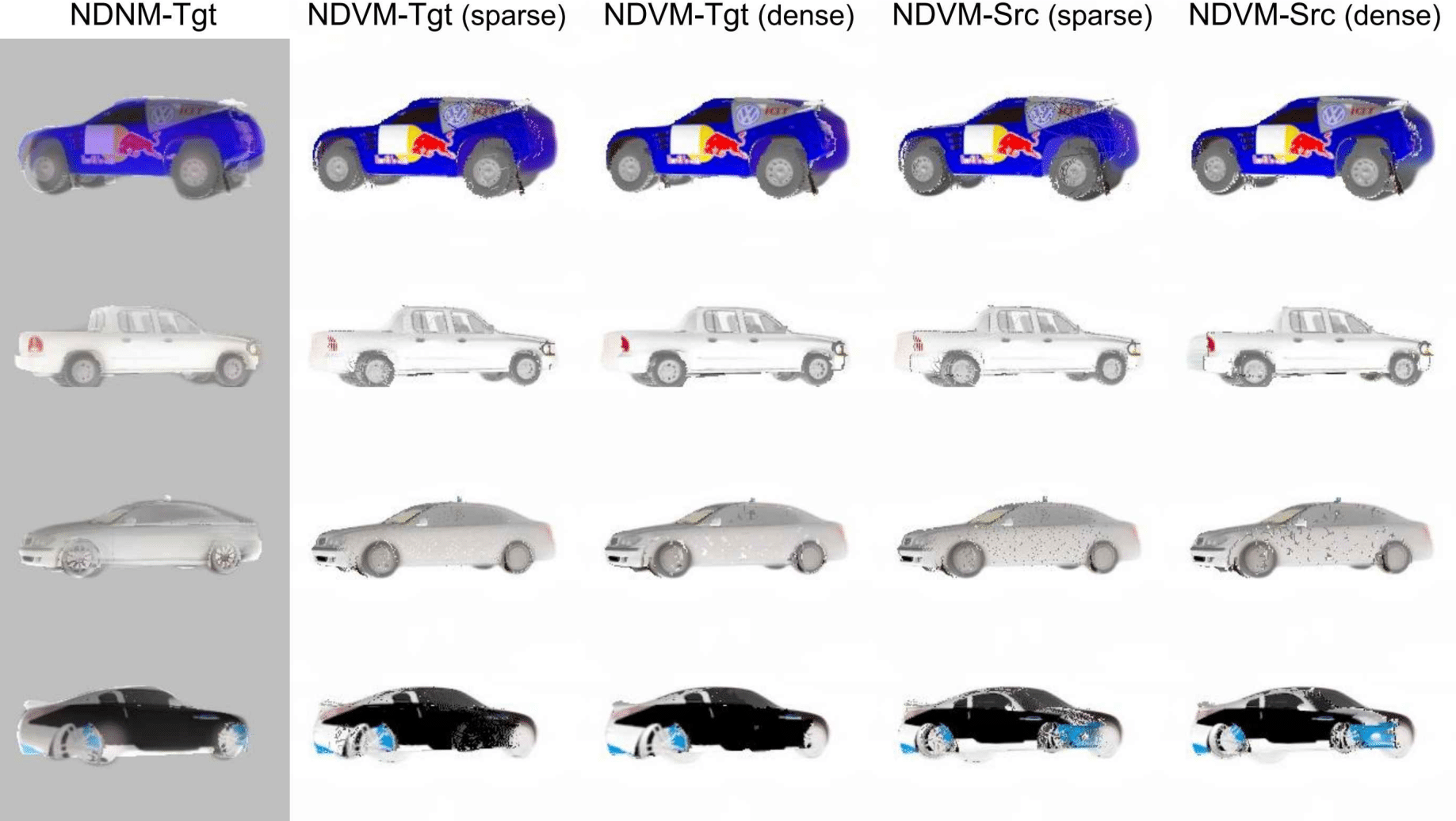}
  \caption{\textbf{Qualitative results of novel view synthesis on synthetic car objects\cite{shapenet}}. The rotation angles between the source and target view are 30\degree, 58\degree, 322\degree,  and 150\degree \space in clock-wise, from top to bottom rows, respectively.  The density of the visibility mask used is indicated inside the parenthesis like \textit{NDVM-Tgt (sparse)}.} 
  \label{fig:syncar_result}
\end{figure}

First, we present the results of the novel view synthesis for synthetic car data. The quantitative results are shown in Table~\ref{tab:car_result}. The \textit{NDVM-Tgt} model, which uses ground-truth target depth and source depth, performs slightly better than other models on both L1 and SSIM metrics. Note that in this model, the source depth is used only to generate the visibility mask. The results demonstrate that using ground-truth target depth generally synthesizes the target view better than using ground-truth source depth. The qualitative results are presented in Figure~\ref{fig:syncar_result}

\begin{table}[h]
    \centering
    \begin{tabular}{||c|c|c|c|c||}
    \hline \hline
        Model & $L_{1}$ & SSIM & Input Depth & Visibility Mask \\
    \hline \hline
        Baseline & 0.0467 & 0.8763 & - & - \\
    \hline
        ND-Tgt &  0.0499 & 0.8737 & Target depth & - \\
        ND-Src &  0.0495 & 0.8721 & Source depth & -\\
    \hline
        NDNM-Tgt & 0.4421 & 0.6861 & Target depth & -\\
    \hline
        NDVM-Tgt & 0.0455 & 0.8766 & Source and Target depth & sparse\\
        NDVM-Tgt & \textbf{0.0454} & \textbf{0.8814} & Source and Target depth & dense \\
    \hline
        NDVM-Src & 0.0467 & 0.8726 & Source depth & sparse \\
        NDVM-Src & 0.0468 & 0.8762 & Source depth & dense \\
        \hline \hline
    \end{tabular}
    \caption{Quantitative results of novel view synthesis on synthetic car objects\cite{shapenet}}
    \label{tab:car_result}
\end{table}

\subsubsection{Influence of using ground-truth depth in view synthesizing}
The influence of ground-truth depth on novel view synthesis performance can be observed by comparing the baseline model with the \textit{ND-Tgt} and \textit{ND-Src} models. Note that in real scenarios, it is challenging to obtain ground-truth target depth; however, we experimented with our models using ground-truth target depth to establish an upper bound.

\begin{figure}[p]
  \centering
  \includegraphics[width = \textwidth]{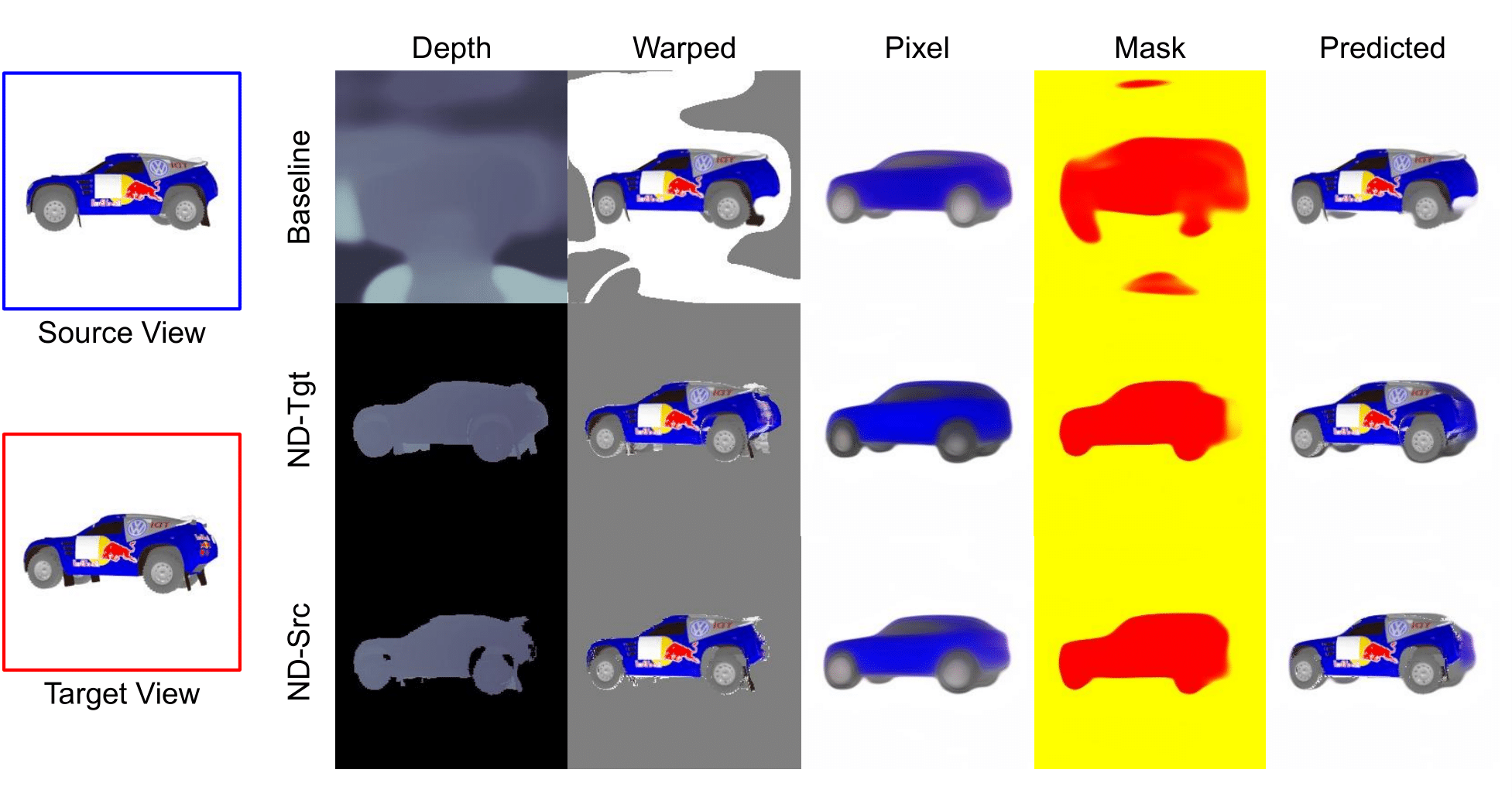}
  \includegraphics[width = \textwidth]{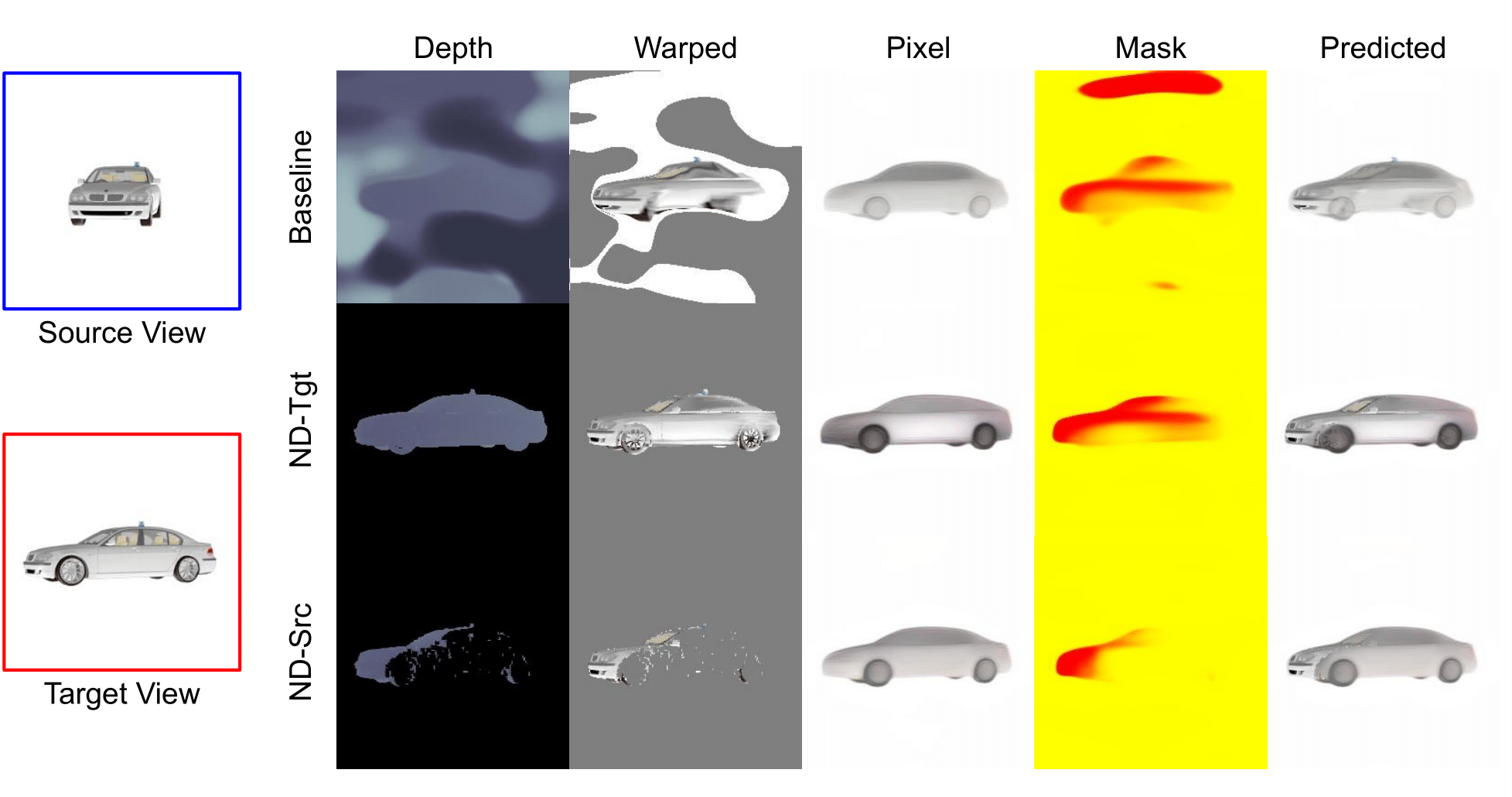}
  \caption{\textbf{Intermediate results of the baseline, \textit{ND-Tgt}, and \textit{ND-Src} models} The \texttt{Depth} column shows images used for inverse-warping projection. The results are shown in the \texttt{Warped} column. Note that for the baseline model, \texttt{Depth} is predicted by the depth decoder. The \textit{ND-Tgt} model uses the ground-truth target depth map for \texttt{Depth}. The \textit{ND-Src} model uses \texttt{Depth} as the warped target depth map computed from forward-warping. The \texttt{Pixel} and \texttt{Mask} columns show the output of the pixel decoder and mask decoder of each model, respectively. The final prediction is shown in the \texttt{Predicted} column.} 
  \label{fig:syncar_intermediate_result}
\end{figure}

Quantitatively, the baseline model performs better than the other two models on both L1 and SSIM metrics. The depth map predicted by the baseline model, as shown in Figure~\ref{fig:syncar_intermediate_result} is a learned depth mask (refer to \texttt{depth}) that optimizes the final predicted output in compliance with the fusion mask (refer to \texttt{mask}). Consequently, naively replacing the baseline depth mask with the ground-truth target map, as in the \textit{ND-Tgt} model, did not improve the model's performance. Indeed, the depth mask from the baseline model differs significantly from the ground-truth target depth map.


Qualitatively, the baseline, \textit{ND-Tgt}, and \textit{ND-Src} models all produce perceptually acceptable results. We observe that using the ground-truth depth map performs better in terms of transferring local details when the transformation between the source view and target view is small. On the other hand, the baseline model performs much better when there is a large view transformation.

As observed in Figure~\ref{fig:syncar_intermediate_result}, the results from the pixel decoder (refer to \texttt{pixel}) show that the pixel decoders are capable of producing well-structured shapes of the target view but lack detailed texture. The \texttt{mask} from the mask decoder determines which result to use for the final prediction. We observe that it learns the parts visible in the source and uses the \textit{warped image} for those parts. For parts not visible in the source, it uses the \textit{pixel image} predicted from the pixel branch. 


\subsubsection{Influence of using visibility mask in view synthesizing}
\begin{figure}[h!]
  \centering
  \includegraphics[width = 0.85\textwidth]{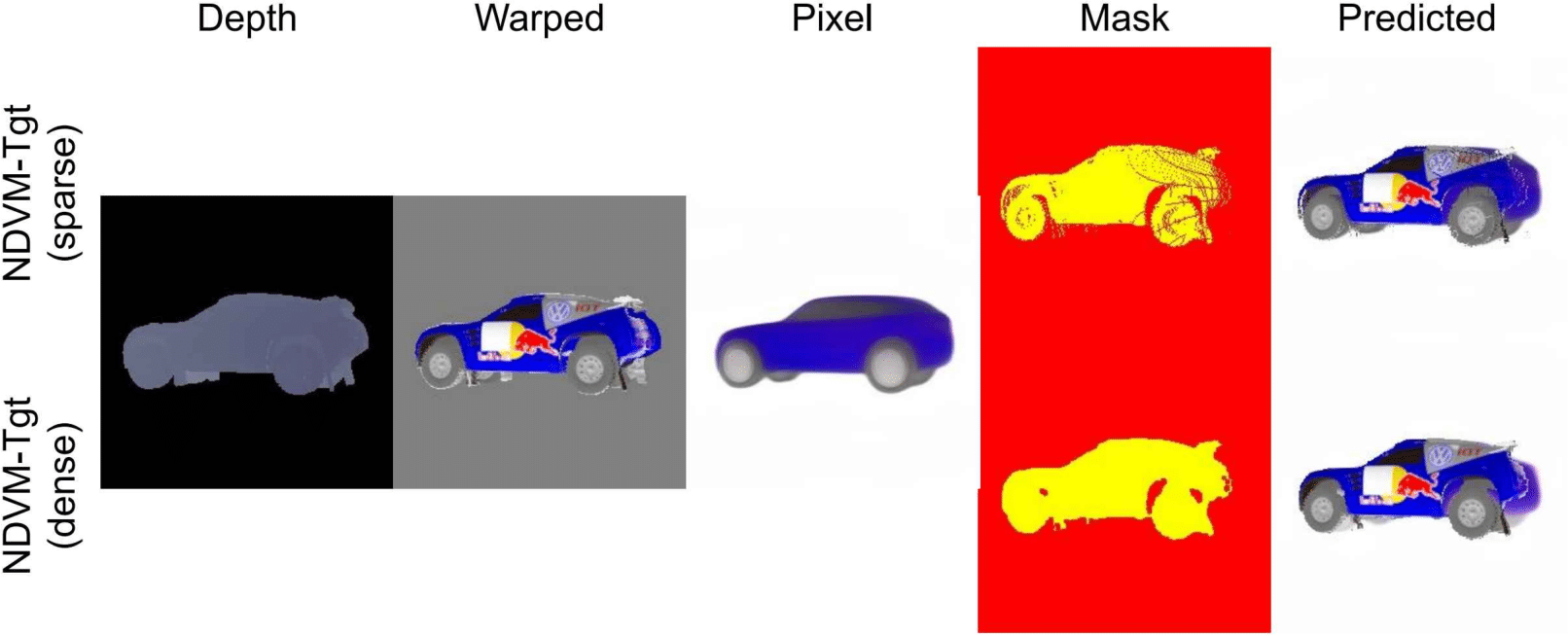}
  \includegraphics[width = 0.85\textwidth]{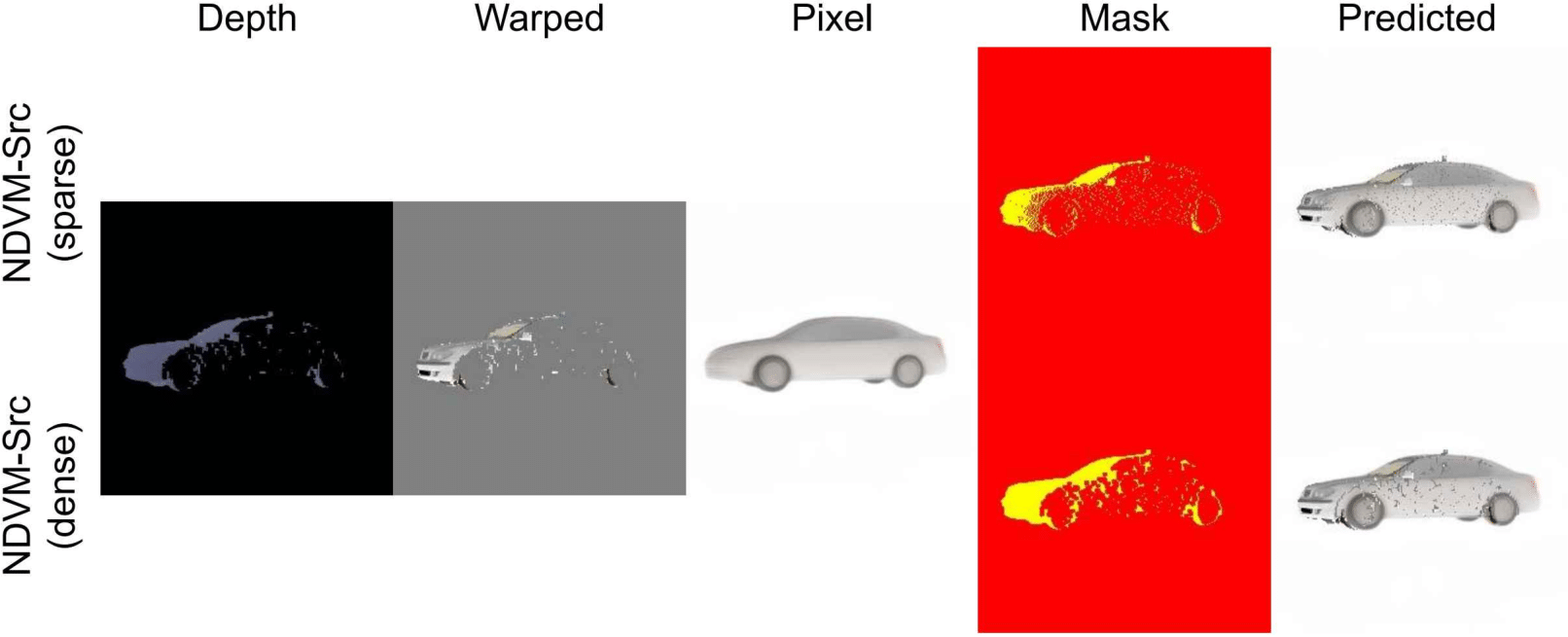}
  \caption{\textbf{Intermediate results of the \textit{NDVM-Tgt} and \textit{NDVM-Src} models} The \texttt{Depth} column shows images used for inverse-warping projection, and the results are shown in the \texttt{Warped} column. The \textit{NDVM-Tgt} model used the ground-truth target depth map for \texttt{Depth}. The \textit{NDVM-Src} model used \texttt{Depth} as a warped target depth map computed from forward-warping. The \texttt{Pixel} column shows the outputs of the pixel decoder, and the \texttt{Mask} column shows the sparse or dense visibility masks computed via forward-warping. } 
  \label{fig:syncar_VM_density}
\end{figure}

Instead of predicting the mask from the mask decoder, we experimented with using a visibility mask computed statistically via forward-warping. The \textit{NDVM-Tgt} and \textit{NDVM-Src} models use the visibility mask to merge the results of the pixel branch and the warped image. For each network, we synthesized the target view using either a sparse or dense visibility mask, as shown in Figure~\ref{fig:syncar_VM_density}. 

Quantitatively, using a statistically computed visibility mask (NDVM models) performs better than using a predicted mask (ND models). Specifically, \textit{NDVM-Tgt} with a dense visibility mask performs better than the baseline model on both L1 and SSIM metrics. Comparing the intermediate results(Figure~\ref{fig:syncar_intermediate_result} and Figure~\ref{fig:syncar_VM_density}) of these models, we observe that the visibility mask provides more precise information about the parts visible in the source view, whereas the predicted mask provides rough information. 

The density of the visibility mask affects the quantitative measures, with the dense visibility mask resulting in higher SSIM values. Qualitatively, the dense visibility mask produces smoother target results, especially in areas visible in the source view. However, the final predicted target results still suffer from speckles in the images for both types of visibility masks.
 
\subsection{Novel view synthesis for Real Scenes}
We also tested our proposed models on real scenes, where dense ground-truth depth maps are usually unavailable and the images are more complex than those in the synthetic car dataset. We trained and tested the baseline model and our proposed models with the KITTI dataset\cite{kitti}. Training with real scenes took significantly longer than training with the synthetic car dataset; it took eight days to train 100 epochs using a 12GB GPU (Tesla K40c). Due to limited resources and time, we could only train the baseline, \textit{ND-Tgt}, and \textit{ND-Src} models. Then, we used the trained encoder and pixel decoder of \textit{ND-Tgt} and \textit{ND-Src} to test the performance of the other models, \textit{NDVM-Tgt} and \textit{NDVM-Src}, respectively.

\begin{figure}[h!]
  \centering
  \includegraphics[height = 7.3cm]{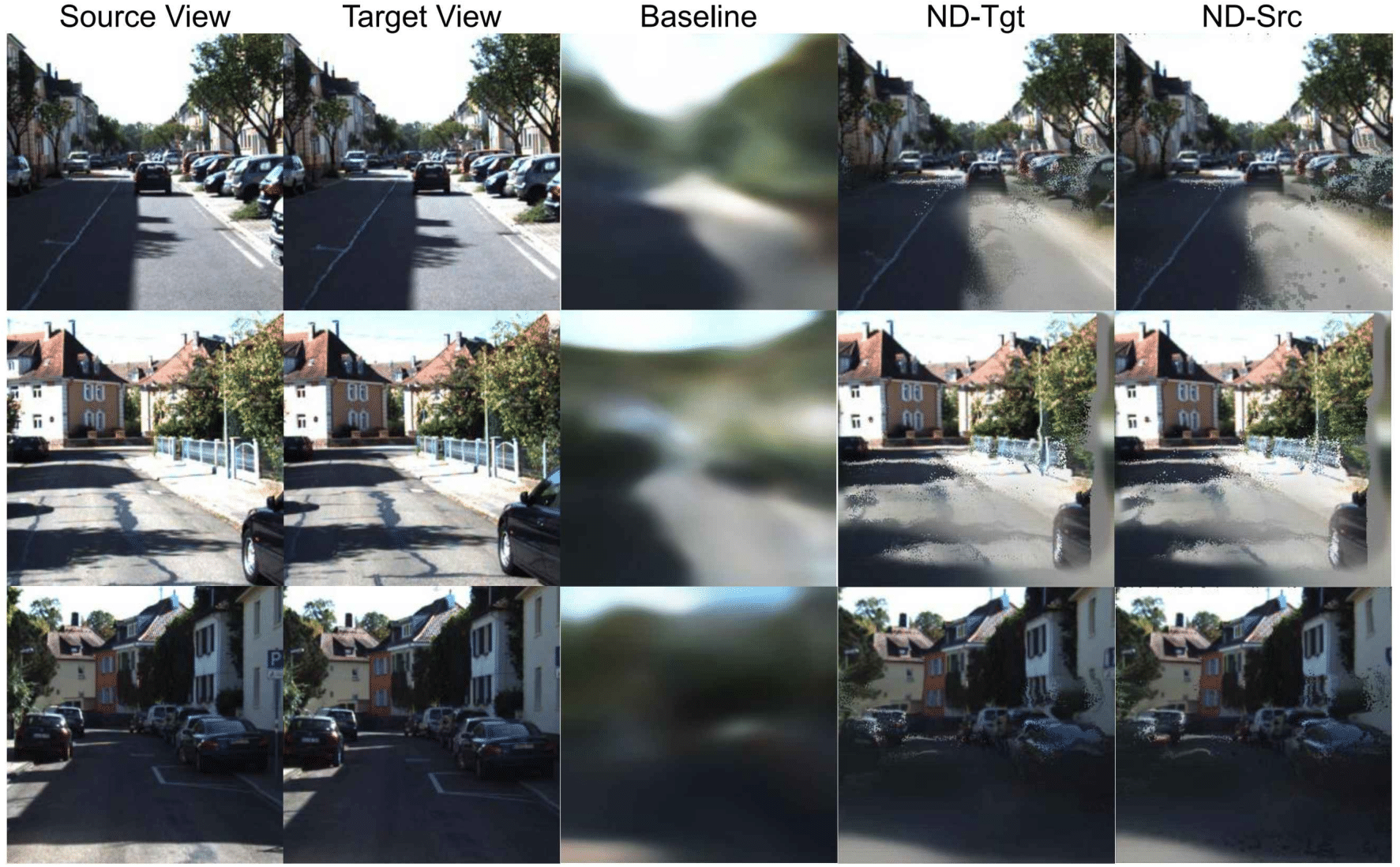}
  \includegraphics[height = 7.3cm]{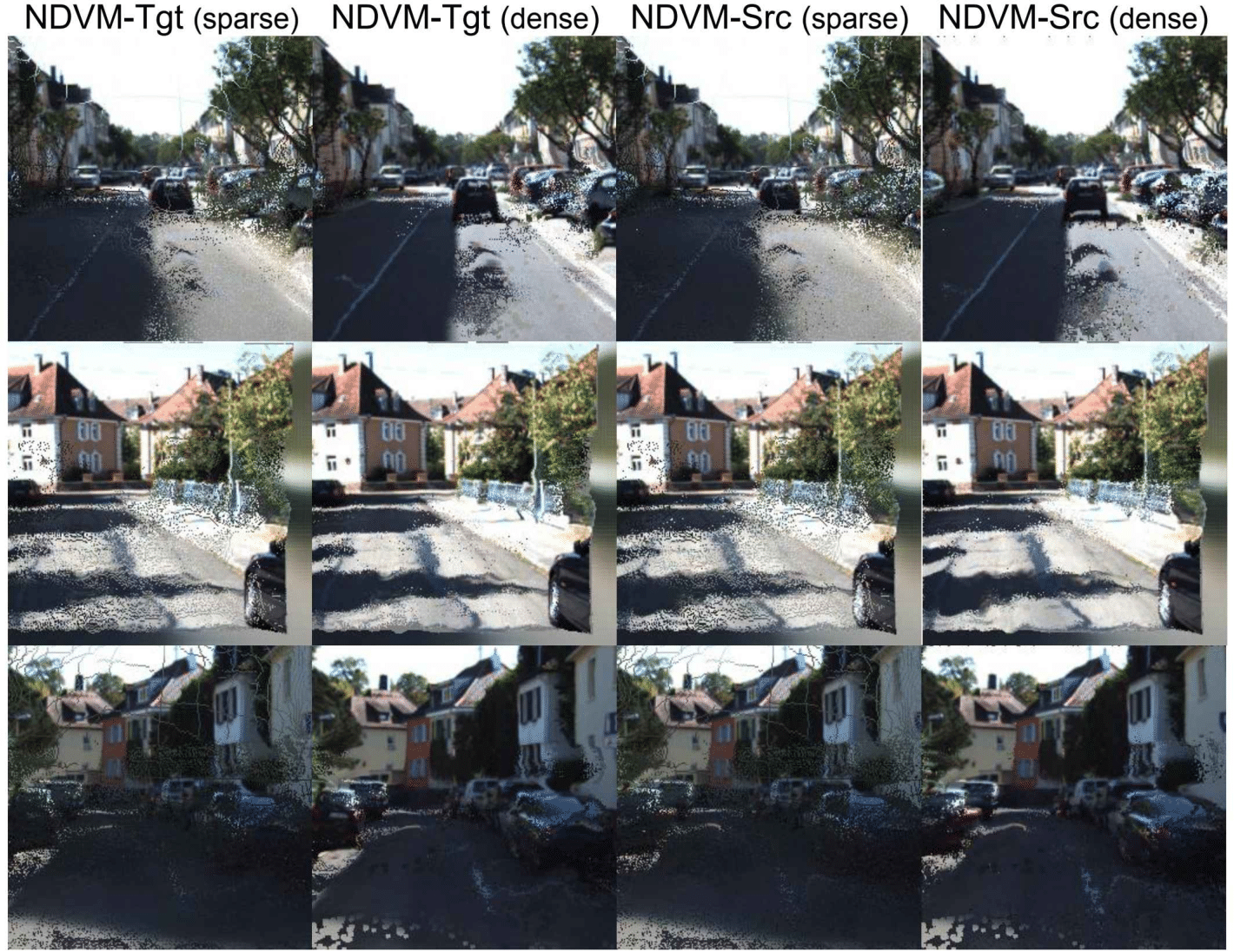}
  \caption{\textbf{Qualitative results of novel view synthesis on real scenes from KITTI dataset\cite{kitti}}. The frame gaps between the source and target views are 1, 2, and 3, from top to bottom rows, respectively. The density of the visibility mask used is indicated in parentheses, e.g., \textit{NDVM-Tgt (sparse)}.}
  \label{fig:kitti_result}
\end{figure}

The quantitative results of the novel view synthesis on real scenes are shown in Table~\ref{tab:kitti_result}. The \textit{ND-Tgt} model performs better than the other models, particularly outperforming the baseline model. Figure~\ref{fig:kitti_result} shows the qualitative results.

\begin{table}[h!]
    \centering
    \begin{tabular}{|c|c|c|c|c|}
    \hline \hline
        Model & $L_{1}$ & SSIM & Input Depth & Visibility Mask \\
    \hline \hline
        Baseline & 0.3409 & 0.2503 & - & - \\
    \hline
        ND-Tgt &  \textbf{0.2026} & \textbf{0.4216} & Target depth & - \\
        ND-Src &  0.2061 & 0.4136 & Source depth & -\\
    \hline
        NDVM-Tgt & 0.2139 & 0.3490 & Source and Target depth & sparse\\
        NDVM-Tgt & 0.2038 & 0.4017 & Source and Target depth & dense \\
    \hline
        NDVM-Src & 0.2141 & 0.3427 & Source depth & sparse \\
        NDVM-Src & 0.2044 & 0.3928 & Source depth & dense \\
        \hline \hline
    \end{tabular}
    \caption{Quantitative results of novel view synthesis on real scenes from KITTI dataset\cite{kitti}}
    \label{tab:kitti_result}
\end{table}

The experiment on real scenes shows that the baseline model fails to predict the target views, while the other methods using ground-truth depth maps can generate at least some parts of the target views. A common issue noticed in all models is that the output from the pixel branch is not satisfactory. One possible explanation could be the small size of the latent vector used during training. Since real scenes are more complex than synthetic car images, using the same size latent vector as the synthetic car dataset may not be sufficient to capture the features of real scenes.

The final output of the baseline model is almost identical to the result of the pixel branch, indicating that the warped target image from the depth branch is not adequate. The depth decoder did not learn the target depth properly, resulting in poor warped target images. While \cite{nvsmachines} demonstrated good performance with real scene images, it is likely that specific parameters need to be applied to make the depth branch work effectively.

\begin{figure}[h!]
  \centering
  \includegraphics[height = 5cm]{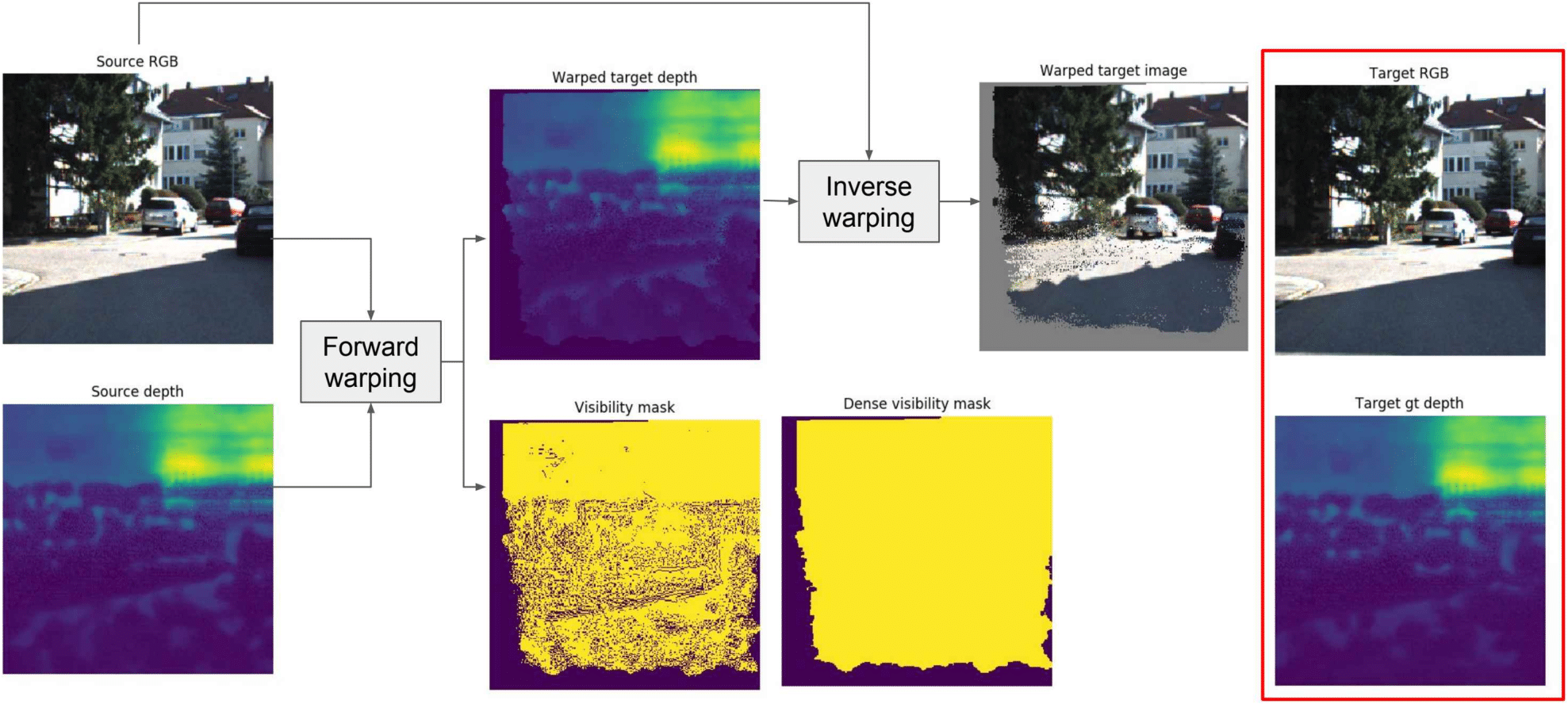}
  \includegraphics[height = 5cm]{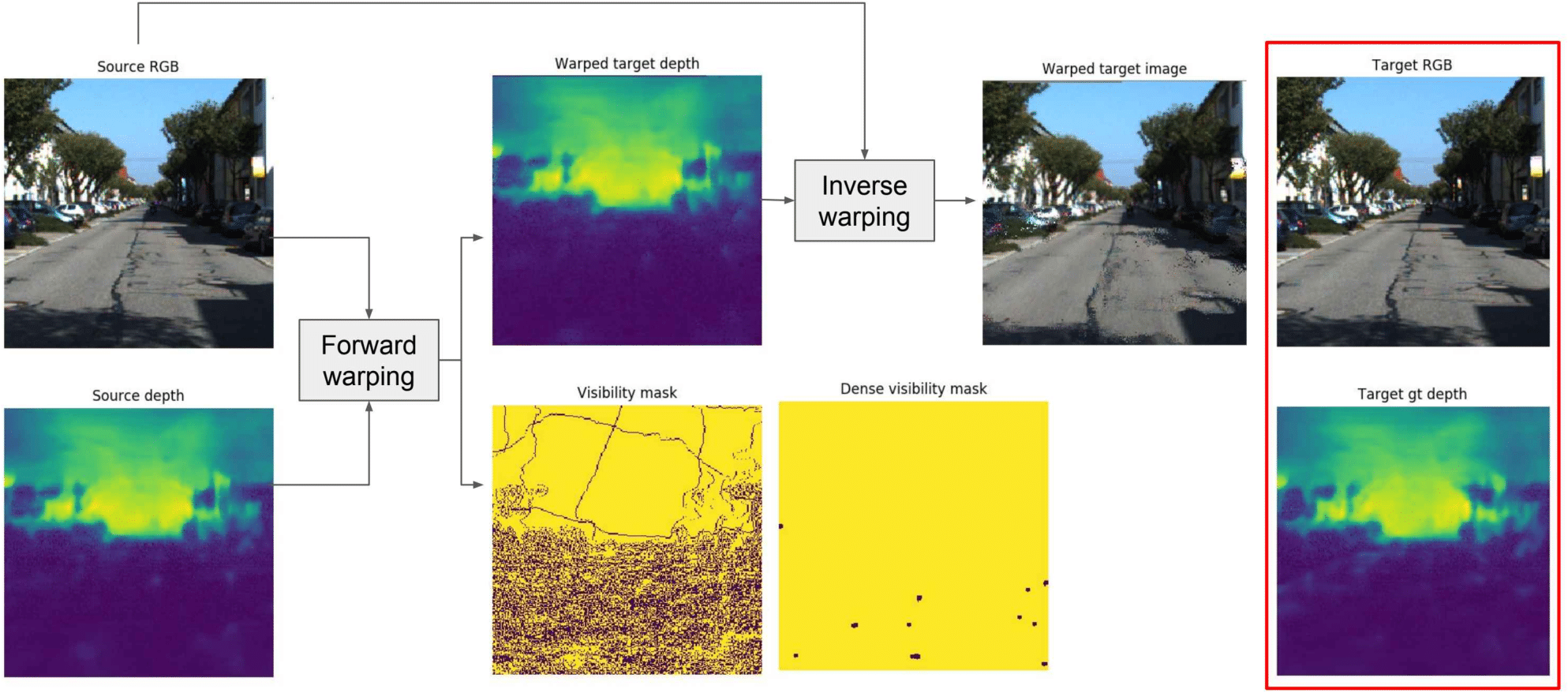}
  \caption{\textbf{3D geometric projection of real scenes}. Source and target depth maps are completed using the \cite{depthcompletion} algorithm. Brighter pixels indicate further distances. Yellow areas in the visibility mask show source pixels visible in the target view.} 
  \label{fig:kitti_warping}
\end{figure}

In contrast, the other models (\textit{ND-Tgt}, \textit{ND-Src}, \textit{NDVM-Tgt}, and \textit{NDVM-Src}) produced reasonable results in some parts of the images, thanks to the ground-truth depth. As shown in  Figure~\ref{fig:kitti_warping}, it is possible to warp the source RGB-D image into a target view using 3D geometrical projections. These models use the results from the pixel branch to inpaint the missing parts from the warped target images, creating the final predicted target images. This result aligns with the observations made with the synthetic car data.

\begin{figure}[h!]
  \centering
  \includegraphics[width = 0.8\textwidth]{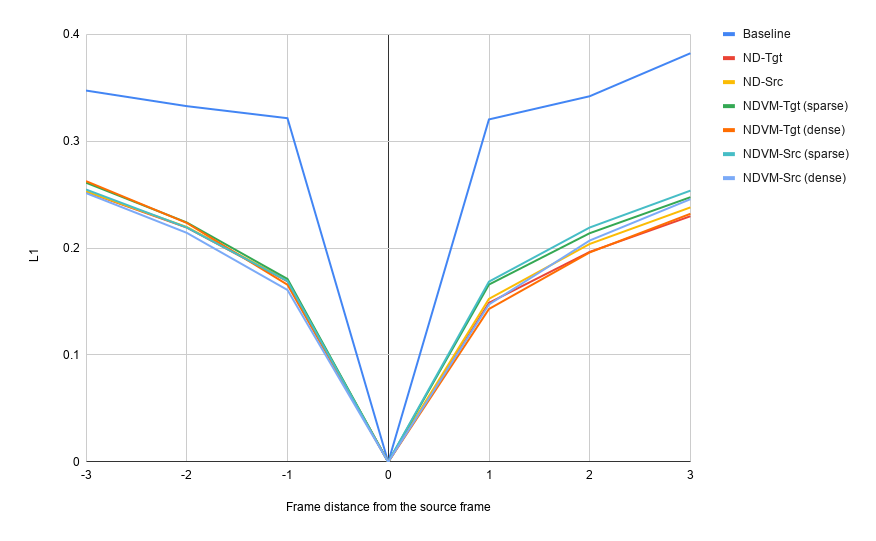}
  \caption{\textbf{Average L1 loss based on frame distance between the source and target images from KITTI dataset}. The horizontal axis shows the frame distance from the source view. Negative values indicate the target view is before the source view; positive values indicate it is after.} 
  \label{fig:kitti_frame_L1}
\end{figure}

Our experiment demonstrated that with a deficient result from the pixel branch, the model's performance relies heavily on the results from the depth branch, specifically the \textit{warped target images}. For example, using the dense visibility mask produces a smoother final result by incorporating more pixels from the warped target images. Among the NDVM models, those using dense visibility masks scored lower L1 and higher SSIM. Additionally, when the source and target images are closer in frames, more parts of the target view are visible from the source view, resulting in better-predicted target images (Fig~\ref{fig:kitti_frame_L1}).

\subsection{Novel view synthesis on the PIV3CAMS dataset }
Finally, we tested one of our models trained on the real scenes on the PIV3CAMS dataset. We used the \textit{ND-Tgt} model for view synthesis as it showed the best performance among the models (Fig~\ref{fig:kitti_result}). The calculated L1 and SSIM scores are 0.2571 and 0.3942, respectively. 

\begin{figure}[h!] 
  \centering
  \includegraphics[width = \textwidth]{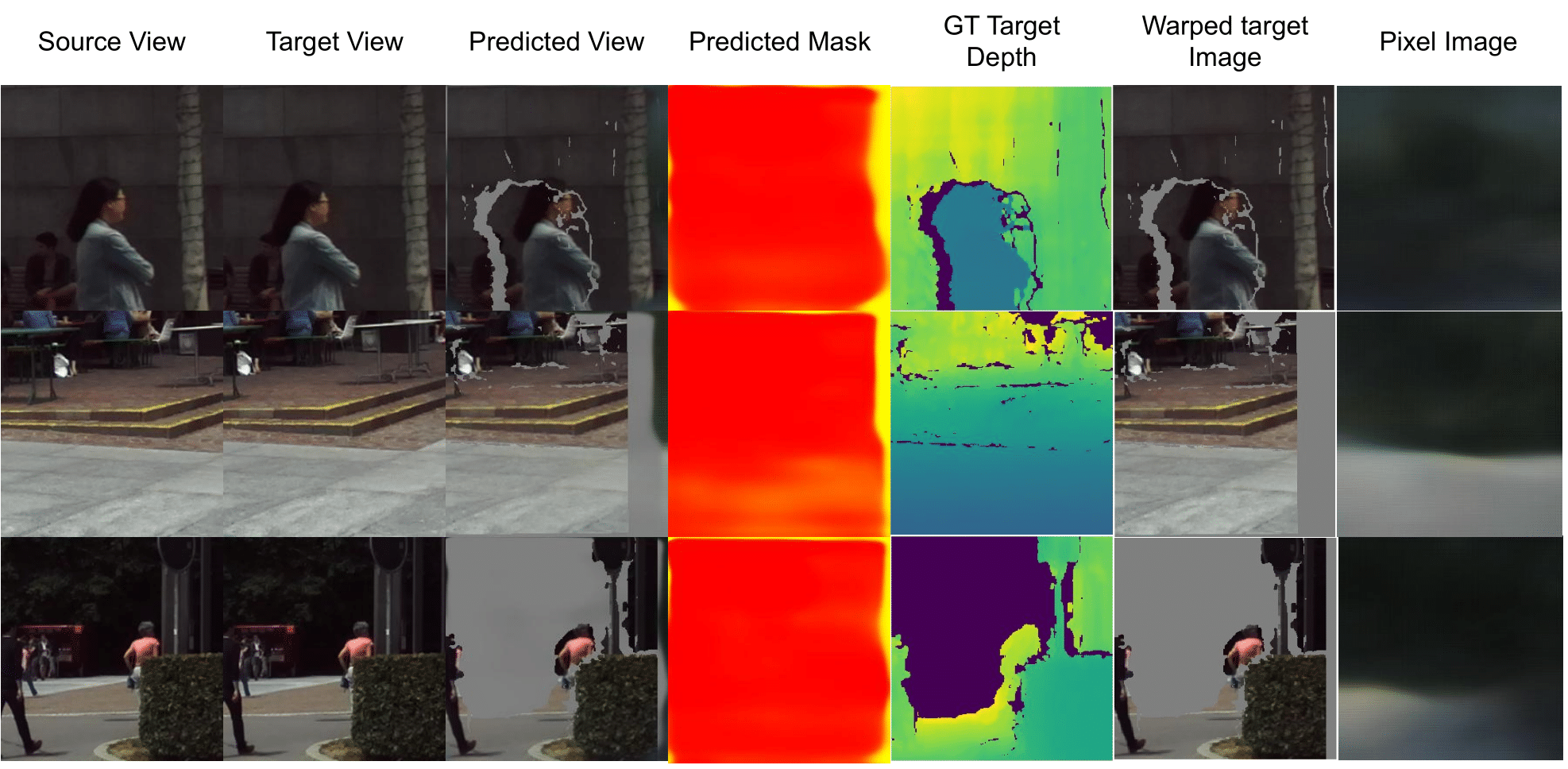}
  \caption{\textbf{The novel view synthesizing results using PIV3CAMS dataset}. The ground-truth target depth map is a sparse depth map, where brighter pixels indicate greater distances than darker pixels. Depth values set to zero are represented in violet.} 
  \label{fig:piv3cam_result}
\end{figure}

In Figure~\ref{fig:piv3cam_result}, we present some visual results. From these results, we observed that the predicted masks prefer the result from the warped target images over the pixel images. Therefore, except for the edge parts of the images, most of the predicted views are the same as the warped target images. Since the ground-truth target depth map used as an input was not a dense depth map, it leaves missing pixel areas as gray in the warped images and thus in the predicted views. Despite this, we have successfully demonstrated the potential of using PIV3CAMS for novel view synthesis applications.

\subsection{Limitations}
\subsubsection{The performance of the pixel branch} 
In the previous section, we have shown that the role of the mask image is to fuse the two results: the warped target image and the pixel image. The mask uses the warped target image for the areas visible in the source view, and the pixel image for the rest. In other words, the occluded parts will be filled in with the results from the pixel branch. Consequently, the performance of the pixel branch plays an essential role in novel view synthesis.

Observing the results from the pixel branches of our proposed models, we noticed that they do not predict the colors of the target view precisely, hence degrading the performance. This issue is evident not only in real scene cases but also in the synthetic car examples. The results of the \textit{NDNM-Tgt} model and the pepper-noised results from the NDVM models in Figure~\ref{fig:syncar_result} show that the predicted colors from the pixel branch do not match perfectly with the ground-truth target view. Additionally, the pixel images in Figure~\ref{fig:syncar_intermediate_result}  are not sharp enough compared to the ground-truth target view. Therefore, designing a network to produce better pixel images could enhance the performance of novel view synthesis.

\subsubsection{The sparsity of the depth map}
From the real scene experiments in Sections 5.2 and 5.3, it becomes clear that having a dense depth map is essential for novel view synthesis. The original KITTI dataset provides an extremely sparse depth map, with less than 10\% of the pixels having depth values. Since the depth map plays a vital role in computing the warped target image, we needed to complete the depth maps of KITTI's. Even if these completed depth maps are not perfect, they provide reasonably good results for view synthesis.
On the other hand, when we tested the model with the PIV3CAMS dataset, which has sparse depth maps, holes appeared in the predicted images. As our proposed models are not able to fill those holes, the methods are limited to dense depth maps. Therefore, adopting a network to fill those holes would expand the boundaries of novel view synthesis using RGB-D images 




\section{Conclusion}\label{ch6}

In this work, we introduced the PIV3CAMS dataset, which consists of paired images and videos from three different cameras (Huawei P20, Canon 5D Mark IV, and ZED stereo camera). The PIV3CAMS dataset comprises 8,385 pairs of images and 82 pairs of videos. Each pair of images includes 2 RGB images, 2 RAW images, 2 RGB-D images (left and right), and one confidence map, resulting in about 75.4 million individual photos in the image set. The videos, captured at 30 frames per second with durations between 25 and 50 seconds, total more than 250 million frames. The data was collected from Zurich, Switzerland, and Cheonan, South Korea, with carefully chosen locations to cover various indoor and outdoor scenes. Additionally, the dataset includes scenes captured at different times of the day, including night scenes. We anticipate that this dataset will advance multiple computer vision algorithms using machine learning approaches, such as view interpolation, image matching, and image/video super-resolution.

We also investigated the novel view synthesis problem as an example of a computer vision application. First, we reproduced one of the state-of-the-art novel view synthesis methods as a baseline. This baseline model consists of three branches that predict the target depth, target pixel image, and a mask for fusion. We then modified the depth branch by using ground-truth target or source depth maps as input. We experimentally showed with synthetic data that having depth information is useful, especially when the transformation between the source and target view is small. Furthermore, we trained our models with real RGB-D images from the KITTI dataset, demonstrating that it is possible to generalize the approaches to real-world imagery. Lastly, we tested the models trained with the KITTI dataset on the PIV3CAMS dataset to demonstrate the potential of using our proposed dataset for the novel view synthesis task.

In conclusion, our work covered the entire process of computer vision, from data collection to application. For future work, we suggest annotating the dataset with several object classes to expand its usage for more computer vision tasks. Additionally, collecting more data from different locations could increase the variety of the dataset. In terms of novel view synthesis tasks, we recommend using the PIV3CAMS dataset to train networks, as it contains more diverse scenes than the KITTI dataset. From our experiments, the warped image using the RGB-D source view has perceptually acceptable results except for some noise and holes. Therefore, an interesting direction for future research is to tackle the novel view synthesis problem with inpainting and denoising techniques.

\bibliographystyle{IEEEtran}
\bibliography{tails/references}

\end{document}